\documentclass[11pt,twoside]{article}
\usepackage{fullpage}

\usepackage{epsf}
\usepackage{fancyhdr}
\usepackage{graphics}
\usepackage{graphicx}
\usepackage{psfrag}
\usepackage{microtype}
\usepackage{subfigure}
\usepackage{algorithmic}
\usepackage{color,xcolor}
\usepackage{caption}
\usepackage{subcaption}
\usepackage{subfigure}

\usepackage{algorithm,algorithmic}

\usepackage{color}

\usepackage{amsthm}
\usepackage{amsfonts}
\usepackage{amsmath}
\usepackage{amssymb,bbm}

\usepackage{mathrsfs}

\usepackage{url}
\usepackage[colorlinks,linkcolor=magenta,citecolor=blue, pagebackref=true,backref=true]{hyperref}
\renewcommand*{\backrefalt}[4]{%
    \ifcase #1 \footnotesize{(Not cited.)}%
    \or        \footnotesize{(Cited on page~#2.)}%
    \else      \footnotesize{(Cited on pages~#2.)}%
    \fi}

\usepackage{nicefrac}
\usepackage{comment}

\usepackage{chngpage}

 \usepackage{tabularx}%

\usepackage{enumitem}
\usepackage{booktabs}
\usepackage{caption}

\usepackage{bm,bbm}
\usepackage{mathtools}
\usepackage{cleveref}

\usepackage{eucal}

\usepackage{scalerel}

\def\ball{\mathbb{B}}

\newcommand{\stepsize}{\gamma}

\newcommand{\real}{\ensuremath{\mathbb{R}}}

\newcommand{\density}{p}

\newcommand{\abss}[1]{\left| #1 \right |}

\newcommand{\sphere}{\ensuremath{\mathbb{S}}}

\newcommand{\mydefn}{\ensuremath{:=}}

\newcommand{\smoothness}{L} 

\newcommand{\defn}{:=}

\newcommand{\matsnorm}[2]{|\!|\!| #1 | \! | \!|_{{#2}}}
\newcommand{\vecnorm}[2]{\| #1\|_{#2}}

\newcommand{\opnorm}[1]{\ensuremath{\matsnorm{#1}{\tiny{\mbox{op}}}}}

\newcommand{\inprod}[2]{\ensuremath{\langle #1 , \, #2 \rangle}}

\newcommand{\kull}[2]{\ensuremath{D_{\text{KL}}(#1\; \| \; #2)}}

\newcommand{\Exs}{\ensuremath{{\mathbb{E}}}}
\newcommand{\Prob}{\ensuremath{{\mathbb{P}}}}

\DeclareMathOperator{\var}{var}

\newtheoremstyle{named}{}{}{\itshape}{}{\bfseries}{.}{.5em}{\thmnote{#3's }#1}
\theoremstyle{named}

\theoremstyle{plain}

\newtheorem{theorem}{Theorem}
\newtheorem{proposition}{Proposition}

\newtheorem{lemma}{Lemma}

\newlength{\widebarargwidth}
\newlength{\widebarargheight}
\newlength{\widebarargdepth}

\makeatletter
\long\def\@makecaption#1#2{
        \vskip 0.8ex
        \setbox\@tempboxa\hbox{\small {\bf #1:} #2}
        \parindent 1.5em  
        \dimen0=\hsize
        \advance\dimen0 by -3em
        \ifdim \wd\@tempboxa >\dimen0
                \hbox to \hsize{
                        \parindent 0em
                        \hfil
                        \parbox{\dimen0}{\def\baselinestretch{0.96}\small
                                {\bf #1.} #2
                                }
                        \hfil}
        \else \hbox to \hsize{\hfil \box\@tempboxa \hfil}
        \fi
        }
\makeatother

\long\def\comment#1{}
\definecolor{battleshipgrey}{rgb}{0.52, 0.52, 0.51}
\definecolor{darkgray}{rgb}{0.66, 0.66, 0.66}
\definecolor{darkgreen}{rgb}{0.0, 0.2, 0.13}
\definecolor{darkspringgreen}{rgb}{0.09, 0.45, 0.27}
\definecolor{dukeblue}{rgb}{0.0, 0.0, 0.61}
\definecolor{olivedrab7}{rgb}{0.24, 0.2, 0.12}
\definecolor{darkblue}{rgb}{0.0, 0.0, 0.55}
\definecolor{darkscarlet}{rgb}{0.34, 0.01, 0.1}
\definecolor{candyapplered}{rgb}{1.0, 0.03, 0.0}
\definecolor{ao(english)}{rgb}{0.0, 0.5, 0.0}
\definecolor{applegreen}{rgb}{0.55, 0.71, 0.0}

\newcommand{\state}{x}

\newcommand{\action}{a}

\newcommand{\kiter}{m}

\newcommand{\MyState}{X}
\newcommand{\State}{\MyState}

\newtheorem{example}{Example}

\newcommand{\rhofix}{\criticalradius^{\mathrm{fix}}}
\newcommand{\policy}{\pi}

\newcommand{\errtermsobo}{Z^{\mathrm{sobo}}}

\newcommand{\dudley}{\mathcal{J}}

\newcommand{\criticalradius}{\rho}
\newcommand{\critradistar}{\criticalradius^*}
\newcommand{\critradicross}{\criticalradius^{\mathrm{cross}}}
\newcommand{\critradismpl}{\criticalradius^{\mathrm{smpl}}}

\def\ll{\lVert}

\newtheorem{assumption}{Assumption}
\newcommand{\simiid}{\stackrel{\mathrm{i.i.d.}}{\sim}}
\newcommand{\numobs}{\ensuremath{n}}

\usepackage{natbib}

\newcommand{\usedim}{\ensuremath{d}}

\newcommand{\Reward}{R}
\newcommand{\reward}{r}

\newcommand{\totalvariation}{d_{\mathrm{TV}}}

\newenvironment{carlist}
 {\begin{list}{$\bullet$}
 {\setlength{\topsep}{0in} \setlength{\partopsep}{0in}
  \setlength{\parsep}{0in} \setlength{\itemsep}{\parskip}
  \setlength{\leftmargin}{0.07in} \setlength{\rightmargin}{0.08in}
  \setlength{\listparindent}{0in} \setlength{\labelwidth}{0.08in}
  \setlength{\labelsep}{0.1in} \setlength{\itemindent}{0in}}}
 {\end{list}}

\newcommand{\ltwospace}{\mathbb{L}^2}

\newcommand{\randreward}{R}

\newcommand{\sfactor}{\alpha}

\newcommand{\fbar}{\ensuremath{\bar{f}}}

\newcommand{\sobopstatnorm}[3]{\vecnorm{#1}{\mathbb{W}^{1, #2} (#3)}}

\newcommand{\sobokpnorm}[4]{\vecnorm{#1}{\mathbb{W}^{#2, #3} (#4)}}

\newcommand{\fakerefassumelip}[1]{\hyperref[assume:smooth-high-order]{{\color{magenta} {\upshape\textbf (}{\upshape{\textbf{Lip}}#1}{\upshape\textbf )}}} }

\newcommand{\convSet}{\Fclass}
\newcommand{\shiftConvSet}{\Fclass_*}

\newcommand{\valuefunc}{v}

\newcommand{\ValFun}{\valuefunc}

\DeclareFontFamily{U}{mathx}{}
\DeclareFontShape{U}{mathx}{m}{n}{<-> mathx10}{}
\DeclareSymbolFont{mathx}{U}{mathx}{m}{n}

\DeclareMathAccent{\widecheck}{0}{mathx}{"71}

\newcommand{\lammin}{\lambda_{\min}}
\newcommand{\lammax}{\lambda_{\max}}

\newcommand{\Event}{\mathcal{E}}

\usepackage{mathbbol}

\long\def\comment#1{}

\newcommand{\ValPol}{\ensuremath{\ValFun^\star}}
\newcommand{\ValTrue}{\ValPol}

\newcommand{\rhohat}{\widehat{\rho}}

\newcommand{\Fclass}{\ensuremath{\mathcal{F}}}

\newcommand{\eucnorm}{\abss}

\usepackage[framemethod=TikZ]{mdframed}
\newcommand{\smoothreg}[1]{\smoothness_{\mathrm{reg}, #1}}

\newcommand{\myframe}[1]{
\begin{mdframed}[backgroundcolor=black!1, roundcorner=5pt]
  #1
\end{mdframed}
}

\newcommand{\rhofixtilde}{\widetilde{\rho}}
\newenvironment{narrowpara}
  {\par\addvspace{\smallskipamount}\narrower\noindent\ignorespaces}
  {\par\addvspace{\smallskipamount}}

\newcommand{\myassumption}[4]{
  \setlist[enumerate,1]{leftmargin=#4}
\myframe{
	\begin{enumerate}[label={ \bf{{{(#1)}}}}]
		\item \label{#2} {#3}
	\end{enumerate}
        }
}

\newcommand{\goodendex}{$\clubsuit$}

\newcommand{\drift}{b}
\newcommand{\covMat}{\Lambda}

\newcommand{\BM}{B}

\newcommand{\Ctwospace}{\holderspace^2}

\newcommand{\generator}{\mathcal{A}}

\newcommand{\sobonorm}[2]{\vecnorm{#1}{\mathbb{H}^1 (#2)}}
\newcommand{\soboinprod}[3]{\inprod{#1}{#2}_{\mathbb{H}^1 (#3)}}

\newcommand{\ltwonorm}[2]{\vecnorm{#1}{\mathbb{L}^2 (#2)}}
\newcommand{\lpnorm}[3]{\vecnorm{#1}{\mathbb{L}^{#2} (#3)}}
\newcommand{\ltwoinprod}[3]{\inprod{#1}{#2}_{\mathbb{L}^2 (#3)}}

\newcommand{\ftil}{\widetilde{f}}

\newcommand{\ctrlMat}{M}
\newcommand{\actCost}{G}
\newcommand{\ExpVal}{f}
\newcommand{\ExpValTrue}{\ExpVal^*}
\newcommand{\termReward}{y}
\newcommand{\termRewardRand}{Y}
\newcommand{\ExpValHat}{\widehat{\ExpVal}}

\newcommand{\pq}{q}

\newcommand{\strucinprod}[3]{\inprod{#1}{#2}_{\mathscr{E}, #3}}

\newcommand{\strucnorm}[2]{\vecnorm{#1}{\mathscr{E}, #2}}

\newcommand{\energynorm}[4]{\vecnorm{#1}{\mathscr{E}^{#2, #3}, #4}}

\newcommand{\dynanorm}[4]{\vecnorm{#1}{\mathscr{V}^{#2, #3}, #4}}

\newcommand{\strucball}{\ball_{\mathscr{E}, T}}

\newcommand{\strongnorm}[2]{\vecnorm{#1}{\mathscr{V}, #2}}
\newcommand{\stronginprod}[3]{\inprod{#1}{#2}_{\mathscr{V}, #3}}

\newcommand{\rmax}{r_{\max}}

\newcommand{\smpltime}[2]{t_{#1}{(#2)}}
\newcommand{\holderspace}{\mathscr{X}}
\newcommand{\errterm}{Z}

\newcommand{\errtermgen}{\errterm^{\mathrm{cross}}}

\newcommand{\errtermsmpl}{\errterm^{\mathrm{smpl}}}
\newcommand{\errtermmain}{\errterm^*}

\begin{document}

\begin{center}
{\bf{\LARGE{Is RL fine-tuning harder than regression? A PDE learning approach for diffusion models}}}

\vspace*{.2in}
{\large{
 \begin{tabular}{cc}
  Wenlong Mou$^{ \diamond}$ 
 \end{tabular}

}

\vspace*{.2in}

 \begin{tabular}{c}
 Department of Statistical Sciences, University of Toronto$^{\diamond}$
 \end{tabular}

}

\begin{abstract}
   We study the problem of learning the optimal control policy for fine-tuning a given diffusion process, using general value function approximation. We develop a new class of algorithms by solving a variational inequality problem based on the Hamilton-Jacobi-Bellman (HJB) equations. We prove sharp statistical rates for the learned value function and control policy, depending on the complexity and approximation errors of the function class. In contrast to generic reinforcement learning problems, our approach shows that fine-tuning can be achieved via supervised regression, with faster statistical rate guarantees.
\end{abstract}
\end{center}

\section{Introduction}
Recent years have seen a surge of interest in the use of reinforcement learning (RL) in post-training optimization of deep generative models. By viewing the progressive generation process as a Markov decision process, RL provides a powerful framework for fine-tuning the generative model to achieve certain objectives, while maintaining the quality of generated samples. This learning paradigm allows us to optimize the generative model using only reward signals, and has been successfully applied to both auto-regressive language models~\cite{ziegler2019fine,ouyang2022training} and diffusion models~\cite{black2023training,fan2023optimizing}. Depending on the nature of the reward signal, RL has been applied to various tasks, including aligning the model with human preferences~\cite{christiano2017deep}, reasoning with verifiable rewards~\cite{shao2024deepseekmath}, and scientific discovery with experimental feedback~\cite{korshunova2022generative,uehara2024understanding}.

Despite the wide applicability in fine-tuning generative models, difficulties in training of RL algorithms have been observed in both theoretical and empirical studies. In practice, deep RL methods often suffer from high variance, training instability, and sample inefficiency~\cite{islam2017reproducibility,azizzadenesheli2018surprising}. In theory, RL with value function approximation typically requires unrealistic structural assumptions on the underlying process~\cite{jin2020provably} and/or strong coverage assumptions~\cite{xie2021batch}, which are known to be information-theoretically unavoidable~\cite{wang2021exponential,weisz2021exponential}. By way of contrast, supervised learning, such as regression problems, often enjoy strong statistical guarantees and computational efficiency, characterized in the form of oracle inequalities~\cite{bartlett2005local,koltchinskii2011oracle}. Given a norm $\vecnorm{\cdot}{}$ and a function class $\mathcal{F}$, an oracle inequality takes the form
\begin{align}
  \vecnorm{\widehat{f}_n - f^*}{} \lesssim \inf_{f \in \mathcal{F}} \vecnorm{f - f^*}{} + \text{complexity}_\numobs (\mathcal{F}),\label{eq:oracle-ineq}
\end{align}
where $f^*$ is the target function and $\widehat{f}_\numobs$ is the learned function based on $\numobs$ samples. The complexity term measures the capacity of the function class $\mathcal{F}$, and is typically characterized by the covering number, VC dimension, or the Rademacher complexity. For supervised learning problems, oracle inequality~\eqref{eq:oracle-ineq} is known to hold under mild assumptions on the function class, achievable by efficient computational methods (see e.g.~\cite{koltchinskii2011oracle}).

This distinction motivates the key question for RL fine-tuning:
\begin{quote}
 \emph{ Can we achieve the oracle inequality~\eqref{eq:oracle-ineq} for RL fine-tuning of deep generative models, under realistic data assumptions?}
\end{quote}
In this paper, we answer this question affirmatively for diffusion generative models~\cite{ho2020denoising,song2020score}, the state-of-the-art model for image generation. We propose and analyze a new class of algorithms based on the Hamilton-Jacobi-Bellman (HJB) equations satisfied by the optimal value functions. The main contributions of this paper are summarized as follows:
\begin{itemize}
  \item For optimal value function estimation in diffusion fine-tuning, we propose a variational problem, the solution to which satisfies the oracle inequality~\eqref{eq:oracle-ineq}. This oracle inequality further provides sub-optimality gap guarantees for the derived policy, balancing approximation and statistical errors.
  \item Through an instance-dependent statistical error analysis, we further show that the complexity term in \Cref{eq:oracle-ineq} satisfies a self-mitigating property -- the effective noise level in defining such a complexity can be related to the approximation error itself. This leads to a non-standard trade-off between approximation and statistical errors, achieving faster rate than supervised learning.
  \item We also develop an efficient computational method to solve the variational problem, by iteratively solving a sequence of least-square regression problems. The algorithm enjoys exponential fast convergence to a local neighborhood governed by the statistical rates.
\end{itemize}
The rest of this paper is organized as follows: we first introduce notations and discuss related work. In \Cref{sec:preliminaries}, we set up the problem and discuss some examples. We present the main results in \Cref{sec:main} and prove the results in \Cref{sec:proofs}. In Section~\ref{sec:discussion}, we conclude the paper with a discussion of future research directions.

\paragraph{Notations:} Here we summarize some notation used throughout the paper. We use $(B_t: t \geq 0)$ to denote $d$-dimensional standard Brownian motion. For a
positive integer $m$, we define the set $[m] \defn \{1,2, \cdots,
m\}$. We slightly abuse the notation and use $\eucnorm{\cdot}$ to denote absolute value for scalars, Euclidean norm for vectors, and Frobenius norm for matrices.

We use $\partial_x f$ to denote the partial derivative $\tfrac{\partial f}{\partial x}$. Given a multi-index $\alpha \in \mathbb{N}^\usedim$, we define the partial derivative $\partial^\alpha f  \mydefn \frac{\partial^{\alpha_1  + \cdots + \alpha_\usedim} }{\partial x_1^{\alpha_1} \partial x_2^{\alpha_2} \cdots \partial x_d^{\alpha_d}} f$. For the time variable $t$. When it is clear from context, we use the nabla operator $\nabla$ as a vector operator, and denote $\nabla \cdot h \mydefn \textstyle\sum_{i = 1}^\usedim \partial_{x_i} h$ for vector field $h$, and $\nabla^2 \cdot \Sigma \mydefn \sum_{i, j \in [\usedim]} \partial_{x_i} \partial_{x_j} \Sigma_{i, j}$ for matrix-valued function $\Sigma$.

Given a function $f$, a density $\density$ and scalars $k \in \mathbb{N}, q \geq 1$, we define the Sobolev norm
\begin{align*}
  \sobokpnorm{f}{k}{q}{\density} \mydefn \Big\{ \int_{\real^\usedim} \sum_{\bm{1}^\top \alpha \leq k} |\partial^\alpha f (x)|^q \density (x) dx \Big\}^{1/q},
\end{align*}
with the shorthand notations in special cases
\begin{align*}
  \sobonorm{f}{\density} \mydefn \sobokpnorm{f}{1}{2}{\density}, \quad \mbox{and} \quad \lpnorm{f}{q}{\density} \mydefn \sobokpnorm{f}{0}{q}{\density}.
\end{align*}
The Sobolev norm also defines an inner product structure
\begin{align*}
  \soboinprod{f}{g}{\density} \mydefn \int_{\real^\usedim} \big(f(x) g(x) + \nabla f (x)^\top \nabla g (x) \big) \density(x) dx.
\end{align*}
Given a center $x$, a radius $r > 0$, and a norm $\vecnorm{\cdot}{}$, we define the ball $\ball_{\vecnorm{\cdot}{}}(x; r ) \mydefn \{y: \vecnorm{y - x}{} \leq r\}$. When it is clear from context, we omit the norm and/or the center, and simply write $\ball (x; r)$ or $\ball(r)$. We define generic chaining and Dudley chaining functionals as
\begin{align*}
  \dudley_q (\mathcal{H}; \|\cdot\|) &\mydefn \int_0^{+\infty} \log^q N(\varepsilon; \mathcal{H}, \|\cdot\|) \, d\varepsilon, \\
  \gamma_q (\mathcal{H}; \|\cdot\|) &\mydefn \inf_{\{\mathcal{A}_k\}} \sup_{h \in \mathcal{H}} \sum_{k=0}^\infty 2^{k/q} \inf_{a \in \mathcal{A}_k} \|h - a\|,
\end{align*}
where $N (\varepsilon; \mathcal{H}, \|\cdot\|)$ is the covering number of $\mathcal{H}$ at scale $\varepsilon$ under norm $\|\cdot\|$, and $\{\mathcal{A}_k\}$ is a sequence of finite subsets of $\mathcal{H}$ with cardinality at most $2^{2^k}$.

Finally, we define the following distances and divergences between probability measures
\begin{align*}
  \totalvariation(\mu, \nu) \mydefn \frac{1}{2} \int_\Omega \big| d\mu - d\nu \big|, \qquad \mbox{and} \qquad
  \kull{\mu}{\nu} \mydefn \int_\Omega \log \left( \frac{d\mu}{d\nu} \right) d\mu.
\end{align*}
where $d\mu/d\nu$ denotes the Radon--Nikodym derivative of $\mu$ with respect to $\nu$ (assuming $\mu \ll \nu$).

\subsection{Related work}
Our paper connects various lines of research, including reinforcement learning, diffusion models, and AI-driven PDEs solvers. Let us briefly discuss some related works in each area.

\paragraph{Continuous-time RL:} The study of reinforcement learning in continuous-time Markov diffusions dates back to last century~\cite{munos1997reinforcement}, which studies finite-difference numerical schemes for HJB equations. A recent line of work~\cite{jia2022policyeval,jia2022policygrad,jia2023q} derived model-free RL algorithms for controlled Markov diffusions, characterizing them using martingale orthogonality conditions. Our derivation of population-level orthogonality condition is similar to theirs, while we consider the HJB equations under an exponential transform, instead of the original ones. Some theoretical analysis in this paper is built upon our own prior work~\cite{mou2024bellman,mou2025statistical}, where uniform ellipticity plays a central role in exhibiting the geometric properties of Bellman operators, and the self-mitigating error properties.

Our paper uses an exponential transformation on the HJB equation, which is closely related to path integral formulation of RL~\cite{kappen2005path,todorov2006linearly,theodorou2010generalized} in literature. On the other hand, we use function approximation to solve the HJB equation directly, instead of resorting to the importance sampling methods.

\paragraph{Post-training RL in generative models:}the fine-tuning objective can be cast as a regularized RL problem, often using KL or entropy regularization to maintain sample quality~\cite{ziegler2019fine,ouyang2022training,black2023training}. This approach enables reward-driven optimization while constraining the learned policy to stay close to the original model.

Motivated by this problem, recent years have seen new developments in RL theory specialized to fine-tuning tasks. For auto-regressive language models, \cite{zhou2025q,brantley2025accelerating} developed online RL algorithms with function approximations. \cite{yuan2025trajectory} established theoretical guarantees for offline RL methods for post-training optimization in LLMs.

Our work utilize the drift control fine-tuning framework for diffusion models by a recent line of developments~\cite{uehara2024fine,tang2024fine,zhao2025score}. Existing work has primarily focused on extending existing continuous-time RL algorithms to this setting~\cite{gao2024reward,zhao2025score}. As discussed in the introduction, when function approximation is used, these method may suffer from common limitations of RL methods, from both theoretical and practical viewpoints. For existing theoretical studies, \cite{han2024stochastic} proved a global convergence results for a population-level policy iteration algorithm, but did not address function approximations.

Additionally, the fine-tuning objective we study is closely related to classifier guidance in diffusion models~\cite{dhariwal2021diffusion}, which was theoretically analyzed in \cite{tang2024stochastic}. We will explore this connection further in our analysis.

\paragraph{Machine learning for PDEs:}
Recent advances in machine learning have led to new methods for solving partial differential equations (PDEs), including the Hamilton--Jacobi--Bellman (HJB) equations considered in our paper. Two notable approaches are the deep Ritz method (DRM)~\cite{yu2018deep} and the deep Galerkin method (DGM)~\cite{sirignano2018dgm}. Existing theoretical analysis~\cite{lu2021priori,lu2021machine} mainly focused on Poisson equations and Schr\"{o}dinger equations, and derived statistical guarantees for these methods under standard smoothness assumptions. We will discuss connection and difference between these methods and our approach in the following sections.

Several existing works explored learning-based numerical methods for HJB equations considered in this paper. The paper~\cite{zhou2021actor} investigated RL algorithms for high-dimensional solving HJB equations. Focusing on the control-affine setting, the work~\cite{grohs2021deep} provides approximation-theoretic results for neural networks. To the best of our knowledge, statistical guarantees for solving HJB equations (even specialized to the diffusion fine-tuning problem) are still lacking.

\section{Preliminaries}\label{sec:preliminaries}
To start with, let us consider an uncontrolled Markov diffusion process in $\real^\usedim$.
\begin{align}
  d \MyState_t = \drift_t (\MyState_t) dt + \covMat_t (\MyState_t)^{1/2} d \BM_t, \quad t \in [0, T], \qquad \MyState_0 \sim p_0.\label{eq:uncontrolled-process}
\end{align}
where $\drift_t : \real^\usedim \rightarrow \real^\usedim, \covMat_t : \real^\usedim \rightarrow \mathbb{S}_+^{\usedim \times \usedim}$ are time-inhomogeneous drift and diffusion functions, the process $(\BM_t: t \geq 0)$ is a standard $\usedim$-dimensional Brownian motion. We denote the diffusion generator
\begin{align*}
  \generator_t f (x) \mydefn \inprod{\drift_t (x)}{\nabla f (x)} + \frac{1}{2} \mathrm{Tr} (\covMat_t (x) \nabla^2 f (x)).
\end{align*}

In many applications including diffusion model fine-tuning, the control operates additively on the drift term, without changing the diffusion term. Given a control policy $\policy_t: \real^\usedim \rightarrow \real^\usedim$, we consider the controlled process under policy $\policy$.
\begin{align}
  d \MyState_t^\policy = \big( \drift_t (\MyState_t^\policy) + \policy_t (\MyState^\policy) \big) dt + \covMat_t (\MyState_t^\policy)^{1/2} d \BM_t, \quad t \in [0, T],\qquad \MyState_0 \sim p_0.\label{eq:controlled-process}
\end{align}
In the fine-tuning problem, we consider an intermediate reward function $\reward_t: \real^\usedim \rightarrow \real$, and a terminal reward function $\termReward: \real^\usedim \rightarrow \real$. The goal is to maximize the total reward, while keeping the path measure of the controlled process~\eqref{eq:controlled-process} close to the uncontrolled one~\eqref{eq:uncontrolled-process}. Let $\sfactor$ be a regularization parameter, the objective function takes the form
\begin{align}
  J (\policy) \mydefn \Exs \big[ \termReward (\MyState_T^\policy) \big] +  \int_0^T \Exs \big[ \reward_t (\MyState_t^\policy) \big] dt -  \sfactor \kull{\Prob_{[0, T]}^\policy}{\Prob_{[0, T]}},\label{eq:objective-function}
\end{align}
where $\Prob_{[0, T]}$ and $\Prob_{[0, T]}^\policy$ denote the path measures of the process~\eqref{eq:uncontrolled-process} and~\eqref{eq:controlled-process}, respectively. The goal is to approximate the optimal policy.
\begin{align}
  \policy^* \mydefn \arg\max_{\policy} J (\policy), \qquad \mbox{s.t. \Cref{eq:controlled-process}}.\label{eq:fine-tuning-problem-formulation}
\end{align}
Following standard derivation based on Girsanov theorem, the pathwise KL divergence can be represented as the integral of second moment of control action, so that the objective function~\eqref{eq:objective-function} simplifies to
\begin{align*}
  J (\policy) \mydefn \Exs \big[ \termReward (\MyState_T^\policy) \big] +  \int_0^T \Exs \big[ \reward_t (\MyState_t^\policy) \big] dt - \frac{\sfactor}{2} \int_0^T \Exs \big[ \policy_t (\MyState_t^\policy)^\top \covMat_t (\MyState_t^\policy)^{-1} \policy_t (\MyState_t^\policy)  \big] dt.
\end{align*}
In other words, the control applies affinely on the diffusion process, and the action-dependent reward term is quadratic. This connects the fine-tuning problem~\eqref{eq:fine-tuning-problem-formulation} to the classical control-affine systems~\cite{kappen2005path}.

\subsection{Hamilton--Jacobi--Bellman (HJB) equations}
Playing a key role in RL and stochastic control problems is the value function. For $t \in [0, T]$ and $x \in \real^\usedim$, we define
\begin{align}
  \ValTrue_t (\state) \mydefn \max_{\policy} \Exs^\policy \Big[ \termReward (\MyState_T) +  \int_t^T \Big\{ \reward_s (\MyState_s) - \frac{\sfactor}{2} \eucnorm{\covMat_t (\MyState_t^\policy)^{-1/2} \policy_t (\MyState_t^\policy)}^2 \Big\} ds \mid \MyState_t = \state \Big].\label{eq:defn-optimal-value-func}
\end{align}
It is known that $\ValTrue$ solves the HJB equation on $\real^\usedim \times [0, T]$.
\begin{align}
  \frac{\partial \ValTrue_t}{\partial t} + \frac{1}{2} \mathrm{Tr} (\covMat_t \nabla^2 \ValTrue_t) + \inprod{\drift_t}{\nabla \ValTrue_t} + \reward_t + \max_{\action} \Big\{ \inprod{\action}{\nabla \ValTrue_t} - \frac{1}{2} \sfactor a^\top \covMat_t^{-1} a \Big\} = 0\label{eq:defn-hjb-eq}
\end{align}
Note that the maximization problem in \Cref{eq:defn-hjb-eq} is quadratic in $\action$, which admits a closed-form solution. Solving this problem, we arrive at the equation
\begin{align}
  \frac{\partial \ValTrue_t}{\partial t} +  \frac{1}{2} \mathrm{Tr} (\covMat_t \nabla^2 \ValTrue_t) + \inprod{\drift_t }{\nabla \ValTrue_t} + \reward_t + \frac{\sfactor}{2} \big(\nabla \ValTrue_t \big)^\top \covMat_t \big( \nabla \ValTrue_t \big) = 0.\label{eq:hjb-eq-under-control-affine-form}
\end{align}
Though \Cref{eq:hjb-eq-under-control-affine-form} is non-linear, a key observation is that it can be turned into a linear PDE, under the Cole--Hopf exponential transformation~\cite{cole1951quasi,hopf1950partial}.
\begin{align*}
  \ExpVal_t (\state) \mydefn \exp \Big(\frac{ \ValFun_t (\state) }{\sfactor} \Big).
\end{align*}
This transformation originates from Madelung~\cite{madelung1927quantetheorie} by relating Hamilton--Jacobi equation to the Schr\"{o}dinger equation, and was first introduced to control theory by Fleming~\cite{fleming1977exit}.

Under such a transformation, \Cref{eq:hjb-eq-under-control-affine-form} becomes
\begin{align}
  \frac{\partial \ExpValTrue_t}{\partial t} + \frac{1}{2} \mathrm{Tr} (\covMat_t \nabla^2 \ExpValTrue_t) + \inprod{\drift_t }{\nabla \ExpValTrue_t} + \sfactor 
  \reward_t \ExpValTrue_t = 0.\label{eq:linear-pde-for-exp-val-true}
\end{align}
The boundary condition is given at the terminal time $\ExpValTrue_T = \exp \big(\termReward / \sfactor)$. When we perform backward induction, the time-reversed process $(\ExpValTrue_{T - t})_{0 \leq t \leq T}$ is a linear parabolic equation.

\paragraph{Function spaces:} To set up the differential equations properly, we need to define appropriate function spaces. For a function $f: [0, T] \times \real^\usedim \rightarrow \real$, scalar $p \geq 1$, and integer $k \geq 0$, we define
\begin{subequations}
\begin{align}
  \energynorm{f}{k}{p}{T}^p &\mydefn \int_0^T \sobokpnorm{f_t}{k}{p}{\density_t}^p dt + \Exs \big[ \abss{f (X_0)}^p \big] + \Exs \big[ \abss{f (X_T)}^p \big], \qquad \mbox{and}\\
  \dynanorm{f}{k}{p}{T}^p &\mydefn \energynorm{f}{k}{p}{T}^p + \int_0^T \lpnorm{\partial_t f_t}{p}{\density_t}^p dt.
\end{align}
\end{subequations}
In other words, the norm $\energynorm{\cdot}{k}{p}{T}$ corresponds to the $(k, p)$-Sobolev energy functional with boundary norm terms; the stronger norm $\dynanorm{\cdot}{k}{p}{T}$ involves an additional velocity term, requiring temporal regularity of the function. For notation simplicity, we omit the supscripts in the first-order Hilbertian case, and define $\strucnorm{f}{T} \mydefn \energynorm{f}{1}{2}{T}$, $\strongnorm{f}{T} \mydefn \dynanorm{f}{1}{2}{T}$. Note that these norms also induce inner product structures.
\begin{subequations}
\begin{align}
  \strucinprod{f}{g}{T} &\mydefn 
 \int_0^T \soboinprod{f_t}{g_t}{\density_t} dt +   \Exs \big[ f_0 (\State_0) g_0 (\State_0) \big] + \Exs \big[ f_T (\State_T) g_T (\State_T) \big], \quad \mbox{and}\\
  \stronginprod{f}{g}{T} &\mydefn \strucinprod{f}{g}{T} + \int_0^T \Exs \big[ \partial_t f_t (\State_t) \cdot \partial_t g_t (\State_t) \big] dt.
\end{align}
\end{subequations}
Additionally, for any integer $k \geq 0$, we define the uniform norms
\begin{align*}
  \vecnorm{f}{\holderspace^k} \mydefn \sup_{t \in [0, T]} \sup_{x \in \real^\usedim} \Big\{ |\partial_t f_t (x)| + \sum_{j = 0}^k \eucnorm{\nabla^j f_t (x)} \Big\}.
\end{align*}

\subsection{Observational model}\label{subsec:observation-model}
For the fine-tuning problem, the coefficients $(\drift, \covMat)$ of the underlying diffusion process~\eqref{eq:uncontrolled-process} are known to the learning algorithm. This assumption naturally holds true, as the coefficients come from a pre-trained model.

We consider an off-policy setting, where the learning algorithm has access to multiple independent copies of discretely-observed trajectories of the diffusion~\eqref{eq:uncontrolled-process}. Concretely, for each $i \in 1,2,\cdots, \numobs$, we let $(\State_t^{(i)})_{t \geq 0}$ be $\mathrm{i.i.d.}$ copies of \eqref{eq:uncontrolled-process}. Independent of the process itself, we additionally sample random observation times $\smpltime{1}{i}, \smpltime{2}{i}, \cdots, \smpltime{K}{i} \simiid \mathrm{Unif} ([0, T])$. The learner is able to observe snapshots of the processes and instantaneous rewards at these observation times, as well as the initial and terminal states. For each process, we observe
\begin{align*}
  \State_0^{(i)}, ~\big( \State_{\smpltime{k}{i}}^{(i)}, \randreward_{\smpltime{k}{i}}^{(i)} \big)_{k = 1}^K, ~ \State_T^{(i)}, ~\mbox{and}~\termRewardRand_i.
\end{align*}
Our framework allow random unbiased observations for the intermediate rewards
\begin{align*}
  \Exs \Big[ \randreward_{\smpltime{k}{i}}^{(i)} \mid (\State_t^{(i)})_{t \geq 0}, \smpltime{k}{i} \Big] = \reward \big( \State_{\smpltime{k}{i}}^{(i)} \big).
\end{align*}
However, for the terminal reward, we assume an exact oracle
\begin{align*}
  \termRewardRand_i = \termReward (\State_T^{(i)}).
\end{align*}
It is easy to see from our proofs that this exact observation assumption can be relaxed when we have an unbiased observation oracle for the terminal quantity $e^{y (\State_T) / \sfactor}$.

Finally, we assume that the reward observations satisfy the bounds
\begin{align}
|\randreward_{\smpltime{k}{i}}^{(i)} | \in [- \rmax - 1, - 1],\quad \mbox{ and } \quad |\termRewardRand_i | \leq 1, \qquad \mbox{almost surely}.\label{eq:bounded-reward-obs}
\end{align}
Note that these assumptions can be made without loss of generality as long as the rewards are uniformly bounded. In particular, if we have $\max(|\randreward_t|, |\termRewardRand| ) \leq B$ almost surely, we can define the new reward functions $\widetilde{\reward} = \frac{\rmax}{2} \big(\frac{\reward}{B} - 1 \big) - 1$ and $\widetilde{y} = \frac{y}{2B}$, which satisfy \Cref{eq:bounded-reward-obs}. By re-scaling the regularization parameter $\sfactor$ accordingly, the optimal policy under the new reward function is exactly the same as the old one. For simplicity of presentation, we assume that~\Cref{eq:bounded-reward-obs} is in place throughout the paper.

\subsection{Some illustrative examples}
Having set up the framework, let us look at some illustrative examples. We will first discuss how our framework fits into fine-tuning problems of diffusion-based generative models. Then we will extend our scope to control-affine policy fine-tuning in robotics, illustrating the applicability of our methodology beyond generative modelling.
\begin{example}[Reward-guided fine-tuning of diffusion generative models]\label{example:reward-guided-fine-tuning}\upshape
  A diffusion generative model learns data's underlying distribution through a forward-backward procedure. In the forward process, we start from the data's distribution, and run a process
  \begin{align*}
    d Z_t = h_t (Z_t) dt + G_t^{1/2} dB_t, ~ Z_0 \sim p_{\mathrm{data}} \qquad t \in [0, T].
  \end{align*}
  Under suitable choice of the functions $(h, G)$, the forward process converges sufficiently fast to some $p_\infty$. We can then approximately sample from $p_{\mathrm{data}}$ by running the backward process
  \begin{align*}
    d \overline{Z}_t = \big( - h_{T - t} (\overline{Z}_t) + 2 G_{T - t} \nabla \log p_{T - t} (\overline{Z}_t) \big) dt + G_{T - t}^{1/2} dB_t, \qquad t \in [0, T],
  \end{align*}
  where $p_{T - t}$ denotes the density of the forward process $(Z_t)_{t \in[0, T]}$ at time $T - t$. In the fine-tuning problem, the uncontrolled process~\eqref{eq:uncontrolled-process} is given by $(\overline{Z}_t: 0 \leq t \leq T)$. Following existing works~\cite{uehara2024fine,tang2024fine,han2024stochastic}, we consider control actions that apply to the drift term of the process, leading to the controlled process~\eqref{eq:controlled-process}.

  The reward signal is determined by the final state $\overline{Z}_T$. Depending on the nature of the reward signal $\Omega_i \in [-1, 1]$, there are two ways of incorporating it into our framework.
  \begin{itemize}
    \item When the reward is noiseless, i.e., a deterministic function of $\overline{Z}_T$, we can directly use it as the terminal reward $\termRewardRand_i = \Omega_i$. We set the intermediate reward $\reward_t = -1$ for $t \in[0, T]$ to be consistent with our problem setup.
    \item When we can only get unbiased observation of the reward, we can choose sufficiently small $\varepsilon > 0$, and compute the unbiased reward oracle on $\MyState_t$ for $t \in [T - \varepsilon, T]$. The intermediate reward can then be given by
    \begin{align*}
    \randreward_t^{(i)}  = \begin{cases} - 1 + \frac{\Omega_i (\MyState_t) - 1}{2 \varepsilon}, & t \in [T - \varepsilon, T],\\
      -1, & t \in [0, T - \varepsilon].
    \end{cases}
    \end{align*}
    Note that the stochastic reward oracle is applied to an intermediate state $\MyState_t$ instead of the final state $\MyState_T$. The resulting reward model is inexact. On the other hand, by choosing $\varepsilon$ sufficiently small, for Lipschitz reward functions, the intermediate reward can serve as a good approximation to the true reward.
  \hfill \goodendex
  \end{itemize}
\end{example}

\begin{example}[Mirror descent policy optimization] \upshape
   It is worth noting that the optimal control problem~\eqref{eq:fine-tuning-problem-formulation} can be used as a building block of mirror descent algorithms for policy optimization~\cite{kakade2002approximately,lan2023policy}. When applied to continuous-time Markov diffusions, the one-step update of mirror descent takes the form
   \begin{align*}
    \policy^{k + 1} &= \arg \max_{\policy} \Big\{ \Exs \big[ \termReward (\MyState_T^\policy) \big] +  \int_0^T \Exs \big[ \reward_t (\MyState_t^\policy) \big] dt - \sfactor_0 \kull{\Prob_{[0, T]}^\policy}{\Prob_{[0, T]}} - \frac{1}{\stepsize} \kull{\Prob_{[0, T]}^\policy}{\Prob_{[0, T]}^{\policy^k}} \Big\},\\
   s.t. &\quad \mbox{$(X_t^\policy: 0 \leq t \leq T)$ follows the controlled process~\eqref{eq:controlled-process},}
   \end{align*}
   where $\sfactor_0$ is the regularization parameter for the original control problem, and $\stepsize$ is the step size for the mirror descent update. Since the expected reward is a linear functional of the path measure $\Prob_{[0, T]}^\policy$, the convergence of mirror descent algorithm follows from standard results in convex optimization~\cite{nemirovski2009robust}. In~\cite{uehara2024feedback}, a variant of this mirror descent algorithms is used in diffusion model fine-tuning, under the assumption that the one-step update can be solved exactly. However, for continuous space RL problems, the mirror descent update is not straightforward to implement.

   Our framework also provides theoretical guarantees for solving the one-step update of mirror descent in a data-driven manner. In particular, by Girsanov theorem, the optimization objective can be written as
   \begin{align*}
     \Exs \big[ \termReward (\MyState_T^\policy) \big] +  \Exs \int_0^T \Big\{ \reward_t (\MyState_t^\policy) - \frac{\sfactor_0}{2} \eucnorm{ \covMat (\MyState_t^\policy)^{-1/2} \policy_t (\MyState_t^\policy)}^2 - \frac{1}{2\stepsize} \eucnorm{ \covMat (\MyState_t^\policy)^{-1/2} \big( \policy_t - \policy_t^k \big) (\MyState_t^\policy)}^2 \Big\} dt,
   \end{align*}
   which corresponds to the problem~\eqref{eq:fine-tuning-problem-formulation} with the regularization factor $\sfactor = \sfactor_0 + \frac{1}{\stepsize}$, and a new drift function given by $\drift + \frac{1}{1 + \sfactor_0 \stepsize} \policy^k$. We will discuss the implication of our main results to mirror descent in the next section. \hfill \goodendex
\end{example}

\begin{example}[Path-integral control in robotics]\upshape
  The stochastic control problem of the form~\eqref{eq:fine-tuning-problem-formulation} is known as path-integral control in robotics~\cite{kappen2005path,theodorou2010generalized}, where the exponential transformation is widely used. Under a more general formulation, the path-integral control problem considers controlled processes of the form
\begin{align*}
  d \MyState_t^\policy = \big( \drift_t (\MyState_t^\policy) + \ctrlMat_t (\MyState_t^\policy) \policy_t (\MyState_t^\policy) \big) dt + \covMat_t (\MyState_t^\policy)^{1/2} dB_t,
\end{align*}
and the objective function takes the form
\begin{align*}
  J (\policy)  \mydefn \Exs \big[ \termReward (\MyState_T^\policy) \big] +  \int_0^T \Exs \big[ \reward_t (\MyState_t^\policy) \big] dt - \frac{1}{2} \int_0^T \Exs \big[ \policy_t (\MyState_t^\policy)^\top \actCost_t (\MyState_t^\policy) \policy_t (\MyState_t^\policy)  \big] dt.
\end{align*}
Under the classical compatibility condition $\sfactor \ctrlMat_t \actCost_t^{-1} \ctrlMat_t^\top = \covMat_t$, the same exponential transformation leads to the linear PDE~\eqref{eq:linear-pde-for-exp-val-true}. Our analysis directly applies to these control problems. However, compared to the importance sampling algorithms in classical path-integral approaches~\cite{theodorou2010generalized}, we learn the value function more efficiently by directly using the PDE structure, with statistical learning error guarantees. \hfill \goodendex
\end{example}
Additionally, RL in control-affine diffusion processes~\eqref{eq:controlled-process} is known to be useful in several applications, including queuing systems~\cite{ata2025drift} and financial hedging with transaction costs~\cite{muhle2023dynamic}. We leave a detailed exploration of these applications for future work.

\subsection{Benchmark: oracle inequalities in regression problems}
To provide some background, let us briefly describe the form of oracle inequalities we expect in regression problems, which serves as a benchmark for our analysis.

Consider $\mathrm{i.i.d.}$ observations $(X_i, Y_i)_{i = 1}^\numobs$. Define the target function $f^* (x) = \Exs [Y \mid X = x]$. Given a convex set $\Fclass$ of functions, a canonical problem in statistical learning theory is to find an estimator $\widehat{f}_\numobs \in \Fclass$ that satisfies \Cref{eq:oracle-ineq} with the standard $\ltwospace$ norm $\vecnorm{\cdot}{} = \vecnorm{\cdot}{\ltwospace (\Prob_X)}$. To achieve this, a natural choice is the constrained least-squares estimator
\begin{align*}
  \widehat{f}_\numobs = \arg\min_{f \in \Fclass} \frac{1}{\numobs} \sum_{i=1}^\numobs (Y_i - f(X_i))^2.
\end{align*}
In statistical learning theory, the complexity terms in \Cref{eq:oracle-ineq} is characterized by some ``critical radii'' determined by the local metric entropy of the function class $\Fclass$. Let $\sigma^2 \mydefn \sup_{x} \var \big( Y |X = x \big)$ be the noise level, and let $\fbar \mydefn \arg\min_{f \in \Fclass} \vecnorm{f - f^*}{\ltwospace (\Prob_X)}$ be the projection of the target function. We consider a pair of fixed-point equations.
\begin{subequations}
  \label{eq:critical-radii-in-regression}
\begin{align}
  \rho^2 &= \sigma \frac{\dudley_2 \big(\Fclass \cap \ball (\fbar; \rho) \big)}{\sqrt{\numobs}} (+ \mbox{high order terms}), \quad \mbox{and} \label{eq:critical-radii-in-regression-i} \\
   \rho^2 &= \rho \frac{ \dudley_2 \big(\Fclass \cap \ball (\fbar; \rho) \big)}{\sqrt{\numobs}} (+ \mbox{high order terms}). \label{eq:critical-radii-in-regression-ii}
\end{align}
\end{subequations}
Let $\rho_\numobs^* (\sigma)$ and $\rho_\numobs^{\mathrm{cross}} $ be the solution to equations \Cref{eq:critical-radii-in-regression-i} and \Cref{eq:critical-radii-in-regression-ii}, respectively. It is known~\cite{lecue2013learning,mendelson2015learning,Wainwright_nonasymptotic} that the least-squares estimator $\widehat{f}_\numobs$ satisfies
\begin{align}
  \vecnorm{\widehat{f}_\numobs - f^*}{\ltwospace (\Prob_X)} \lesssim \inf_{f \in \Fclass}  \vecnorm{f - f^*}{\ltwospace (\Prob_X)} + \rho_\numobs^* (\sigma) + \rho_\numobs^{\mathrm{cross}} ,\label{eq:oracle-ineq-regression}
\end{align}
with high probability. The term $\rho_\numobs^* (\sigma)$ captures the statistical error due to observation noise, while the term $\rho_\numobs^{\mathrm{cross}}$ corresponds to the estimation error in noiseless settings. They come from the localized empirical process bounds for the main noise and the cross terms, respectively. These rates are also known to be minimax optimal under some regularity assumptions~\cite{lecue2013learning}.

In reinforcement learning, we aim to approximate the optimal value function $\ValTrue$ using a function class $\Fclass$. Compared to standard regression problems, the main challenge is that the value function is not directly observed. Instead, we only have access to noisy observations of the reward signals along sampled trajectories. As a result, oracle inequalities in the form of \Cref{eq:oracle-ineq-regression} are not available in general. However, as we will see in the next section, by exploiting the PDE structure of the fine-tuning problem, we can still establish oracle inequalities that are analogous to \Cref{eq:oracle-ineq-regression}.

\section{Main results}\label{sec:main}
We present the main algorithms and theoretical results in this section. The goal is to approximate the target function $\ExpValTrue$ using a function class $\Fclass$. In doing so, we first introduce a population-level variational condition for the projected solution, and establish its theoretical guarantees. Moving to data-driven algorithms, we then present statistical guarantees as well as efficient computational methods. Finally, we discuss how the value estimation guarantees lead to sub-optimality gap bounds for the optimization problem~\eqref{eq:fine-tuning-problem-formulation}.

\subsection{Population-level variational problem and approximation guarantees}
To start with, we consider the bilinear form defined on a pair $(f, g)$ of functions.
\begin{align}
  B [f, g] \mydefn \Exs \big[ f_T (\MyState_T) g_T (\MyState_T) \big] - \int_0^T  \Exs \Big[ \Big\{ \frac{\partial f_t}{\partial t} + \generator_t f_t + \sfactor \reward_t f_t \Big\} (\MyState_t) \cdot g_t (\MyState_t)  \Big] dt.\label{eq:defn-bilinear-form}
\end{align}
The bilinear form $B[f, g]$ captures the correlation between the functions $f$ and $g$ over space and time, akin to classical weak formulations and energy estimates.

Let $\Fclass$ be a convex set of functions, which is compact under the $\ltwospace$ norm. Given $\Fclass$, we seek to find $\fbar \in \Fclass$, such that
\begin{align}
  B \big[\ExpValTrue - \fbar, g - \fbar \big]\leq 0, \quad \mbox{for any } g \in \Fclass.\label{eq:orthogonality-condition-general}
\end{align}
When the class $\Fclass$ is a linear subspace, equality holds for \Cref{eq:orthogonality-condition-general}, which is similar to the Galerkin orthogonality conditions for elliptic equations. However, our construction deviates from existing analysis of Galerkin methods in parabolic equations, as we impose orthogonality only on the time integral, instead of on the time marginals. Moreover, we relax the equality to inequality, accommodating general approximating classes.

The rest of this subsection is devoted to the error analysis of the population-level variational problem~\eqref{eq:orthogonality-condition-general}. We will establish approximation properties for its solution $\fbar$.

\paragraph{Technical assumptions:} To start with, let us list some assumptions used in our analysis.

First, we assume the underlying diffusion process to be uniformly elliptic.
\myassumption{UE$(\lammin, \lammax)$}{assume:uniform-elliptic}
{
  There exists positive constants $\lammin, \lammax$, such that
  \begin{align*}
    \lammin I_d \preceq \covMat_t (x) \preceq \lammax I_d, \quad \mbox{for any }x \in \real^\usedim, ~ t \in [0, T].
  \end{align*}
}{3cm}
Uniform ellipticity is a standard assumption central to theory of Markov diffusion (see e.g.~\cite{pavliotis2016stochastic}). In the context of diffusion generative modelling, such an assumption is satisfied by most SDE models, but not by ODE models.

Additionally, we require some moment bounds on the drift and diffusion terms, as well as their derivatives.
\myassumption{MB$(\smoothness)$}{assume:moment-bound}
{
  There exists $L > 0$, such that for any $q \geq 1$ such that for any $t \in [0, T]$, we have
  \begin{align*}
    &\Exs \big[ \eucnorm{\drift_t (\State_t)}^q \big] + \Exs \big[ \abss{\nabla \cdot \drift_t (\State_t)}^q \big] \leq  (\pq \smoothness)^{\pq}, \quad \mbox{and} \\
    &\Exs \big[ \eucnorm{\covMat_t (\State_t)}^q \big] + \Exs \big[ \eucnorm{\nabla \cdot \covMat_t (\State_t)}^q \big] +  \Exs \big[ \eucnorm{\nabla^2 \cdot \covMat_t (\State_t)}^q \big] \leq (\pq \smoothness)^{\pq},
  \end{align*}
}{1.7cm}
Assumption~\ref{assume:moment-bound} requires the coefficients in the diffusion process~\eqref{eq:uncontrolled-process}, and their derivatives, to be sub-exponential. For diffusion models, the conditions on the diffusion term are easily satisfied, as we typically work with bounded diffusion matrices that are independent of the state. As for the drift term, the assumption is satisfied when $\drift_t$ and $\nabla \cdot \drift_t$ satisfy linear growth conditions, as the finite-time marginal distribution $\density_t$ is known to be sub-Gaussian (see e.g. \cite{nualart2006malliavin}, Section 2.2).

We also need a regularity condition on the density of the diffusion process.
\myassumption{DR$(\smoothness_{\mathrm{reg}})$}{assume:density-regular}{
  For any $q \geq 1$, there exists $\smoothreg{q} > 0$ such that
  \begin{align*}
    &\max \big( \Exs \big[ \eucnorm{\nabla \log \density_t (\State_t)}^q \big]^{1/q}, \Exs \big[ \eucnorm{\tfrac{\nabla^2\density_t}{\density_t} (\State_t)}^q \big]^{1/q} \big)  \leq \smoothreg{q}, \quad \mbox{for any } t \in [0, T],\\
   & \Exs \big[ \eucnorm{\nabla_{x, y} \log \density_{s, t} (\State_t \mid \State_s)}^q \big]^{1/q}  \leq \frac{1}{\sqrt{t - s}}\smoothreg{q}, \quad \mbox{for } 0 \leq s < t \leq T,\\
    &\Exs \Big[ \eucnorm{\frac{\nabla_{x, y}^2 \density_{s, t}}{\density_{s, t}} (\State_t \mid \State_s)}^q \Big]^{1/q}  \leq \frac{1}{t - s}\smoothreg{q}, \quad \mbox{for } 0 \leq s < t \leq T,
  \end{align*}
}{2cm}
This type of regularity estimates are standard in Malliavin calculus literature~\cite{nualart2006malliavin}. In particular, Assumptions~\ref{assume:uniform-elliptic} and \ref{assume:moment-bound} together imply the regularity estimates in \ref{assume:density-regular}. We refer the readers to the papers~\cite{menozzi2021density,li2024error} for detailed arguments.

Finally, we need a moment comparison condition.
\myassumption{MC$(\tau, \kappa)$}{assume:moment-comparison}{
   For any $f \in \Fclass - \Fclass$, we have
   \begin{align*}
    \energynorm{f}{1}{4 \pq}{T} \leq \tau \strucnorm{f}{T}, \quad \dynanorm{f}{1}{4 \pq}{T} \leq \tau \strongnorm{f}{T}, \quad \mbox{and} \quad \strongnorm{f}{T} \leq \kappa \strucnorm{f}{T}.
   \end{align*}
}{1.9cm}
The first two inequalities are standard in statistical learning theory literature~\cite{mendelson2015learning,lecue2013learning}. The last inequality requires the second moment of the time derivative $\partial_t f$ to be controlled by the second moment of $f$ itself. The following proposition provides a simple sufficient condition for this to hold.
\begin{proposition}\label{prop:bandwidth-limited-func-norm-domination}
  Let $(\phi_j)_{j = 1}^{+ \infty}$ be Legendre polynomials on $[0, T]$. Suppose that there exists constants $m, c_1, c_2 > 0 $, such that a function $f = \sum_{j = 1}^{+ \infty} \alpha_j (x) \phi_j (t)$ satisfies
  \begin{itemize}
    \item $f$ is bandwidth-limited in time, i.e., $\alpha_j \equiv 0$ for $j > m$.
    \item The coefficients satisfy a time-uniform covariance bound, i.e., for any $s, t \in [0, T]$
    \begin{align*}
      c_1 \Exs \big[ \alpha (\MyState_t) \alpha (\MyState_t)^\top \big] \preceq \Exs \big[ \alpha (\MyState_s) \alpha (\MyState_s)^\top \big] \preceq c_2 \Exs \big[ \alpha (\MyState_t) \alpha (\MyState_t)^\top \big].
    \end{align*}
  \end{itemize}
  Then we have $\strongnorm{f}{T} \leq \sqrt{m^3 c_2 / c_1} \strucnorm{f}{T}$.
\end{proposition}
\noindent See \Cref{subsec:app-proof-prop-bandwidth-limited-func-norm-domination} for the proof of this proposition. The bandwidth bound is a natural assumption when we use positional embedding to encode the time variable~\cite{ho2020denoising}. The time-uniform covariance bound requires the second moment of the feature vectors under marginal distributions $\density_t$ not to vary too much over time. This assumption is satisfied by many diffusion processes under natural choice of feature vectors.

\paragraph{Approximation guarantees:} Under above assumptions, we can establish approximation error guarantees for the projected fixed point $\fbar$ defined by \Cref{eq:orthogonality-condition-general}.
\begin{lemma}\label{lemma:approx-error}
  Under the Assumptions~\ref{assume:uniform-elliptic},~\ref{assume:moment-bound},~\ref{assume:density-regular}, and~\ref{assume:moment-comparison}, the solution $\fbar$ to the variational problem~\eqref{eq:orthogonality-condition-general} exists and is unique. Moreover, we have the approximation error bound
  \begin{align*}
    \strucnorm{\fbar - \ExpValTrue}{T} \leq c \frac{  1 + \sfactor \rmax + \tau \smoothness (1 + \smoothreg{8})  }{\min (1, \lammin, \sfactor)} \cdot \inf_{f \in \Fclass} \strongnorm{f - \ExpValTrue}{T},
  \end{align*}
  for a universal constant $c > 0$.
\end{lemma}
\noindent See \Cref{subsec:proof-lemma-approx-error} for the proof of this lemma.

A few remarks are in order. First, we showed that the approximation error of $\fbar$ is comparable to the best approximation error within the class $\Fclass$, albeit under a slightly stronger norm. This is similar to classical C\'{e}a's lemma in finite element methods~\cite{cea1964approximation}, while we extend the analysis to general function class $\Fclass$ and parabolic equations. The norm $\strongnorm{\cdot}{T}$ on the right-hand-side requires the time derivative of the target function $\ExpValTrue$ to be well-approximated by the class $\Fclass$, which is expected, as \Cref{eq:linear-pde-for-exp-val-true} itself involves time derivatives.

Similar to our prior work~\cite{mou2024bellman}, the approximation factor in \Cref{lemma:approx-error} is a problem-dependent constant. Though this constant can be much larger than $1$, it does not depend on the sample size or the discretization step length, which in sharp contrast with the contraction-based bounds in discrete-time RL literature~\cite{mou2023optimal}. This is made possible through the positive-definite nature of the bilinear operator $B$ we constructed.

Finally, we note that \Cref{lemma:approx-error} and the results to follow all rely on convexity of the class $\Fclass$. This assumption is satisfied by many natural models, and enabled by ensemble methods. It is an interesting direction of future work to extend the variational problem and our analysis to non-convex classes, such as neural networks.

\subsection{Data-driven estimation of the value function}
In practice, we cannot directly compute the bilinear form $B$ defined in \Cref{eq:defn-bilinear-form}. As discussed in \Cref{subsec:observation-model}, we can only observe discretely-sampled trajectories of the uncontrolled process~\eqref{eq:uncontrolled-process}, as well as noisy observations of the reward functions. In this subsection, we present a data-driven algorithm for approximating the solution to the variational problem~\eqref{eq:orthogonality-condition-general}, and establish its statistical guarantees.

To start with, we define the empirical counterpart of the bilinear form $B$. Given the observations $\{ \State_0^{(i)}, (\State_{\smpltime{k}{i}}^{(i)}, \randreward_{\smpltime{k}{i}}^{(i)})_{k = 1}^K, \State_T^{(i)}, \termRewardRand_i: i = 1,2,\cdots, \numobs\}$, we define
\begin{multline}
   \widehat{\mathcal{B}}_n (f, g) \mydefn \frac{1}{\numobs} \sum_{i = 1}^\numobs \big( \exp \big( \termReward^{(i)} / \sfactor \big) - f_T (\State_T^{(i)} ) \big) \cdot g_T (\State_T^{(i)})\\
 + \frac{T}{\numobs K} \sum_{i = 1}^\numobs \sum_{k = 1}^K \Big\{\partial_t + \generator_{\smpltime{k}{i}} + \sfactor \randreward_{\smpltime{k}{i}}^{(i)} \Big\} \ExpVal_{\smpltime{k}{i}} (\MyState_{\smpltime{k}{i}}^{(i)})  \cdot g_{\smpltime{k}{i}}  (\MyState_{\smpltime{k}{i}}^{(i)}).\label{eq:defn-empirical-bilinear-form}
\end{multline}
The empirical version of the orthogonality condition~\eqref{eq:orthogonality-condition-general} then takes the form
\begin{align}
  \widehat{\mathcal{B}}_n (\ExpValHat_\numobs, g - \ExpValHat_\numobs) \leq 0 , \quad \mbox{for any $g \in \Fclass$}.\label{eq:optimality-condition-in-finite-sample-variational-problem}
\end{align}
This section focuses on the statistical properties of $\ExpValHat_\numobs$, assuming it exists and can be computed. In the next section, we will present an efficient algorithm that computes an approximate solution that satisfies the same statistical guarantees.

\paragraph{Additional assumptions:}
We additionally impose the following assumptions on the function class $\Fclass$. First, we assume that the following interpolation inequality holds true between the $\energynorm{\cdot}{2}{p}{T}$-norm and the $\strucnorm{\cdot}{T}$-norm, for functions in the class $\Fclass$.
\myassumption{Interp$(\sigma, \eta)$}{assume:subgaussian-class}{
Given a function $f \in \Fclass_*$, for $p \geq 2$, we have
\begin{align*}
 \energynorm{f}{2}{p}{T} \leq \sigma \sqrt{p} \strucnorm{f}{T}^{\eta}, \quad \mbox{and} \quad \dynanorm{f}{2}{p}{T} \leq \sigma \sqrt{p} \strongnorm{f}{T}^{\eta}.
\end{align*}
}{2.5cm}
This assumption is trivially satisfied by bounded classes for $\eta = 0$. As we will see in \Cref{thm:main-error} to follow, a rate of convergence can still be derived even when $\eta = 0$. When $\eta > 0$, this type of assumptions are known as weakly-subGaussian classes~\cite{ziemann2024sharp}, which relaxes the subGaussian class assumption in~\cite{lecue2013learning}.
Under smoothness assumptions on functions in the class $\Fclass$, the Gargliardo--Nirenberg interpolation inequalities imply Assumption~\ref{assume:subgaussian-class}, with the exponent $\eta$ depending on the order of smoothness and the problem dimension. See the paper~\cite{ziemann2024sharp} for more details.

Finally, we need the functions in the class $\Fclass$ and the true solution $\ExpValTrue$ to satisfy a uniform boundedness condition.
\myassumption{BC$(\smoothness_\Fclass)$}{assume:smoothness-class}{
  The functions in the class $\Fclass \cup \{\ExpValTrue\}$ are uniformly bounded by $\smoothness_\Fclass$ in $\vecnorm{\cdot}{\holderspace^2}$, i.e., for any $f \in \Fclass$, we have $\vecnorm{f}{\holderspace^2} \leq \smoothness_\Fclass$.
}{1.7cm}
Using more involved technical arguments, we can also relax this boundedness assumption to weaker tail assumptions.

\paragraph{Critical radii:}
Analogous to the definition~\eqref{eq:critical-radii-in-regression} of critical radii in regression problems, we introduce the following critical radii as solution to certain fixed-point equations. For notational simplicity, let us first define
\begin{align*}
  \shiftConvSet = \Fclass - \Fclass = \{f - g: f,g \in \Fclass \}, \quad \mbox{and} \quad \shiftConvSet (\rho) \mydefn \shiftConvSet \cap \ball (0; \rho, \strucnorm{\cdot}{T}).
\end{align*}
For given $\numobs, K \in \mathbb{N}^+$, $\pq > 1$, and $\delta \in (0, 1)$, we define the following quantities

\begin{subequations}\label{eq:defn-critical-radii}
\begin{carlist}
  \item Given a function $h$, we define the radius $\critradistar_{\numobs, \delta} (h)$ as the largest non-negative real $\rho$ such that
  \begin{multline}
    \rho^2 \leq c_1 \tau \dynanorm{h}{1}{4 \pq}{T} \sqrt{\log K} \Big\{ \frac{ \dudley_2 \big( \shiftConvSet (\rho); \strucnorm{\cdot}{T} \big)}{\sqrt{\numobs}}
     + \rho  \sqrt{\frac{\log (1/ \delta)}{\numobs}}\Big\}  \\
   +   c_1 \frac{\vecnorm{h}{\Ctwospace}\log \numobs }{\numobs} \big( \dudley_1 \big( \convSet; \vecnorm{\cdot}{\holderspace^0} \big) + \smoothness_\Fclass \log (1 / \delta) \big) .\label{eq:defn-critical-radius-star}
  \end{multline}
  \item We define the radius $\critradicross_{\numobs, \delta}$ as the largest non-negative real $\rho$ such that
  \begin{align}
    \rho^2 \leq c_1 \tau^2 \kappa  \rho \sqrt{\log K} \Big\{ \frac{ \dudley_2 \big( \shiftConvSet (\rho); \strucnorm{\cdot}{T} \big)}{\sqrt{\numobs}} + \rho  \sqrt{\frac{ \log (1/ \delta)}{\numobs}}\Big\} 
   +  c_1  \frac{\kappa \sigma \rho^\eta \log \numobs }{\numobs } \big( \dudley_1 \big( \convSet; \vecnorm{\cdot}{\holderspace^1} \big) + \smoothness_\Fclass \log (1 / \delta)\big).\label{eq:defn-critical-radius-cross}
  \end{align}
  \item Finally, we define the radius $\critradismpl_{\numobs, K, \delta}$ as the largest non-negative real $\rho$ such that
  \begin{multline}
    \rho^2 \leq c_1 \frac{\tau^2 \kappa \rho^\eta}{\sqrt{\numobs K}} \Big\{ \dudley_2 (\convSet; \vecnorm{\cdot}{\Ctwospace}) + \smoothness_\Fclass \sqrt{\log (1/ \delta)} \Big\}\\
     + c_1 \smoothness_\Fclass \Big\{  \frac{\dudley_2 (\shiftConvSet (\rho); \strucnorm{\cdot}{T} )  + \rho \sqrt{\log (1 / \delta)}}{\sqrt{\numobs K}} +  \frac{\dudley_1 (\convSet; \vecnorm{\cdot}{\holderspace^0} )  +\smoothness_\Fclass \log (1 / \delta)}{\numobs \sqrt{K}} \log \numobs \Big\}.
  \end{multline}
\end{carlist}
\end{subequations}
The quantities $\critradistar_{\numobs, \delta} (\fbar - \ExpValTrue)$ and $\critradicross_{\numobs, \delta}$ are analogous to $\rho_\numobs^* (\sigma)$ and $\rho_\numobs^{\mathrm{cross}}$ defined in \Cref{eq:critical-radii-in-regression}, respectively. The high-order terms in these definitions are due to the Bernstein-type concentration inequalities we used in our analysis, which is similar to the prior works~\cite{lecue2013learning,ziemann2024sharp}. We use the Sobolev norm $\dynanorm{h}{1}{4 \pq}{T}$ of function $h$ in the place of noise level for the critical radius. In \Cref{thm:main-error} to follow, we will relate this to the approximation error in $\fbar$.

The additional critical radius term $\critradismpl_{\numobs, K, \delta}$ comes from the discrete-time sampling of the continuous-time process. When the number of discrete time steps $K$ is sufficiently large, this term is negligible compared to the two main terms.

\paragraph{Statistical guarantees:} With the above assumptions and definitions in place, we can establish the statistical guarantee for the solution $\ExpValHat_\numobs$ to the variational prokblem~\eqref{eq:optimality-condition-in-finite-sample-variational-problem}.
\begin{theorem}\label{thm:main-error}
  Under the setup of \Cref{lemma:approx-error}, and assuming that Assumptions~\ref{assume:subgaussian-class},~\ref{assume:smoothness-class} are in place. For $\delta \leq \frac{1}{2 \log \numobs + 2 \log K} $, with probability $1 - \delta$, the output $\ExpValHat$ of the empirical variational inequality problem~\eqref{eq:optimality-condition-in-finite-sample-variational-problem} satisfies
 \begin{align*}
  \strucnorm{\ExpValHat_\numobs - \ExpValTrue}{T} \leq c \tau  \inf_{f \in \Fclass} \strongnorm{\ExpValTrue - f}{T} + \critradistar_{\numobs, \delta} (\fbar - \ExpValTrue) + \critradicross_{\numobs, \delta} +  \critradismpl_{\numobs, K, \delta},
 \end{align*}
 for constants $c, c_1 > 0$ depending on the quantities $(\pq, \sfactor, \rmax)$, and problem parameters in Assumptions~\ref{assume:uniform-elliptic},~\ref{assume:moment-bound} and~\ref{assume:density-regular}.
\end{theorem}
\noindent See \Cref{subsec:proof-thm-main-error} for the proof of this theorem.

A few remarks are in order. First, the error bound in \Cref{thm:main-error} is analogous to the oracle inequality~\eqref{eq:oracle-ineq-regression} in regression problems. The first term corresponds to the approximation error, while the other three terms are complexity terms that capture the statistical error due to finite samples and discrete-time observations. Note that in the critical radius $\critradistar_{\numobs, \delta} (\fbar - \ExpValTrue)$, the Sobolev norm $\dynanorm{\fbar - \ExpValTrue}{1}{4 \pq}{T}$ plays the role of noise level. By \Cref{lemma:approx-error}, the $\mathbb{H}^1$-norm of this function is controlled by the approximation error $\inf_{f \in \Fclass} \strongnorm{\ExpValTrue - f}{T}$. In many cases, the $\dynanorm{\cdot}{1}{4 \pq}{T}$-norm of this error term is comparable to the $\strongnorm{\cdot}{T}$-norm, so the critical radius $\critradistar_{\numobs, \delta} (\fbar - \ExpValTrue)$ is dominated by the approximation error term. This self-mitigating error property is similar to the phenomenon observed in~\cite{mou2025statistical}. However, since the uncontrolled process is known in the fine-tuning problem, the bound in \Cref{thm:main-error} does not involve the additional martingale error term due to the unknown transition kernel, which is present in~\cite{mou2025statistical}.

Additionally, we note that the constants $(c, c_1)$ in \Cref{thm:main-error} can hide dependence on problem-dependent parameters such as $(\lammin, \lammax, \smoothness, \smoothreg, \sfactor)$. On the other hand, we track the dependence on the parameters $(\tau, \kappa, \sigma, \eta)$, as they may depend on structures and complexities of the function class $\Fclass$.

\subsubsection{Some concrete exampples}
Let us illustrate the statistical rates in \Cref{thm:main-error} through some concrete examples.

For notational simplicity, we define the gradient classes
\begin{align*}
  \nabla \Fclass \mydefn \big\{ \partial_{x_j} f: f \in \Fclass, j \in [\usedim] \big\}, \quad \mbox{and} \quad \nabla^2 \Fclass \mydefn \big\{ \partial_{x_i} \partial_{x_j} f: f \in \Fclass, i, j \in [\usedim] \big\}.
\end{align*}
Our first example considers parametric function classes, where the log-covering number grows logarithmically with respect to the inverse of the covering radius. This type of behavior is often observed in function classes with finite VC-dimension.
\begin{proposition}\label{prop:example-vc-dimension}
  Suppose that the class $\mathcal{G} \mydefn \Fclass \cup \nabla \Fclass \cup \nabla^2 \Fclass$ satisfies that $\log \mathcal{N}(\varepsilon; \mathcal{G}, \vecnorm{\cdot}{\infty}) \leq d_0 \log(1/\varepsilon)$ for all $\varepsilon $. Under the setup of \Cref{thm:main-error}, for $K \asymp \numobs^2$ and $\delta \asymp \frac{1}{n^2 K^2}$, when $\numobs \gtrsim d_0 \log d_0$, with probability $1 - \delta$, we have
  \begin{align*}
    \strucnorm{\ExpValHat_\numobs - \ExpValTrue}{T} \lesssim \inf_{f \in \Fclass} \strongnorm{\ExpValTrue - f}{T} + \sqrt{\vecnorm{\ExpValTrue - \fbar}{\Ctwospace} \frac{d_0\log \numobs}{\numobs} } + \Big( \frac{d_0 \log \numobs}{\numobs} \Big)^{\frac{1}{2 - \eta}},
  \end{align*}
  where we hide dependence on problem parameters in the $\lesssim$ notation.
\end{proposition}
\noindent See \Cref{subsec:app-proof-prop-example-vc-dimension} for the proof of this proposition. When the approximation error satisfies $\vecnorm{\ExpValTrue - \fbar}{\Ctwospace} = o (1)$, and the weakly subGaussian parameter satisfies $\eta > 0$, the bound in \Cref{prop:example-vc-dimension} is faster than the standard $O (\sqrt{d_0 / \numobs})$ rate in parametric models, due to the self-mitigating error phenomenon discussed above. The choice $K \asymp \numobs^2$ and $\delta \asymp \frac{1}{n^2 K^2}$ are made for notational convenience, and the full dependence on these parameters can be found in the proofs. Note, however, that the first statistical error term in the bound actually comes from ``high-order'' term in the defining equation~\eqref{eq:defn-critical-radius-star} for the critical radius $\critradistar$, which is due to the Bernstein-type concentration inequalities we used in our analysis. With more refined arguments, this term can be improved.

Next, let us move on to non-parametric function classes.

\begin{proposition}\label{prop:example-nonparametric}
  Suppose that there exists $\beta_0, \beta_1 \in (0, 1)$, such that
  \begin{align*}
    \log N (\varepsilon; \Fclass \cup \nabla \Fclass , \vecnorm{\cdot}{\infty}) \leq (c/\varepsilon)^{\beta_0}, \quad \mbox{and} \quad  \log N(\varepsilon; \nabla^2 \Fclass  , \vecnorm{\cdot}{\infty}) \leq (c/\varepsilon)^{\beta_1}.
  \end{align*}
   for all $\varepsilon> 0$. Under the setup of \Cref{thm:main-error}, for $K \asymp \numobs^2$ and $\delta \asymp \frac{1}{n^2 K^2}$, with probability $1 - \delta$, we have
  \begin{align*}
    \strucnorm{\ExpValHat_\numobs - \ExpValTrue}{T} \lesssim \inf_{f \in \Fclass} \strongnorm{\ExpValTrue - f}{T} +\Big(\dynanorm{\ExpValTrue - \fbar}{1}{4 \pq}{T}^2 \frac{\log \numobs}{\numobs} \Big)^{\frac{1}{2 + \beta_0}} + \sqrt{\vecnorm{\ExpValTrue - \fbar}{\Ctwospace}\frac{ \log \numobs}{\numobs}} + \Big( \frac{\log \numobs}{\numobs} \Big)^{\frac{1}{\beta_0} \wedge \frac{1}{2 - \eta}},
  \end{align*}
  where we hide dependence on problem parameters in the $\lesssim$ notation.
\end{proposition}
\noindent See \Cref{subsec:app-proof-prop-example-nonparametric} for the proof of this proposition. Similar to the parametric case, when the approximation error satisfies $\vecnorm{\ExpValTrue - \fbar}{\Ctwospace} = o (1)$, the bound in \Cref{prop:example-nonparametric} is faster than the standard $O (\numobs^{- \frac{1}{2 + \beta}})$ rate in non-parametric models, once again due to the self-mitigating statistical errors. It is also worth noting that the upper bound in \Cref{prop:example-nonparametric} depends only on $\beta_0$ but not on $\beta_1$, as long as $\beta_0, \beta_1 < 1$. In other words, only the metric entropy of the first-order derivative class plays a role in determining the convergence rate. This dependence, on the other hand, is necessary, as we measure the errors in terms of first-order Sobolev norm.

\subsubsection{Comparison to existing methods}

It is useful to compare our algorithm to other existing methods for solving the fine-tuning problem~\eqref{eq:fine-tuning-problem-formulation} and the PDE~\eqref{eq:linear-pde-for-exp-val-true}. We discuss two types of methods below.

\paragraph{Comparison to classifier guidance:} Under the reward-guided diffusion framework in \Cref{example:reward-guided-fine-tuning} (with exact reward observation), the solution to fine-tuning problem~\eqref{eq:fine-tuning-problem-formulation} admits an explicit form
\begin{align*}
  \policy_t^* (x) = \nabla \log \Exs \big[ \exp (\termReward (\State_T) / \sfactor) \mid \State_t = x \big].
\end{align*}
A direct approach, therefore is to learn a time-dependent regression model for the conditional expectation $\Exs [ \exp (\termReward (\State_T) / \sfactor) \mid \State_t = x]$. This approach is first known as classifier guidance in the generative modelling literature~\cite{dhariwal2021diffusion}, as the conditional expectation can be interpreted as the likelihood ratio between the target distribution and the original distribution. However, from statistical learning viewpoint, there are two limitations of the na\"{i}ve classifier guidance method: first, accurate estimation of the conditional expectation does not always imply accurate estimation of its gradient; second, when $t$ is far from $T$, $X_t$ is uninformative to predict $Y$, leading to large noise.

In a recent theoretical study~\cite{tang2024stochastic}, the first challenge is addressed by learning the gradient separately through a martingale approach after learning the classifier guidance. Since their method uses the learned classifier guidance as a plug-in, the high variance is inherited in the second step. By way of contrast, our method addresses the two challenges simultaneously using the stochastic control formulation~\eqref{eq:fine-tuning-problem-formulation}, which not only allows for a broader class of reward observation models, but also provides sharper statistical rates.

\paragraph{Comparison to deep Ritz and deep Galerkin methods:} Our method solves the linearized HJB equation~\eqref{eq:linear-pde-for-exp-val-true} using function approximations. This method is closely related to existing ML approaches for PDEs, such as deep Ritz method~\cite{yu2018deep} and deep Galerkin method~\cite{sirignano2018dgm}. Let us provide a comparison.

The population-level bilinear form $B$ defined in \Cref{eq:defn-bilinear-form} can be interpreted as a weak formulation of the PDE~\eqref{eq:linear-pde-for-exp-val-true}. Through integration-by-parts, it can be related to the loss function in deep Ritz method. However, the deep Ritz method is not applicable to our settings due to two reasons: DRM typically deals with elliptic equations while our equation is parabolic. More importantly, if we apply integration-by-parts formula to \Cref{eq:defn-bilinear-form} to derive the deep Ritz loss, it will involve the gradient of log-density, which does not admit an empirical version. Our approach, on the other hand, solves a variational inequality problem motivated from reinforcement learning, which avoids the need of density estimation.

Compared to deep Galerkin method, we minimize the projected residual in the space of test functions, instead of the norm of the residual. This allows us to operate in the first-order (instead of second-order) Sobolev space, which is more natural for RL fine-tuning problems. As a result, the approximation and statistical errors mainly depend on the geometry of first-order derivatives (see examples above), even if the bilinear form $B$ involves second-order derivatives. This is in sharp contrast with deep Galerkin method, where the approximation and statistical errors depend on the geometry of second-order derivatives.

\subsection{Computing the fixed-point: an iterative algorithm}\label{eq:iterative-optimization}

\Cref{thm:main-error} provides value learning guarantees assuming computational oracles for solving the variational inequality problem~\eqref{eq:optimality-condition-in-finite-sample-variational-problem}. Though the population-level version \eqref{eq:orthogonality-condition-general} is strongly monotonic in suitably defined Sobolev spaces, the finite-sample version~\eqref{eq:optimality-condition-in-finite-sample-variational-problem} is not linear. Therefore, the existence of a solution is not always guaranteed, not to mention efficient computation.

In this section, we propose an iterative algorithm to approximately compute the solution of the variational inequality problem~\eqref{eq:optimality-condition-in-finite-sample-variational-problem}. Our theoretical results will show that the output of the algorithm enjoys the same statistical guarantees as the solution of the variational inequality problem.

To start with, let us first consider an iterative algorithm that solves the population-level problem~\eqref{eq:orthogonality-condition-general}.
\begin{algorithm}[H]
  \caption{Iterative algorithm for solving population-level problem~\eqref{eq:orthogonality-condition-general}}\label{alg:iterative-optimization-population}
  \begin{algorithmic}
    \STATE \textbf{Input:} Initial function $\ExpVal^{(0)} \in \Fclass$, number of iterations $M$, step size $\stepsize > 0$.
    \FOR{$\kiter = 0, 1, \cdots, M - 1$}
      \STATE Update the function
      \begin{align*}
        \ExpVal^{(\kiter + 1)} = \arg\min_{\ExpVal \in \Fclass} \Big\{ \strucnorm{\ExpVal - \ExpVal^{(\kiter)}}{T}^2 - 2 \stepsize B \big[\ExpValTrue - \ExpVal^{(\kiter)}, \ExpVal - \ExpVal^{(\kiter)} \big] \Big\}.
      \end{align*}
    \ENDFOR
    \STATE Output $\ExpVal^{(M)}$.
  \end{algorithmic}
\end{algorithm}
Each iterates of Algorithm~\ref{alg:iterative-optimization-population} can be viewed as a gradient descent step for the variational problem~\eqref{eq:orthogonality-condition-general}. Since the problem geometry aligns with the first-order Sobolev space, we use the $\strucnorm{\cdot}{T}$-norm as the proximal term.

We have the following convergence guarantee for Algorithm~\ref{alg:iterative-optimization-population}.
\begin{proposition}\label{prop:iterative-optimization-population}
  Under the setup of \Cref{lemma:approx-error}, there exists a stepsize threshold $\stepsize_{\max} > 0$ depending on $(\lammax, \lammax, \smoothness, \smoothreg{8}, \tau, \kappa)$, such that when $\stepsize \leq \stepsize_{\max}$, the iterates $\ExpVal^{(M)}$ generated by Algorithm~\ref{alg:iterative-optimization-population} satisfy
  \begin{align*}
    \strucnorm{\ExpVal^{(M)} - \fbar}{T}^2 \leq \exp \Big( - \frac{M \stepsize}{4} \min (\sfactor, \lammin, 1) \Big) \cdot \strucnorm{\ExpVal^{(0)} - \fbar}{T}^2.
  \end{align*}
\end{proposition}
\noindent See \Cref{subsec:proof-iterative-optimization-population} for the proof of this proposition. \Cref{prop:iterative-optimization-population} establishes exponentially fast convergence of the iterates $\ExpVal^{(M)}$ to the target function $\fbar$, which is aligned with classical results on variational inequalities. The monotonicity structure we use lives in a first-order Sobolev space, which is crucial for the analysis.

By replacing the population-level iterates with the empirical version, we can also obtain a data-driven algorithm. The empirical version of Algorithm~\ref{alg:iterative-optimization-population} is given in \Cref{alg:iterative-optimization-empirical}.
To start with, we define the following empirical inner product structure over $n$ training samples.
\begin{align*}
  \widehat{\mathcal{E}}_n (f, g) &\mydefn \frac{1}{\numobs} \sum_{i = 1}^\numobs f_T (\State_T^{(i)}) \cdot g_T (\State_T^{(i)}) + f_0 (\State_0^{(i)}) \cdot g_0 (\State_T^{(i)}) \\
  &\qquad + \frac{T}{\numobs K} \sum_{i = 1}^\numobs \sum_{k = 1}^K \big( f_{\smpltime{k}{i}} (\State_{\smpltime{k}{i}}^{(i)}) \cdot g_{\smpltime{k}{i}} (\State_{\smpltime{k}{i}}^{(i)}) + \nabla f_{\smpltime{k}{i}} (\State_{\smpltime{k}{i}}^{(i)})^\top \nabla g_{\smpltime{k}{i}} (\State_{\smpltime{k}{i}}^{(i)}) \big),
\end{align*}
The functionals $\widehat{\mathcal{E}}_n$ and $\widehat{\mathcal{B}}_n$ are empirical versions of the energy functional and bilinear form used in \Cref{alg:iterative-optimization-population}. Based on these loss functionals, we can solve the empirical version of the variational inequality problem~\eqref{eq:optimality-condition-in-finite-sample-variational-problem} using an iterative algorithm, as described in \Cref{alg:iterative-optimization-empirical}.
\begin{algorithm}[H]
  \caption{Iterative algorithm for solving empirical version of the problem~\eqref{eq:optimality-condition-in-finite-sample-variational-problem}}\label{alg:iterative-optimization-empirical}
  \begin{algorithmic}
    \STATE \textbf{Input:} Initial function $\ExpValHat^{(0)} \in \Fclass$, number of iterations $N$, step size $\stepsize > 0$.
    \FOR{$\kiter = 0, 1, \cdots, M - 1$}
      \STATE Update the function
\begin{align*}
  \ExpValHat^{(\kiter + 1)} = \arg\min_{\ExpVal \in \Fclass} \Big\{ \widehat{\mathcal{E}}_n (\ExpVal - \ExpVal^{(\kiter)}, \ExpVal - \ExpVal^{(\kiter)}) - 2 \stepsize \widehat{\mathcal{B}}_n \big(\ExpVal^{(\kiter)}, \ExpVal - \ExpVal^{(\kiter)} \big) \Big\}.
\end{align*}
    \ENDFOR
    \STATE Output $\ExpValHat^{(M)}$.
  \end{algorithmic}
\end{algorithm}
Similar to \Cref{alg:iterative-optimization-population}, each step of \Cref{alg:iterative-optimization-empirical} involves solving a quadratic minimization problem over a known convex set $\Fclass$, which is a standard subroutine in regression. The theoretical guarantees of the algorithm is given in the following theorem.
\begin{theorem}\label{thm:iterative-optimization-empirical}
  Under the setup of \Cref{thm:main-error}, for any $\delta < \frac{1}{2 \log n + 2 \log K + M^2}$, given a stepsize $\stepsize \in (0, \stepsize_{\max}]$, with probability $1 - \delta$, the iterates $\ExpValHat^{(M)}$ generated by Algorithm~\ref{alg:iterative-optimization-empirical} satisfy
  \begin{multline*}
     \strucnorm{\ExpValHat^{(M)} - \fbar}{T} \leq \exp \Big( - \frac{\min (\sfactor, \lammin, 1)}{8} M \stepsize \Big) \strucnorm{\ExpValHat^{(0)} - \fbar}{T} \\
     + c \Big\{ \critradistar_{\numobs, \delta} (\ExpValTrue - \fbar) + \critradicross_{\numobs, \delta} +  \critradismpl_{\numobs, K, \delta} \Big\} + \frac{c}{ \sqrt{\stepsize}} \critradicross_{\numobs, \delta},
  \end{multline*}
  where the constants $c, c_1 > 0$ depend on the problem parameters in Assumptions~\ref{assume:uniform-elliptic},~\ref{assume:moment-bound}, and~\ref{assume:density-regular}.
\end{theorem}
\noindent See \Cref{subsec:proof-iterative-optimization-empirical} for the proof of this theorem. \Cref{thm:iterative-optimization-empirical} shows that the output of \Cref{alg:iterative-optimization-empirical} enjoys the same statistical guarantees as \Cref{thm:main-error}, up to an exponentially-decaying optimization error term captured by \Cref{prop:iterative-optimization-population}. The stepsize $\stepsize$ can be chosen to balance the optimization and statistical errors. When we choose it to be $\stepsize_{\max}$, the last term in the bound can be absorbed into other statistical error terms. With a constant $\stepsize$, we can choose the number of iterates $M$ growing only logarithmically with the sample size $\numobs$ to ensure that the optimization error is dominated by the statistical error.

\subsection{Sub-optimality gap guarantees}
\Cref{thm:main-error,thm:iterative-optimization-empirical} provide error guarantees for estimating the optimal value function. It remains to turn the estimated value function into a control policy. This section provides guarantees about learning the control policy.

From the derivation of the HJB equation~\eqref{eq:hjb-eq-under-control-affine-form}, the optimal policy is given by
\begin{align*}
  \policy_t^* = \frac{1}{\sfactor} \covMat_t \nabla \ValTrue_t =  \covMat_t \nabla \log \ExpValTrue_t .
\end{align*}
and therefore, given an estimator $\ExpValHat$, we can use the plug-in policy
\begin{align*}
  \widehat{\policy}_t (\state) \mydefn \covMat_t (\state) \nabla \log \ExpValHat_t (\state).
\end{align*}
To measure the performance of a policy $\policy$, we use the sub-optimality gap $J (\policy) - J (\policy^*)$, where $J (\policy)$ is the objective function defined in \Cref{eq:objective-function}. We need the following two additional conditions, which are naturally satisfied by the reward-guided fine-tuning problem in \Cref{example:reward-guided-fine-tuning}.
\begin{itemize}
  \item The diffusion matrix $\covMat_t$ is a function of $t$ only, independent of the state variable.
  \item The process reward $\reward_t$ satisfies $\tfrac{1}{T}\int_0^T \max_x |\reward_t (x)| dt \leq 2$.
\end{itemize}

The following proposition shows that objective function is Lipschitz in terms of the first-order Sobolev norm of the value function.
\begin{proposition}\label{prop:from-valfunc-to-policy}
  Under the setup above, let $\ExpVal^{(1)}$ and $\ExpVal^{(2)}$ be a pair of functions satisfying $\abss{\nabla \ExpVal^{i}_t (x)} \leq \smoothness_\Fclass$ for any $t \in [0, T]$ and $x \in \real^\usedim$. Given any $\varepsilon \in (0, 1)$, the control policy $\policy_t^{(i)} \mydefn \covMat_t \nabla \log \ExpValHat_t^{(i)} $ satisfies
  \begin{align*}
    |J (\policy^{(1)}) - J (\policy^{(2)})| \leq C \sqrt{ \int_0^T \sobopstatnorm{\ExpVal_t^{(1)} - \ExpVal_t^{(2)}}{2 (1 + \varepsilon)}{\density_t}^2 dt},
  \end{align*}
  where $C$ is a constant that depends on the parameters $\sfactor, \varepsilon, \rmax, T$.
\end{proposition}
\noindent See \Cref{subsec:proof-from-valfunc-to-policy} for the proof of this proposition. Applying this proposition with $\ExpVal^{(1)} = \ExpValHat_\numobs$ and $\ExpVal^{(2)} = \ExpValTrue$ gives us a bound on the sub-optimality gap
\begin{align*}
  J (\policy^*) - J (\widehat{\policy}_t)  \leq C \sqrt{ \int_0^T \sobopstatnorm{\ExpValHat_t - \ExpValTrue_t}{2 (1 + \varepsilon)}{\density_t}^2 dt} \leq C (2\smoothness_\Fclass)^{\frac{2}{2 + \varepsilon}} \strucnorm{\ExpValHat_\numobs - \ExpValTrue}{T}^{\frac{2}{2 + \varepsilon}},
\end{align*}
where we use H\"{o}lder's inequality in the last step.
Taking $\varepsilon$ arbitrarily close to $0$, we obtain a convergence rate in the sub-optimality gap that enjoys nearly the same convergence rates in \Cref{thm:main-error,thm:iterative-optimization-empirical}, up to constant factors.

\section{Proofs}\label{sec:proofs}
We collect the proof of main results in this section.
\subsection{Proof of \Cref{lemma:approx-error}}\label{subsec:proof-lemma-approx-error}
We start with the following pair of lemmas, which characterizes the geometric properties of the bi-linear operator $B$.
\begin{lemma}\label{lemma:parabolic-positive-definite}
  Under the setup of \Cref{lemma:approx-error}, for any smooth function $f: [0, T] \times \real^\usedim \rightarrow \real$, we have
  \begin{align*}
   B[f, f] &=  \frac{1}{2}  \Exs \big[ f_0 (\State_0)^2 \big] + \frac{1}{2} \Exs \big[ f_T (\State_T)^2 \big] +  \frac{1}{2} \int_0^T \Exs \Big[ \eucnorm{\covMat_t^{1/2} (\MyState_t) \nabla f_t (\MyState_t)}^2 +2 \sfactor  |\reward_t (\State_t)| f_t^2 (\State_t) \Big] dt\\
    &\geq \frac{\min (\sfactor, \lammin, 1)}{2} \strucnorm{f}{T}^2.
  \end{align*}
\end{lemma}
\noindent See \Cref{subsubsec:lemma-parabolic-positive-definite} for the proof of this lemma.

\begin{lemma}\label{lemma:parabolic-op-upper-bound}
  Under the setup of \Cref{lemma:approx-error}, given a function $g \in \Fclass - \Fclass$ and a smooth functions $f$ defined on $[0, T] \times \real^\usedim$, we have
  \begin{align*}
   \abss{B [f, g]} \leq c \max \Big\{ 1 + \sfactor \rmax, \tau \smoothness (1 + \smoothreg{8})  \Big\} \cdot \strucnorm{g}{T} \cdot \strongnorm{f}{T}. 
  \end{align*}
\end{lemma}
\noindent See \Cref{subsubsec:lemma-parabolic-op-upper-bound} for the proof of this lemma.

Based on these lemmas, let us first establish the existence and uniqueness of the solution to the population-level variational inequality problem~\eqref{eq:orthogonality-condition-general}. To show existence, we let $\fbar_B \mydefn \arg\min_{f \in \Fclass} B [f - \ExpValTrue, f - \ExpValTrue]$. Since $\Fclass$ is compact and \Cref{lemma:parabolic-positive-definite} implies that $B[f, f]$ is non-negative, the minimizer $\fbar_B$ exists. The optimality condition for $\fbar_B$ implies that
\begin{align*}
  B [\ExpValTrue - \fbar_B, f - \fbar_B] \leq 0, \quad \mbox{for any } f \in \Fclass,
\end{align*}
which establishes the existence of $\fbar_B$ as a solution to the population-level variational inequality problem~\eqref{eq:orthogonality-condition-general}.

As for uniqueness, suppose that both $\fbar, \fbar' \in \Fclass$ solve the variational problem~\eqref{eq:orthogonality-condition-general}. We have
\begin{align*}
  B [\ExpValTrue - \fbar, \fbar' - \fbar] \leq 0, \quad \mbox{and} \quad B [\ExpValTrue - \fbar', \fbar - \fbar'] \leq 0.
\end{align*}
Adding them up yields $B [\fbar - \fbar', \fbar - \fbar'] \leq 0$. By \Cref{lemma:parabolic-positive-definite}, we conclude that $\strucnorm{\fbar - \fbar'}{T} = 0$, which establishes the uniqueness of the solution to the variational inequality problem~\eqref{eq:orthogonality-condition-general}.

Now we turn to analyze the approximation error guarantees. Given a function $\ftil \in \Fclass$ fixed, by the optimality condition~\eqref{eq:orthogonality-condition-general}, we have
\begin{align*}
  \ltwoinprod{\ExpValTrue_T - \fbar_T }{\ftil_T - \fbar_T}{\density_T}
  + \int_0^T \Exs \Big[ \Big\{ \frac{\partial \fbar_t}{\partial t} + \generator_t \fbar_t + \sfactor \reward_t \fbar_t \Big\} (\MyState_t) \cdot (\ftil_t - \fbar_t) (\MyState_t)  \Big] dt \leq 0.
\end{align*}
Note that $ \frac{\partial \ExpValTrue_t}{\partial t} + \generator_t \ExpValTrue_t + \sfactor \reward_t \fbar_t = 0$. We can re-write the inequality as
\begin{align*}
  \ltwoinprod{\ExpValTrue_T - \fbar_T}{\ftil_T - \fbar_T}{\density_T}
  + \int_0^T  \ltwoinprod{\Big\{ \frac{\partial }{\partial t} + \generator_t + \sfactor  \reward_t  \Big\} (\fbar_t - \ExpValTrue_t)}{ \ftil_t - \fbar_t }{\density_t} dt \leq 0,
\end{align*}
i.e., $B [\ExpValTrue - \fbar, \ftil - \fbar] \leq 0$. Re-arranging yields
\begin{align}
  B [\ftil - \fbar, \ftil - \fbar] \leq B [\ExpValTrue - \ftil, \ftil - \fbar].\label{eq:basic-ineq-approx-factor-analysis}
\end{align}

Applying \Cref{lemma:parabolic-positive-definite} and \Cref{lemma:parabolic-op-upper-bound} to \Cref{eq:basic-ineq-approx-factor-analysis}, we conclude that
\begin{multline*}
  \frac{\min (\sfactor, \lammin, 1)}{2} \strucnorm{\fbar - \ftil}{T}^2 \leq  B [\ftil - \fbar, \ftil - \fbar] \leq B [\ExpValTrue - \ftil, \ftil - \fbar] \\
  \leq c \max \Big\{ 1 + \sfactor \rmax, \tau \smoothness (1 + \smoothreg{8})  \Big\} \cdot \strucnorm{\ftil - \fbar}{T} \cdot \strongnorm{\ExpValTrue - \ftil}{T}.
\end{multline*}
Note that the function $\ftil$ is arbitrary, and we can take the infimum over all functions in $\Fclass$. Under this choice, re-arranging above inequality yields the conclusion of \Cref{lemma:approx-error}:
\begin{align*}
  \strucnorm{\fbar - \ftil}{T} \leq 2c \frac{ \max \big( 1 + \sfactor \rmax, \tau \smoothness (1 + \smoothreg{8})  \big) }{\min (1, \lammin, \sfactor)} \cdot \inf_{\ftil \in \Fclass} \strongnorm{\ExpValTrue - \ftil}{T}.
\end{align*}

\subsubsection{Proof of \Cref{lemma:parabolic-positive-definite}}\label{subsubsec:lemma-parabolic-positive-definite}
For the time derivative, we note that
\begin{align}
  \Exs \Big[ \frac{\partial f_t}{\partial t} (\State_t) f_t (\State_t) \Big] &= \frac{1}{2} \int \frac{\partial}{\partial t} (f_t^2) (x) \density_t (x) dx \nonumber\\
  &= \frac{1}{2} \frac{d}{dt} \int f_t^2 (x) \density_t (x) dx - \frac{1}{2} \int f_t^2 (x) \frac{\partial \density_t}{\partial t} (x) dx.\label{eq:time-derivative-in-pos-def-proof}
\end{align}
Now we study the generator term. Applying integration-by-parts formula, we have
\begin{align}
  &\Exs \Big[ \big( \frac{1}{2} \mathrm{Tr} (\covMat_t \nabla^2 f_t) + \inprod{\drift_t }{\nabla f_t} \big) (\State_t) f_t (\State_t) \Big] = \int \big( \frac{1}{2} \mathrm{Tr} (\covMat_t \nabla^2 f_t) + \inprod{\drift_t }{\nabla f_t} \big) f_t \density_t dx \nonumber \\
  &= - \frac{1}{2} \int (\nabla f_t)^\top \covMat_t \nabla f_t \density_t dx - \frac{1}{2} \int f_t \nabla f_t^\top \nabla \cdot (\density_t \covMat_t) dx + \frac{1}{2} \int \drift_t^\top \nabla (f_t^2) \density_t dx \nonumber \\
  &= - \frac{1}{2} \int (\nabla f_t)^\top \covMat_t \nabla f_t \density_t dx + \frac{1}{4}  \int f_t^2 \nabla^2 \cdot (\density_t \covMat_t) dx  - \frac{1}{2} \int f_t^2 \nabla \cdot (\density_t \drift_t) dx. \label{eq:generator-basic-expression-in-pos-def-proof}
\end{align}
The density $\density_t$ satisfies the Fokker--Planck equation
\begin{align}
  \frac{\partial \density_t}{\partial t} = - \nabla \cdot (\density_t \drift_t) + \frac{1}{2} \nabla^2 \cdot (\density_t \covMat_t).\label{eq:fokker-planck}
\end{align}
Substituting \Cref{eq:fokker-planck} to \Cref{eq:generator-basic-expression-in-pos-def-proof}, we obtain
\begin{align}
  \Exs \Big[ \big( \frac{1}{2} \mathrm{Tr} (\covMat_t \nabla^2 f_t) + \inprod{\drift_t }{\nabla f_t} \big) (\State_t) f_t (\State_t) \Big]
  = - \frac{1}{2} \ltwonorm{\covMat_t^{1/2} \nabla f_t}{\density_t}^2 + \frac{1}{2} \int f_t^2 \frac{\partial \density_t}{\partial t} dt.\label{eq:organized-generator-bound-in-pos-def-proof}
\end{align}
Combining \Cref{eq:time-derivative-in-pos-def-proof,eq:organized-generator-bound-in-pos-def-proof}, we have the time integral bound
\begin{align*}
  &\int_0^T \Exs \Big[ \big( \frac{\partial f_t}{\partial t} + \frac{1}{2} \mathrm{Tr} (\covMat_t \nabla^2 f_t) + \inprod{\drift_t }{\nabla f_t} + \sfactor \cdot \reward_t f_t \big) (\MyState_t) \cdot f_t (\MyState_t)  \Big] dt\\
  &\leq  - \frac{1}{2} \int_0^T \ltwonorm{\covMat_t^{1/2} \nabla f_t}{\density_t}^2 dt + \int_0^T \sfactor \Exs \big[ \reward_t (\State_t) f_t^2 (\State_t) \big] dt + \frac{1}{2} \Exs \big[ f_T^2 (\State_T) \big] - \frac{1}{2} \Exs \big[ f_0^2 (\State_0) \big].
\end{align*}
Note that $\reward_t (\State_t) < 0$ almost surely by assumption. We have the equation
\begin{align*}
   & \Exs \big[ f_T (\State_T)^2 \big] - \int_0^T \Exs \Big[ \Big\{ \frac{\partial f_t}{\partial t} + \frac{1}{2} \mathrm{Tr} (\covMat_t \nabla^2 f_t) + \inprod{\drift_t }{\nabla f_t} + \sfactor  \reward_t f_t \Big\} (\MyState_t) \cdot f_t (\MyState_t)  \Big] dt \\
   &= \frac{1}{2}  \Exs \big[ f_0 (\State_0)^2 \big] + \frac{1}{2} \Exs \big[ f_T (\State_T)^2 \big] +  \frac{1}{2} \int_0^T \Exs \Big[ \eucnorm{\covMat_t^{1/2} (\MyState_t) \nabla f_t (\MyState_t)}^2 + 2 \sfactor  |\reward_t (\State_t)| f_t^2 (\State_t) \Big] dt.
\end{align*}
Invoking \Cref{assume:uniform-elliptic}, and noting that $\reward_t \leq -1$ almost surely, we conclude that
\begin{align*}
  B [f, f] \geq\frac{1}{2}  \Exs \big[ f_0 (\State_0)^2 \big] + \frac{1}{2} \Exs \big[ f_T (\State_T)^2 \big] + \frac{\min(\lammin, \sfactor)}{2} \int_0^T \sobonorm{f_t}{\density_t}^2 dt,
\end{align*}
which completes the proof of this lemma.

\subsubsection{Proof of \Cref{lemma:parabolic-op-upper-bound}}\label{subsubsec:lemma-parabolic-op-upper-bound}
For the generator term, applying integration-by-parts formula, we note that
\begin{align*}
  &\Exs \big[ (\generator_t f_t) (\State_t) g_t (\State_t) \big]\\
  &= \int \Big\{ \inprod{\drift_t}{\nabla f_t} + \frac{1}{2} \mathrm{Tr} \big( \covMat_t \nabla^2 f_t \big) \Big\} g_t p_t dx\\
  &=- \frac{1}{2} \int \Big(\nabla f_t \covMat_t \nabla g_t \Big) \density_t dx - \frac{1}{2} \int g_t \inprod{\nabla f_t}{\nabla \cdot (\density_t \covMat_t)} dx + \int \inprod{\nabla f_t }{\drift_t} g_t \density_t dx\\
  &= - \frac{1}{2} \int \Big(\nabla f_t \covMat_t \nabla g_t \Big) \density_t dx + \frac{1}{2} \int f_t \nabla \cdot \big(g_t \nabla (\density_t \covMat_t) \big) dx - \int f_t \drift_t^\top \nabla g_t \density_t dx - \int f_t g_t \nabla \cdot (\density_t \drift_t) dx\\
  &= - \frac{1}{2} \int \Big(\nabla f_t \covMat_t \nabla g_t \Big) \density_t dx + \int f_t g_t \generator_t^* p_t dx +  \int f_t \inprod{\nabla g_t}{\frac{1}{2} \nabla \covMat_t + \frac{1}{2} \covMat_t \nabla \log \density_t - \drift_t} \density_t dx.
\end{align*}
Now we bound the terms in this decomposition. By Cauchy--Schwarz inequality, we have
\begin{align*}
  &\abss{\int_0^T \int \Big(\nabla f_t \covMat_t \nabla g_t \Big) \density_t dx dt} \leq \lammax \int_0^T \sobonorm{f_t}{\density_t} \cdot \sobonorm{g_t}{\density_t} dt \leq \lammax \strucnorm{f}{T} \cdot \strucnorm{g}{T}, \qquad \mbox{and}\\
  &\abss{\int_0^T\int f_t g_t \generator_t^* p_t dx dt} \leq \int_0^T \ltwonorm{g_t}{\density_t} \cdot \lpnorm{f_t}{4}{\density_t} \cdot \lpnorm{\frac{\generator_t^* \density_t}{\density_t}}{4}{\density_t} dt\\
  & \qquad \leq \Big( \int \ltwonorm{g_t}{\density_t} dx \Big)^{1/2} \cdot \Big( \int_0^T  \lpnorm{f_t}{4}{\density_t}^2 \cdot \lpnorm{\frac{\generator_t^* \density_t}{\density_t}}{4}{\density_t}^2 dt \Big)^{1/2}
  \leq  \tau \strucnorm{g}{T} \cdot \strucnorm{f}{T} \cdot \sup_{t \in [0, T]} \lpnorm{\frac{\generator_t^* \density_t}{\density_t}}{4}{\density_t}.
\end{align*}
For the last term, we have
\begin{align*}
  &\abss{\int_0^T \Exs \Big[ f_t (\State_t) \cdot \inprod{\nabla g_t}{\frac{1 }{2} \nabla \covMat_t + \frac{1}{2} \covMat_t \nabla \log \density_t -  \drift_t} (\State_t) \Big] dt}\\
  &\leq \int_0^T \ltwonorm{\nabla g_t}{\density_t} \cdot \lpnorm{f_t}{4}{\density_t} \cdot \lpnorm{\frac{1}{2} \nabla \covMat_t + \frac{1}{2} \covMat_t \nabla \log \density_t -  \drift_t}{4}{\density_t} dt \\
  &\leq \tau \strucnorm{g}{T} \cdot \strucnorm{f}{T} \cdot \sup_{t \in [0, T]} \lpnorm{\frac{1}{2} \nabla \covMat_t + \frac{1}{2} \covMat_t \nabla \log \density_t -  \drift_t}{4}{\density_t} .
\end{align*}
By Assumption~\ref{assume:moment-bound} and~\ref{assume:density-regular}, for any $t \in [0, T]$, we have
\begin{align*}
  \lpnorm{\frac{\generator_t^* \density_t}{\density_t}}{4}{\density_t} &\leq \lpnorm{\nabla \cdot \drift_t}{4}{\density_t} + \lpnorm{\drift_t^\top \nabla \log \density_t}{4}{\density_t} + \lpnorm{\tfrac{\nabla^2 \density_t}{\density_t} \cdot \covMat_t}{4}{\density_t} + 2 \lpnorm{\nabla \cdot \covMat_t^\top \nabla \log \density_t}{4}{\density_t} + \lpnorm{\nabla^2 \covMat_t}{4}{\density_t}\\
  &\leq c \smoothness (1 + \smoothreg{8}),
\end{align*}
for a universal constant $c > 0$ independent of other parameters. Similarly, we have
\begin{align*}
   \lpnorm{\frac{1}{2} \nabla \covMat_t + \frac{1}{2} \covMat_t \nabla \log \density_t -  \drift_t}{4}{\density_t} \leq c \smoothness (1 + \smoothreg{8}).
\end{align*}
Putting them together, we have
\begin{align*}
  \abss{\int_0^T \Exs \big[ (\generator_t f_t) (\State_t) g_t (\State_t) \big] dt} \leq c \tau \smoothness (1 + \smoothreg{8}) \strucnorm{g}{T} \cdot \strucnorm{f}{T}.
\end{align*}
For the additional terms in $B[f, g]$, we note that
\begin{align*}
  \abss{\ltwoinprod{f_T}{g_T}{\density_T}} & \leq \strucnorm{f}{T} \cdot \strucnorm{g}{T},\\
  \abss{\int_0^T \Exs \Big[ g_t (\State_t) \big( \partial_t + \sfactor \reward_t \big) f_t (\State_t) \Big] dt} &\leq \strucnorm{g}{T} \cdot \Big\{ \sfactor \rmax \strucnorm{f}{T} + \sqrt{\int_0^T \ltwonorm{\partial_t f_t}{\density_t}^2 dt} \Big\}.
\end{align*}
Collecting these bounds, we conclude that
\begin{align*}
  \abss{B [f, g]} &\leq c \max \Big\{ 1 + \sfactor \rmax, \tau \smoothness (1 + \smoothreg{8})  \Big\} \cdot \strucnorm{g}{T} \cdot \strongnorm{f}{T}. 
\end{align*}
which completes the proof of this lemma.

\subsection{Proof of \Cref{thm:main-error}}\label{subsec:proof-thm-main-error}
To start with, we define the process
\begin{multline}
  \xi_i (f, g) \mydefn f_T (X_T^{(i)}) g_T (X_T^{(i)}) - \frac{T}{K} \sum_{k = 1}^K   \big( \partial_t + \generator_{\smpltime{k}{i}} + \sfactor \Reward_{\smpltime{k}{i}} \big)  f_{\smpltime{k}{i}} (\MyState_{\smpltime{k}{i}}^{(i)})  \cdot  g_{\smpltime{k}{i}} (\MyState_{\smpltime{k}{i}}^{(i)}),\label{eq:defn-xi-process}
\end{multline}
Clearly, we have that $\Exs [\xi_i (f, g)] = B [f, g]$, where $B [f, g]$ is defined in \Cref{eq:defn-bilinear-form}. We bound the pointwise variance of this process in the following lemma.
\begin{lemma}\label{lemma:variance-bound-xi-process}
  For two pairs of functions $f^{(1)}, g^{(1)}, f^{(2)}, g^{(2)}: [0, T] \times \real^\usedim \rightarrow \real$, we have
  \begin{multline*}
    \var \Big( \xi_i (f^{(1)}, g^{(1)}) - \xi_i (f^{(2)}, g^{(2)}) \Big) \\
    \leq  c \log K \Big\{ \dynanorm{g^{(1)} - g^{(2)}}{1}{4 \pq}{T}^2  \energynorm{f^{(1)}}{1}{4 \pq}{T}^2 + \dynanorm{f^{(1)} - f^{(2)}}{1}{4 \pq}{T}^2  \energynorm{g^{(2)}}{1}{4 \pq}{T}^2 \Big\} \\
    +  \frac{c T}{K} \Big\{ \energynorm{g^{(1)} - g^{(2)}}{0}{4 \pq}{T}^2  \dynanorm{f^{(1)}}{2}{4 \pq}{T}^2 + \dynanorm{f^{(1)} - f^{(2)}}{2}{4 \pq}{T}^2  \energynorm{g^{(2)}}{0}{4 \pq}{T}^2 \Big\},
  \end{multline*}
  where the constant $c$ depends on the parameters in Assumptions~\ref{assume:moment-bound} and~\ref{assume:density-regular}, time horizon $T$, and the exponent $\pq$.
\end{lemma}
\noindent See \Cref{subsubsec:proof-lemma-variance-bound-xi-process} for the proof of this lemma.

We also use the following auxiliary process
\begin{align*}
  \zeta_i (h) \mydefn \frac{T}{K} \sum_{k = 1}^K   \big( \partial_t + \generator_{\smpltime{k}{i}} + \sfactor \Reward_{\smpltime{k}{i}} \big)  \ExpValTrue_{\smpltime{k}{i}} (\MyState_{\smpltime{k}{i}}^{(i)})  \cdot  h_{\smpltime{k}{i}} (\MyState_{\smpltime{k}{i}}^{(i)}).
\end{align*}

Now we use the optimality condition~\eqref{eq:optimality-condition-in-finite-sample-variational-problem} to derive the basic inequality for the estimation error. Applying \Cref{eq:optimality-condition-in-finite-sample-variational-problem} with $g = \fbar$, we have
\begin{align*}
  \frac{1}{\numobs} \sum_{i = 1}^\numobs \xi_i (\ExpValTrue - \ExpValHat, \fbar - \ExpValHat)  + \frac{1}{\numobs} \sum_{i = 1}^\numobs \zeta_i (\fbar - \ExpValHat) \leq 0.
\end{align*}

Consider the empirical processes
\begin{align*}
  \errtermgen (\rho) & \mydefn \sup_{f, g\in \shiftConvSet \cap \strucball (\rho)} \frac{1}{\numobs} \sum_{i = 1}^\numobs \Big\{ \xi_i (f, g) - B (f, g) \Big\},\\
  \errtermmain (\rho) &\mydefn \sup_{h \in \shiftConvSet \cap \strucball (\rho)} \frac{1}{\numobs} \sum_{i = 1}^\numobs \Big\{ \xi_i (\ExpValTrue - \fbar, h) - B (\ExpValTrue - \fbar, h) \Big\},\\
  \errtermsmpl (\rho) &\mydefn \sup_{h \in \shiftConvSet \cap \strucball (\rho)} \frac{1}{\numobs} \sum_{i = 1}^\numobs \zeta_i (h) .
\end{align*}
Recall the bilinear form $B [f, g]$ defined in \Cref{eq:defn-bilinear-form}.
By substituting $g = \fbar$ in the optimality condition~\eqref{eq:optimality-condition-in-finite-sample-variational-problem}, we have
\begin{align}
  B \big[ \ExpValHat - \fbar, \ExpValHat - \fbar \big] \leq \errtermmain (\rhohat) + \errtermgen (\rhohat) + \errtermsmpl (\rhohat),\label{eq:basic-inequality-estimation-error}
\end{align}
where we define $\rhohat \mydefn \strucnorm{\ExpValHat - \fbar}{T}$.

By \Cref{lemma:parabolic-positive-definite}, we have the lower bound
\begin{align}
  B \big[ \ExpValHat - \fbar, \ExpValHat - \fbar \big] \geq \frac{\min (\lammin, \sfactor, 1)}{2} \strucnorm{\ExpValHat - \fbar }{T}^2.\label{eq:basic-inequality-lower-bound}
\end{align}
It suffices to bound the error terms. We provide non-asymptotic bounds for the error terms in the following lemmas.

\begin{lemma}\label{lemma:errterm-main-bound}
  For any $\rho > 0$, with probability $1 - \delta$, we have
  \begin{align*}
    \errtermmain (\rho) &\leq c \Big\{ \tau \dynanorm{ \ExpValTrue - \fbar }{1}{4 \pq}{T} \sqrt{\frac{\log K}{\numobs}} + \frac{ \dynanorm{ \ExpValTrue - \fbar }{2}{4 \pq}{T} }{\sqrt{\numobs K}} \Big\} \Big\{ \dudley_2 \big( \shiftConvSet (\rho); \strucnorm{\cdot}{T} \big) + \rho \sqrt{\log (1 / \delta)} \Big\}\\
  &\qquad + c \vecnorm{\ExpValTrue - \fbar}{\Ctwospace} \frac{\log \numobs}{\numobs} \cdot \Big\{ \dudley_1 \big( \convSet; \vecnorm{\cdot}{\holderspace^0} \big) + \smoothness_\Fclass \log (1 / \delta) \Big\}.
  \end{align*}
\end{lemma}
\noindent See \Cref{subsubsec:proof-lemma-errterm-main-bound} for the proof of this lemma.

\begin{lemma}\label{lemma:errterm-gen-bound}
  For any $\rho > 0$, with probability $1 - \delta$, we have
  \begin{multline*}
    \errtermgen (\rho) \leq c \tau^2 \kappa \rho \sqrt{\frac{\log K}{\numobs}} \Big\{ \dudley_2 \big( \shiftConvSet (\rho); \strucnorm{\cdot}{T} \big) + \rho \sqrt{\log (1 / \delta)} \Big\}\\
     +  \frac{c \kappa \sigma \rho^\eta}{\numobs} \Big\{ \dudley_1 \big( \convSet; \vecnorm{\cdot}{\Ctwospace} \big) + \smoothness_\Fclass \log (1 / \delta) \Big\}
  + \frac{\tau^2 \kappa \rho^\eta }{\sqrt{\numobs K}} \Big\{ \dudley_2 (\convSet; \vecnorm{\cdot}{\Ctwospace}) + \smoothness_\Fclass \sqrt{\log (1/ \delta)} \Big\}.
  \end{multline*}
\end{lemma}
\noindent See \Cref{subsubsec:proof-lemma-errterm-gen-bound} for the proof of this lemma.

As for the sampling error term, we have the following lemma.
\begin{lemma}\label{lemma:errterm-sampling-bound}
  For any $\rho > 0$, with probability $1 - \delta$, we have
  \begin{align*}
    \errtermsmpl (\rho) \leq  c \smoothness_\Fclass \sfactor T \Big\{  \frac{\dudley_2 (\shiftConvSet (\rho); \strucnorm{\cdot}{T} )  + \rho \sqrt{\log (1 / \delta)}}{\sqrt{\numobs K}} +  \frac{\dudley_1 (\convSet; \vecnorm{\cdot}{\holderspace^0} )  +\smoothness_\Fclass \log (1/ \delta)}{\numobs \sqrt{K}} \log \numobs \Big\}.
  \end{align*}
\end{lemma}
\noindent See \Cref{subsubsec:proof-lemma-errterm-sampling-bound} for the proof of this lemma.

Note that \Cref{lemma:errterm-main-bound,lemma:errterm-gen-bound,lemma:errterm-sampling-bound} provide upper bounds on localized empirical process suprema for fixed radius $\rho$. However, the basic inequality~\eqref{eq:basic-inequality-estimation-error} involves a random radius $\rhohat = \strucnorm{\ExpValHat - \fbar}{T}$. To handle this, we need the notion of sub-quadratic functions. In particular, we call a non-decreasing function $s: \real_+ \rightarrow \real_+$ sub-quadratic if for any $\rho_1, \rho_2 > 0, a > 1$, we have
\begin{align}
  s (\rho_1 + \rho_2) \leq 2 s (\rho_1) + 2 s(\rho_2), \quad \mbox{and} \quad s (a \rho_1) \leq a^2 s (\rho_1). \label{eq:defn-quasi-sub-additive}
\end{align}
Clearly, the functions $\rho \mapsto \rho$ and $\rho \mapsto \rho^2$ are both sub-quadratic. We use the following lemma to establish the sub-quadratic of some functions involved in our analysis.
\begin{lemma}\label{lemma:quasi-sub-additive}
  Let $\vecnorm{\cdot}{X}, \vecnorm{\cdot}{Y}$ be a pair of norms. For $c \in [0, 1]$ and $\gamma > 0$, the function $h : \real_+ \mapsto \real_+$ defined as
  \begin{align*}
    h (\rho) \mydefn \rho^c \dudley_\gamma \big( \shiftConvSet \cap \ball_X (\rho); \vecnorm{\cdot}{Y} \big)
  \end{align*}
  is sub-quadratic. Furthermore, if $h_1$ and $h_2$ are both sub-quadratic, then for any pair $a_1, a_2 > 0$, the function $a_1 h_1 + a_2 h_2$ is also sub-quadratic.
\end{lemma}
\noindent See \Cref{app:subsec-proof-quasi-subadditive} for the proof of this lemma.

By Lemma~\ref{lemma:quasi-sub-additive}, the right-hand-side of the bounds in \Cref{lemma:errterm-main-bound,lemma:errterm-gen-bound,lemma:errterm-sampling-bound} are all sub-quadratic functions of $\rho$. We can now use the following lemma to bound the estimation error with respect to the random radius $\rhohat$.

\begin{lemma}\label{lemma:random-radius-empirical-process}
  Let $Z: \real^+ \rightarrow \real^*$ be a random non-decreasing function such that for any $\delta \in (0, 1)$,
  \begin{align*}
    \Prob \big( Z (\rho) \leq s_\delta (\rho) \big) \geq 1 - \delta, \quad \mbox{for any } \rho > 0,
  \end{align*}
  for some deterministic function $s$ such that $s_\delta$ is sub-quadratic for any $\delta \in (0, 1)$. Then, for any random variable $\rhohat \in [0, D]$, and deterministic scalar $a > 1$, suppose that $\rhofix$ is a deterministic scalar satisfying
  \begin{align*}
    \rho^2 \geq 8 a s_\delta (\rho), \quad \mbox{and} \quad \rho \geq \frac{D}{\numobs^2 K^2}.
  \end{align*}
  Then we have
  \begin{align*}
   \Prob \Big( a Z(\rhohat) - \rhohat^2 \geq 3 a (\rhofix)^2 \Big) \leq  2 (\log \numobs + \log K)\delta. 
  \end{align*}
\end{lemma}
\noindent See \Cref{app:subsec-proof-lemma-random-radius-empirical-process} for the proof of this lemma.

Now we use these lemmas to bound the estimation error. To simplify the notation, we use $s_\delta^{*}, s_\delta^{\mathrm{cross}}, s_\delta^{\mathrm{smpl}}$ to denote the right-hand-sides of the bounds in \Cref{lemma:errterm-main-bound,lemma:errterm-gen-bound,lemma:errterm-sampling-bound}, as functions of $\rho$ and $\delta$, respectively. For $\diamond \in \{*, \mathrm{cross}, \mathrm{smpl}\}$, also define $\rhofixtilde_\diamond$ as the largest deterministic radius such that
\begin{align*}
  (\rhofixtilde)^2 \leq \frac{16}{\min (\lammin, \sfactor, 1)} s_\delta^{(\diamond)} (\rhofixtilde).
\end{align*}
It is straightforward to verify that $\rhofixtilde_\diamond \geq \frac{\smoothness_\Fclass}{\numobs^2 K^2}$, for $\diamond \in \{*, \mathrm{cross}, \mathrm{smpl}\}$. By \Cref{lemma:random-radius-empirical-process}, for $\delta \leq \frac{1}{2 (\log \numobs + \log K)}$, with probability $1 - \delta$, we have
\begin{align*}
  Z^{\diamond} (\rhohat) \leq 3 (\rhofixtilde_\diamond)^2 + \frac{\min (\lammin, \sfactor, 1)}{8} (\rhohat)^2, \quad \mbox{for } \diamond \in \{*, \mathrm{cross}, \mathrm{smpl}\}.
\end{align*}
Substituting to Equations~\eqref{eq:basic-inequality-estimation-error} and~\eqref{eq:basic-inequality-lower-bound}, with probability $1 - 3 \delta$, we have
\begin{align}
  \frac{\min (\lammin, \sfactor, 1)}{8} (\rhohat)^2 \leq 3 (\rhofixtilde_*)^2 + 3 (\rhofixtilde_{\mathrm{cross}})^2 + 3 (\rhofixtilde_{\mathrm{smpl}})^2.\label{eq:rate-theorem-near-complete-form}
\end{align}
By taking the constant $c_1 \geq \frac{32 c}{\min (\lammin, \sfactor, 1)} s_\delta^{(\diamond)}$ in \Cref{eq:defn-critical-radii}, where the constant $c$ is the one in \Cref{lemma:errterm-main-bound,lemma:errterm-gen-bound,lemma:errterm-sampling-bound}, we have
\begin{align*}
  \rhofixtilde_* \leq \max (\critradistar_{\numobs, \delta} (\ExpValTrue - \fbar) , \critradismpl_{\numobs, K , \delta} ),\quad
  \rhofixtilde_{\mathrm{cross}} \leq \max (\critradicross_{\numobs, \delta}  , \critradismpl_{\numobs, K , \delta} ),\quad
  \rhofixtilde_{\mathrm{smpl}} \leq \critradismpl_{\numobs, K , \delta}.
\end{align*}
Substituting them to \Cref{eq:rate-theorem-near-complete-form}, we complete the proof of \Cref{thm:main-error}.

\subsubsection{Proof of \Cref{lemma:variance-bound-xi-process}}\label{subsubsec:proof-lemma-variance-bound-xi-process}
To facilitate the analysis, we define the auxiliary error term
\begin{align*}
  \widetilde{\xi}_i (f, g) &\mydefn \Exs \big[ \xi_i (f, g) \mid (\State_t^{(i)})_{t \in [0, T]}\big]
  =  f_T (\State_T^{(i)}) g_T( \State_T^{(i)} )   - \int_0^T  \Big\{ \big( \partial_t + \generator_t + \sfactor \reward_t \big) f_t (\MyState_t^{(i)})  \cdot g_t (\MyState_t^{(i)}) \Big\} dt.
\end{align*}
For two pairs $(f^{(1)}, g^{(1)})$, $(f^{(2)}, g^{(2)})$ of functions, we have
\begin{align*}
  &\widetilde{\xi}_i (f^{(1)}, g^{(1)}) - \widetilde{\xi}_i (f^{(2)}, g^{(2)})\\
   &= f^{(1)}_T (\State_T) \cdot \big( g^{(1)} - g_T^{(2)} \big) (\State_T) -  \int_0^T  \big( \partial_t + \generator_t + \sfactor \reward_t \big)  f_t^{(1)} (\MyState_t^{(i)})  \cdot \big( g_t^{(1)} - g_t^{(2)}\big) (\MyState_t^{(i)}) dt \\
   &\qquad \qquad + \big( f^{(1)} - f_T^{(2)} \big) (\State_T) \cdot g^{(2)}_T (\State_T) - \int_0^T  \big( \partial_t + \generator_t + \sfactor \reward_t \big) \big( f_t^{(1)} - f_t^{(2)} \big) (\MyState_t^{(i)})  \cdot g_t^{(2)} (\MyState_t^{(i)}) dt.
\end{align*}
First, for the terms not involving gradients, we directly use Cauchy--Schwarz inequality to note that
\begin{subequations}\label{eq:naive-bound-1-in-xi-var-bound}
\begin{align}
  \Exs \big[ \abss{f^{(1)}_T (\State_T) \cdot \big( g^{(1)} - g_T^{(2)} \big) (\State_T)}^2 \big] &\leq \lpnorm{f_T^{(1)}}{4}{\density_T}^2 \lpnorm{g_T^{(1)} - g_T^{(2)}}{4}{\density_T}^2 \leq \energynorm{f^{(1)}}{1}{4}{T}^2 \energynorm{g^{(1)} - g^{(2)}}{1}{4}{T}^2,\\
  \Exs \big[ \abss{ \big( f^{(1)}_T - f^{(2)}_T \big) (\State_T) \cdot g_T^{(2)}  (\State_T)}^2 \big] &\leq \lpnorm{f_T^{(1)} - f_T^{(2)}}{4}{\density_T}^2 \lpnorm{ g_T^{(2)}}{4}{\density_T}^2 \leq \energynorm{f^{(1)} - f^{(2)}}{1}{4}{T}^2 \energynorm{ g^{(2)}}{1}{4}{T}^2.
\end{align}
\end{subequations}
And for the time integral,
\begin{subequations}\label{eq:naive-bound-2-in-xi-var-bound}
\begin{multline}
  \Exs \Big[ \abss{\int_0^T \reward_t   f_t^{(1)} (\MyState_t^{(i)})  \cdot \big( g_t^{(1)} - g_t^{(2)}\big) (\MyState_t^{(i)}) dt}^2 \Big] \leq 4 \Exs \Big[ \int_0^T  |f_t^{(1)} (\MyState_t^{(i)})|^2 dt \cdot \int_0^T  |\big( g_t^{(1)} - g_t^{(2)}\big) (\MyState_t^{(i)})|^2  dt  \Big]\\
  \leq 4 \sqrt{ \int_0^T \Exs \big[ |f_t^{(1)} (\MyState_t^{(i)})|^4 \big] dt} \cdot \sqrt{\int_0^T  |\big( g_t^{(1)} - g_t^{(2)}\big) (\MyState_t^{(i)})|^4  dt} \leq 4 \energynorm{f^{(1)}}{1}{4}{T}^2 \energynorm{g^{(1)} - g^{(2)}}{1}{4}{T}^2,
\end{multline}
and similarly
\begin{align}
  \Exs \Big[ \abss{\int_0^T \reward_t  \big(f_t^{(1)} - f_t^{(2)} \big) (\MyState_t^{(i)})  \cdot g_t^{(2)} (\MyState_t^{(i)}) dt}^2 \Big]  \leq 4 \energynorm{f^{(1)} - f^{(2)}}{1}{4}{T}^2 \energynorm{g^{(2)}}{1}{4}{T}^2.
\end{align}
\end{subequations}
Now we turn to bound the terms involving the diffusion generator. We use the following lemma, which is adapted from the paper~\cite{mou2025statistical}.
\begin{lemma}\label{lemma:cross-cov-bound-diffusion}
  Given a pair of functions $f, g: [0, T] \times \real^\usedim \rightarrow \real$, for any $\delta \in (0, T/2)$, we have
  \begin{align*}
    \Exs \Big[ \abss{\int_0^T \generator_t f_t (\State_t) \cdot g_t (\State_t) dt}^2 \Big] \leq c \Big\{ \log (T/\delta) \energynorm{f}{1}{4 \pq}{T}^2 \energynorm{g}{1}{4 \pq}{T}^2 + \delta \energynorm{f}{2}{4 \pq}{T}^2 \energynorm{g}{0}{4 \pq}{T}^2  \Big\},
  \end{align*}
  where the constant $c$ depends on the parameters in Assumptions~\ref{assume:moment-bound} and~\ref{assume:density-regular} and the exponent $\pq$.
\end{lemma}
\noindent See \Cref{subsubsec:proof-lemma-cross-cov-bound-diffusion} for the proof of this lemma.

Invoking \Cref{lemma:cross-cov-bound-diffusion}, for any $\delta \in (0, T/2)$, we have
\begin{subequations}\label{eq:generator-bound-in-xi-var-bound}
\begin{multline}
  \Exs \Big[ \abss{\int_0^T \generator_t  f_t^{(1)} (\MyState_t)  \cdot \big( g_t^{(1)} - g_t^{(2)}\big) (\MyState_t) dt}^2 \Big] \\
  \leq c \energynorm{g^{(1)} - g^{(2)}}{1}{4 \pq}{T}^2  \energynorm{f^{(1)}}{1}{4 \pq}{T}^2  \log (T/\delta) + c\delta \energynorm{g^{(1)} - g^{(2)}}{0}{4 \pq}{T}^2   \energynorm{f^{(1)}}{2}{4 \pq}{T}^2,
\end{multline}
and similarly,
\begin{multline}
  \Exs \Big[ \abss{\int_0^T \generator_t \big( f_t^{(1)} - f_t^{(2)} \big) (\MyState_t)  \cdot g_t^{(2)} (\MyState_t) dt}^2 \Big] \\
  \leq c \energynorm{f^{(1)} - f^{(2)}}{1}{4 \pq}{T}^2 \energynorm{g^{(2)}}{1}{4 \pq}{T}^2  \log (T/ \delta) + c \delta \energynorm{f^{(1)} - f^{(2)}}{2}{4 \pq}{T}^2  \energynorm{g^{(2)}}{0}{4 \pq}{T}^2 .
\end{multline}
\end{subequations}
For the time derivative term, we use the following lemma.
\begin{lemma}\label{lemma:time-derivative-cross-var-bound}
  For a pair of functions $f, g$, we have
  \begin{align*}
    \Exs \Big[ \abss{\int_0^T \partial_t f_t (\State_t) g_t (\State_t) dt}^2 \Big] \leq \min \Big\{ 9 \energynorm{f}{1}{4}{T}^2 \dynanorm{g}{1}{4}{T}^2, \energynorm{g}{1}{4}{T}^2 \dynanorm{f}{1}{4}{T}^2  \Big\}
  \end{align*}
\end{lemma}
\noindent See \Cref{subsubsec:proof-lemma-time-derivative-cross-var-bound} for the proof of this lemma.

Applying this lemma to the time derivative terms, we have
\begin{subequations}\label{eq:time-derivative-bound-in-xi-var-bound}
\begin{align}
  \Exs \Big[ \abss{\int_0^T \partial_t f_t^{(1)} (\MyState_t) \cdot \big( g_t^{(1)} - g_t^{(2)}\big) (\MyState_t) dt}^2 \Big] &\leq 9 \energynorm{f^{(1)}}{1}{4}{T}^2 \dynanorm{g^{(1)} - g^{(2)}}{1}{4}{T}^2,\\
   \Exs \Big[ \abss{\int_0^T \partial_t  \big( f_t^{(1)} - f_t^{(2)} \big) (\MyState_t) \cdot g_t^{(2)}  (\MyState_t) dt}^2 \Big] &\leq 9 \energynorm{f^{(1)} - f^{(2)}}{1}{4}{T}^2 \dynanorm{g^{(2)}}{1}{4}{T}^2.
\end{align}
\end{subequations}
Combining \Cref{eq:naive-bound-1-in-xi-var-bound,eq:naive-bound-2-in-xi-var-bound,eq:generator-bound-in-xi-var-bound,eq:time-derivative-bound-in-xi-var-bound}, we conclude the bound
\begin{align}
 &\Exs \Big[ \abss{\widetilde{\xi}_i (f^{(1)}, g^{(1)}) - \widetilde{\xi}_i (f^{(2)}, g^{(2)})}^2 \Big] \nonumber\\
&\leq 2 c \log (T / \delta) \Big\{ \energynorm{f^{(1)}}{1}{4 \pq}{T}^2 \dynanorm{g^{(1)} - g^{(2)}}{1}{4 \pq}{T}^2 + \energynorm{f^{(1)} - f^{(2)}}{1}{4 \pq}{T}^2 \dynanorm{g^{(2)}}{1}{4 \pq}{T}^2 \Big\} \nonumber\\
&\qquad + 2 c \delta \Big\{ \energynorm{f^{(1)}}{2}{4 \pq}{T}^2 \dynanorm{g^{(1)} - g^{(2)}}{0}{4 \pq}{T}^2 + \energynorm{f^{(1)} - f^{(2)}}{2}{4 \pq}{T}^2 \dynanorm{g^{(2)}}{0}{4 \pq}{T}^2 \Big\}.\label{eq:xii-var-bound-in-errterm-gen-bound-proof}
\end{align}
As for the additional noises from discrete sampling of the continuous-time process, we use the following lemma to bound its variance.

\begin{lemma}\label{lemma:discrete-sampling-in-xi-var-bound}
  Under the setup of \Cref{lemma:variance-bound-xi-process}, we have
  \begin{multline*}
    \Exs \Big[ \var \Big( \xi_i (f^{(1)}, g^{(1)}) - \xi_i (f^{(2)}, g^{(2)}) \mid (\State_t^{(i)})_{t \in [0, T]} \Big) \Big] \\
    \leq \frac{c}{K} \Big\{ \dynanorm{f^{(1)}}{2}{4}{T}^2  \energynorm{g^{(1)} - g^{(2)}}{0}{4}{T}^2 + 2 \dynanorm{f^{(1)} - f^{(2)}}{2}{4}{T}^2  \energynorm{ g^{(2)}}{0}{4}{T}^2 \Big\}.
  \end{multline*}
\end{lemma}
\noindent We prove this lemma in \Cref{subsec:proof-lemma-discrete-sampling-in-xi-var-bound}. Let us now complete the proof of \Cref{lemma:variance-bound-xi-process} taking this lemma as given.

Combining \Cref{eq:xii-var-bound-in-errterm-gen-bound-proof} and \Cref{lemma:discrete-sampling-in-xi-var-bound}, and letting $\delta = T / K$, we have
\begin{align*}
  &\var \Big(  \xi_i (f^{(1)}, g^{(1)}) - \xi_i (f^{(2)}, g^{(2)}) \Big)\\
  & = \var \Big(  \widetilde{\xi}_i (f^{(1)}, g^{(1)}) - \widetilde{\xi}_i (f^{(2)}, g^{(2)}) \Big) + \Exs\Big[\var \Big( \xi_i (f^{(1)}, g^{(1)}) - \xi_i (f^{(2)}, g^{(2)}) \mid (\State_t)_{t \in [0, T]} \Big)\Big]\\
  & \leq 3 c \log K \Big\{ \dynanorm{g^{(1)} - g^{(2)}}{1}{4 \pq}{T}^2  \energynorm{f^{(1)}}{1}{4 \pq}{T}^2 + \dynanorm{f^{(1)} - f^{(2)}}{1}{4 \pq}{T}^2  \energynorm{g^{(2)}}{1}{4 \pq}{T}^2 \Big\} \\
    &\qquad +  \frac{3 c T}{K} \Big\{ \energynorm{g^{(1)} - g^{(2)}}{0}{4 \pq}{T}^2  \dynanorm{f^{(1)}}{2}{4 \pq}{T}^2 + \dynanorm{f^{(1)} - f^{(2)}}{2}{4 \pq}{T}^2  \energynorm{g^{(2)}}{0}{4 \pq}{T}^2 \Big\},
\end{align*}
which proves \Cref{lemma:variance-bound-xi-process}.

\subsubsection{Proof of \Cref{lemma:errterm-main-bound}}\label{subsubsec:proof-lemma-errterm-main-bound}
Similar to the proof of \Cref{lemma:errterm-gen-bound}, we bound the variance and Orlicz norm of the process $\xi_i (\ExpValTrue - \fbar, h^{(1)} - h^{(2)})$ for $h^{(1)}, h^{(2)} \in \shiftConvSet \cap \strucball (\rho)$. Applying Lemma~\ref{lemma:variance-bound-xi-process} with $f^{(1)} = \ExpValTrue - \fbar$, $g^{(1)} = h^{(1)} - h^{(2)}$, and $f^{(2)} = g^{(2)} = 0$, we have
\begin{align*}
  &\var \Big( \xi_i (\ExpValTrue - \fbar, h^{(1)}) - \xi_i (\ExpValTrue - \fbar, h^{(2)}) \Big)\\
  &\leq c \log K \energynorm{h^{(1)} - h^{(2)}}{1}{4 \pq}{T}^2 \dynanorm{ \ExpValTrue - \fbar }{1}{4 \pq}{T}^2 + \frac{c T}{K} \energynorm{h^{(1)} - h^{(2)}}{0}{4 \pq}{T}^2 \dynanorm{ \ExpValTrue - \fbar}{2}{4 \pq}{T}^2\\
  &\leq c  \tau^2  \strucnorm{h^{(1)} - h^{(2)}}{T}^2 \Big\{  \log K \cdot \dynanorm{ \ExpValTrue - \fbar }{1}{4 \pq}{T}^2 + \frac{1}{K} \dynanorm{ \ExpValTrue - \fbar }{2}{4 \pq}{T}^2 \Big\}.
\end{align*}
As for the Orlicz norm, we have
\begin{align*}
  &\vecnorm{\xi_i (\ExpValTrue - \fbar, h^{(1)}) - \xi_i (\ExpValTrue - \fbar, h^{(2)})}{\psi_1} \\
  &\leq \vecnorm{\ExpValTrue - \fbar}{\Ctwospace} \cdot \Big\{ \vecnorm{h^{(1)}_T (\State_T) - h^{(2)}_T (\State_T)}{\psi_1} + \frac{1}{K} \sum_{k = 1}^K \vecnorm{h^{(1)}_{\smpltime{k}{i}} (\State_{\smpltime{k}{i}}) - h^{(2)}_{\smpltime{k}{i}} (\State_{\smpltime{k}{i}})}{\psi_1} \Big\}\\
  &\leq 2 \vecnorm{\ExpValTrue - \fbar}{\Ctwospace} \cdot \vecnorm{h^{(1)} - h^{(2)}}{\holderspace^0}.
\end{align*}
By Bernstein's inequality, for any $\delta \in (0, 1)$, with probability $1 - \delta$, we have
\begin{multline*}
  \abss{\frac{1}{\numobs} \sum_{i = 1}^\numobs \xi_i (\ExpValTrue - \fbar, h^{(1)} - h^{(2)})} \leq c \tau \strucnorm{h^{(1)} - h^{(2)}}{T} \sqrt{\frac{\log (1/ \delta)}{\numobs}} \Big\{  \dynanorm{ \ExpValTrue - \fbar }{1}{4 \pq}{T}  \sqrt{\log K} + \frac{\dynanorm{ \ExpValTrue - \fbar }{2}{4 \pq}{T} }{\sqrt{K}}\Big\}\\
  +c  \vecnorm{\ExpValTrue - \fbar}{\Ctwospace} \cdot \vecnorm{h^{(1)} - h^{(2)}}{\holderspace^0} \frac{\log (\numobs / \delta)}{\numobs}. 
\end{multline*}
Applying a mixed-norm chaining argument, we can bound the supremum of the empirical process over $h \in \shiftConvSet \cap \strucball (\rho)$ as
\begin{align*}
  \errtermmain (\rho) &\leq c \Big\{ \tau \dynanorm{ \ExpValTrue - \fbar }{1}{4 \pq}{T} \sqrt{\frac{\log K}{\numobs}} + \frac{ \dynanorm{ \ExpValTrue - \fbar }{2}{4 \pq}{T} }{\sqrt{\numobs K}} \Big\} \Big\{ \dudley_2 \big( \shiftConvSet (\rho); \strucnorm{\cdot}{T} \big) + \rho \sqrt{\log (1 / \delta)} \Big\}\\
  &\qquad + c \vecnorm{\ExpValTrue - \fbar}{\Ctwospace} \frac{\log \numobs}{\numobs} \cdot \Big\{ \dudley_1 \big( \convSet; \vecnorm{\cdot}{\holderspace^0} \big) + \smoothness_\Fclass \log (1 / \delta) \Big\},
\end{align*}
which proves \Cref{lemma:errterm-main-bound}.

\subsubsection{Proof of \Cref{lemma:errterm-gen-bound}}\label{subsubsec:proof-lemma-errterm-gen-bound}
Recall the definition~\eqref{eq:defn-xi-process} of the process $\xi_i (f, g)$. We can rewrite the target empirical process supremum as
\begin{align*}
  \errtermgen (\rho) = \sup_{f, g \in \shiftConvSet \cap \strucball (\rho)} \Upsilon_\numobs (f, g), \quad \mbox{where} ~ \Upsilon_\numobs (f, g) \mydefn \frac{1}{\numobs} \sum_{i = 1}^\numobs \Big( \xi_i (f, g) - \Exs \big[ \xi_i (f, g) \big] \Big).
\end{align*}

Clearly, we have $\Exs [\Upsilon_\numobs (f, g)] = 0$. In order to bound its supremum, we use chaining techniques. For functions $f^{(1)}, f^{(2)}, g^{(1)}, g^{(2)} \in \shiftConvSet \cap \strucball (\rho)$, \Cref{lemma:variance-bound-xi-process} implies that
\begin{align*}
  &\var \big( \xi_i (f^{(1)}, g^{(1)}) - \xi_i (f^{(2)}, g^{(2)}) \big)\\
  &\leq c \log K \Big\{ \dynanorm{g^{(1)} - g^{(2)}}{1}{4 \pq}{T}^2  \energynorm{f^{(1)}}{1}{4 \pq}{T}^2 + \dynanorm{f^{(1)} - f^{(2)}}{1}{4 \pq}{T}^2  \energynorm{g^{(2)}}{1}{4 \pq}{T}^2 \Big\} \\
    &\qquad +  \frac{c T}{K} \Big\{ \energynorm{g^{(1)} - g^{(2)}}{0}{4 \pq}{T}^2  \dynanorm{f^{(1)}}{2}{4 \pq}{T}^2 + \dynanorm{f^{(1)} - f^{(2)}}{2}{4 \pq}{T}^2  \energynorm{g^{(2)}}{0}{4 \pq}{T}^2 \Big\}\\
  &\leq  2 c \tau^4 \kappa^2 \Big( \strucnorm{f^{(1)} - f^{(2)}}{T}^2 + \strucnorm{g^{(1)} - g^{(2)}}{T}^2 \Big) \cdot \Big\{ \rho^2 \log K + \frac{\sigma^2 T}{K} \rho^{2 \eta} \Big\}\\
  &\qquad  + \frac{cT}{K}\tau^2  \Big(  \vecnorm{f^{(1)} - f^{(2)}}{\Ctwospace}^2 + \vecnorm{g^{(1)} - g^{(2)}}{\Ctwospace}^2  \Big) \rho^2 .
\end{align*}
where in the last inequality, we use Assumptions~\ref{assume:moment-comparison} and~\ref{assume:subgaussian-class} to obtain the norm dominance relations
\begin{align*}
  \energynorm{h}{1}{4 \pq}{T} \leq \tau \strucnorm{h}{T},\qquad & \energynorm{h}{1}{4 \pq}{T} \leq \tau \strucnorm{h}{T} ,\\
  \energynorm{h}{2}{4 \pq}{T} \leq \sigma \strucnorm{h}{T}^\eta, \qquad &\strongnorm{h}{T} \leq \kappa \strucnorm{h}{T},
\end{align*}
for any function $h \in \shiftConvSet$.

In order to prove high-probability bounds for the pointwise difference $\Upsilon_\numobs (f^{(1)}, g^{(1)}) - \Upsilon_\numobs (f^{(2)}, g^{(2)})$, we also need an Orlicz norm upper bound. We note that
\begin{align}
  &\vecnorm{\xi_i (f^{(1)}, g^{(1)}) - \xi_i (f^{(2)}, g^{(2)})}{\psi_1}\nonumber \\
  & \leq \vecnorm{f^{(1)} (\State_T) g^{(1)} (\State_T) - f^{(2)} (\State_T) g^{(2)} (\State_T)}{\psi_1} \nonumber \\
  &\qquad + \frac{T}{K} \sum_{k = 1}^K  \vecnorm{ ( \partial_t + \generator_{\smpltime{k}{i}}  + \randreward_{\smpltime{k}{i}} ) f^{(1)}_{\smpltime{k}{i}} (\MyState_{\smpltime{k}{i}}^{(i)})  \cdot \big( g^{(1)}_{\smpltime{k}{i}} - g^{(2)}_{\smpltime{k}{i}} \big)(\MyState_{\smpltime{k}{i}}^{(i)}) }{\psi_1} \nonumber\\
  &\qquad \qquad +  \frac{T}{K} \sum_{k = 1}^K  \vecnorm{ ( \partial_t + \generator_{\smpltime{k}{i}}  + \randreward_{\smpltime{k}{i}} )  \big( f^{(1)}_{\smpltime{k}{i}} - f^{(2)}_{\smpltime{k}{i}} \big)(\MyState_{\smpltime{k}{i}}^{(i)})  \cdot  g^{(2)}_{\smpltime{k}{i}} (\MyState_{\smpltime{k}{i}}^{(i)}) }{\psi_1}\nonumber \\
  &\leq T \cdot  \vecnorm{g^{(1)} - g^{(2)}}{\infty} \cdot ( \vecnorm{( \partial_t + \generator_{\smpltime{k}{i}}  + \randreward_{\smpltime{k}{i}} ) f^{(1)}_{\smpltime{k}{i}} (\MyState_{\smpltime{k}{i}}^{(i)})}{\psi_1} + \vecnorm{f^{(1)} (\State_T)}{\psi_1} + \vecnorm{f^{(2)} (\State_T)}{\psi_1} )\nonumber\\
  &\qquad  + T \cdot \vecnorm{ ( \partial_t + \generator_{\smpltime{k}{i}}  + \randreward_{\smpltime{k}{i}} )  \big( f^{(1)}_{\smpltime{k}{i}} - f^{(2)}_{\smpltime{k}{i}} \big)(\MyState_{\smpltime{k}{i}}^{(i)})}{\psi_2} \cdot \vecnorm{g^{(2)} (\MyState_{\smpltime{k}{i}}^{(i)})}{\psi_2} \nonumber\\
 &\qquad \qquad + T \cdot \vecnorm{f^{(1)} - f^{(2)}}{\infty} \cdot \big(  \vecnorm{g^{(1)} (\State_T)}{\psi_1} + \vecnorm{g^{(2)} (\State_T)}{\psi_1} \big).
  \label{eq:orlicz-norm-initial-in-errterm-gen-proof}
\end{align}
We first analyze the terms involving the diffusion generator.
By Assumptions~\ref{assume:moment-bound} and~\ref{assume:subgaussian-class}, we note that
\begin{align*}
  \vecnorm{ \generator_{\smpltime{k}{i}} h_{\smpltime{k}{i}} (\MyState_{\smpltime{k}{i}}^{(1)})}{\psi_1} &\leq \sup_{p \geq 2} \frac{1}{p} \Big\{ \frac{1}{T} \int_0^T \Exs \Big[ \abss{\inprod{\drift_t}{h_t} (\State_t) + \frac{1}{2} \mathrm{Tr} \big( \covMat_t \nabla^2 h_t \big) (\State_t)}^p \Big] dt \Big\}^{1/p}\\
  &\leq \sup_{p \geq 2} \frac{1}{p} \Big\{ \frac{1}{T} \int_0^T \Big[ \big( 2 \sqrt{p} \smoothness \sobopstatnorm{h_t}{p}{\density_t} \big)^p + \big( 2 \sqrt{p} \smoothness \sobokpnorm{h_t}{2}{p}{\density_t} \big)^p \Big] dt \Big\}^{1/p}\\
  &\leq 4 \sigma \smoothness \strucnorm{h}{T}^\eta,
\end{align*}
and the $\psi_2$ norm bound
\begin{align*}
  \vecnorm{ \generator_{\smpltime{k}{i}} h_{\smpltime{k}{i}} (\MyState_{\smpltime{k}{i}}^{(1)})}{\psi_2} &\leq \sup_{p \geq 2} \frac{1}{\sqrt{p}} \Big\{ \frac{1}{T} \int_0^T \Exs \Big[ \abss{\inprod{\drift_t}{h_t} (\State_t) + \frac{1}{2} \mathrm{Tr} \big( \covMat_t \nabla^2 h_t \big) (\State_t)}^p \Big] dt \Big\}^{1/p}\\
  &\leq \sup_{p \geq 2} \frac{1}{\sqrt{p}} \Big\{ \frac{1}{T} \int_0^T \Big[ \big( 2 \sqrt{p} \smoothness \vecnorm{\nabla h_t}{\infty} \big)^p + \big( 2 \sqrt{p} \smoothness \vecnorm{\nabla^2 h_t}{\infty} \big)^p \Big] dt \Big\}^{1/p}\\
  &\leq 2 \smoothness \vecnorm{h}{\holderspace^2}.
\end{align*}
For time derivative and constant terms, we note that for any $h \in \shiftConvSet$, we have
\begin{align*}
  \vecnorm{( \partial_t  + \randreward_{\smpltime{k}{i}} ) h_{\smpltime{k}{i}} (\MyState_{\smpltime{k}{i}}^{(1)})}{\psi_2} \leq \sigma \strongnorm{h}{T}^\eta \leq \kappa \sigma \strucnorm{h}{T}^\eta.
\end{align*}
As for the terminal value terms, Assumption~\ref{assume:subgaussian-class} implies that
\begin{align*}
  \vecnorm{h_T (\State_T) }{\psi_1} \leq \sigma \strucnorm{h}{T}^\eta.
\end{align*}

Putting them together, and substituting them into the Orlicz norm bound~\eqref{eq:orlicz-norm-initial-in-errterm-gen-proof}, we have
\begin{align*}
  \vecnorm{\xi_i (f^{(1)}, g^{(1)}) - \xi_i (f^{(2)}, g^{(2)})}{\psi_1}
  &\leq 4 \kappa \sigma \smoothness T \vecnorm{g^{(1)} - g^{(2)}}{\infty}  \strucnorm{f^{(1)}}{T}^\eta + 2 \kappa \sigma \smoothness T \vecnorm{f^{(1)} - f^{(2)}}{\Ctwospace} \cdot \strucnorm{g^{(2)} (\State_{t_k})}{T}\\
  &\leq  6 \sigma \kappa \smoothness T \rho^\eta \big(\vecnorm{f^{(1)} - f^{(2)}}{\Ctwospace} + \vecnorm{g^{(1)} - g^{(2)}}{\infty} \big).
\end{align*}
Invoking Bernstein's inequality for unbounded random variables (\cite{adamczak2008tail}, Theorem 4), with probability $1 - \delta$, we have
\begin{align*}
  &\abss{ \Upsilon_\numobs (f^{(1)}, g^{(1)}) - \Upsilon_\numobs (f^{(2)}, g^{(2)}) } \\
 & \leq  c \tau^2 \kappa \big( \strucnorm{f^{(1)} - f^{(2)}}{T} + \strucnorm{g^{(1)} - g^{(2)}}{T} \big) \rho \sqrt{\frac {\log K \log (1 / \delta)}{\numobs}} \\
 &\qquad + c \sigma \kappa \rho^\eta \big( \vecnorm{f^{(1)} - f^{(2)}}{\Ctwospace} + \vecnorm{g^{(1)} - g^{(2)}}{\Ctwospace} \big) \frac{\log (1 / \delta)}{\numobs} \\
 &\qquad + c \Big\{ \tau^2 \kappa \big( \strucnorm{f^{(1)} - f^{(2)}}{T} + \strucnorm{g^{(1)} - g^{(2)}}{T} \big) \sigma \rho^\eta + \tau \rho \big( \vecnorm{f^{(1)} - f^{(2)}}{\Ctwospace} + \vecnorm{g^{(1)} - g^{(2)}}{\Ctwospace} \big) \Big\}  \sqrt{\frac {\log (1 / \delta)}{\numobs K}}.
\end{align*}
Now we invoke the generic chaining tail bound (\cite{dirksen2015tail}, Theorem 3.5) to bound the supremum of the empirical process $\Upsilon_\numobs (h)$ over the set $\shiftConvSet \cap \strucball (\rho)$. With probability $1 - \delta$, we have
\begin{multline*}
  \sup_{f, g \in \shiftConvSet \cap \strucball (\rho)} |\Upsilon_\numobs (f, g)| \\
  \leq  c \tau^2 \kappa \rho \sqrt{\frac{\log K}{\numobs}} \Big\{ \gamma_2 \big( \shiftConvSet (\rho)^{\otimes 2}; \strucnorm{\cdot}{T} \big) + \rho \sqrt{\log (1 / \delta)} \Big\} +  \frac{c \kappa \sigma \rho^\eta}{\numobs} \Big\{ \gamma_1 \big( \convSet^{\otimes 2}; \vecnorm{\cdot}{\Ctwospace} \big) + \smoothness_\Fclass \log (1 / \delta) \Big\}\\
  + \frac{\tau^2 \kappa \rho^\eta }{\sqrt{\numobs K}} \Big\{ \gamma_2 (\convSet^{\otimes 2}; \vecnorm{\cdot}{\Ctwospace}) + \smoothness_\Fclass \sqrt{\log (1/ \delta)} \Big\},
\end{multline*}
where we use $\mathcal{H}^{\otimes 2}$ to denote the tensor product $\{ (h_1, h_2): h_1, h_2 \in \mathcal{H}\}$ for a given function class $\mathcal{H}$.

For the generic chaining functionals, we note that
\begin{align*}
  \gamma_2 \big(\shiftConvSet (\rho)^{\otimes 2}; \strucnorm{\cdot}{T} \big) &\leq c \dudley_2 \big( \shiftConvSet (\rho)^{\otimes 2}; \strucnorm{\cdot}{T} \big) \leq \sqrt{2} c \dudley_2 \big( \shiftConvSet (\rho); \strucnorm{\cdot}{T} \big), \qquad \mbox{and}\\
  \gamma_1 \big(\convSet^{\otimes 2}; \vecnorm{\cdot}{\holderspace^2}\big) &\leq c \dudley_1 \big( \convSet^{\otimes 2}; \vecnorm{\cdot}{\holderspace^2} \big) \leq 2 c \dudley_1 \big( \convSet; \vecnorm{\cdot}{\holderspace^2} \big).
\end{align*}
Substituting to the previous bound, we conclude the proof of \Cref{lemma:errterm-gen-bound}.

\subsubsection{Proof of \Cref{lemma:errterm-sampling-bound}}\label{subsubsec:proof-lemma-errterm-sampling-bound}
Note that the optimal value function $\ExpValTrue$ solves the parabolic PDE~\eqref{eq:hjb-eq-under-control-affine-form}. So we have
\begin{align*}
  \zeta_i (h) &=  \frac{T}{K} \sum_{k = 1}^K   \big( \partial_t + \generator_{\smpltime{k}{i}} + \sfactor \Reward_{\smpltime{k}{i}} \big)  \ExpValTrue_{\smpltime{k}{i}} (\MyState_{\smpltime{k}{i}}^{(i)})  \cdot  h_{\smpltime{k}{i}} (\MyState_{\smpltime{k}{i}}^{(i)})\\
  & = \frac{\sfactor T}{K} \sum_{k = 1}^K   (\Reward_{\smpltime{k}{i}}^{(i)} - \reward (\State_{\smpltime{k}{i}}^{(i)})) \ExpValTrue_{\smpltime{k}{i}} (\State_{\smpltime{k}{i}}^{(i)})  h_{\smpltime{k}{i}} (\MyState_{\smpltime{k}{i}}^{(i)}).
\end{align*}
We note the variance bound
\begin{align*}
  \Exs \big[ |\zeta_i (h^{(1)} - h^{(2)})|^2 \big] &= \frac{\sfactor^2 T^2}{K} \Exs \Big[ \abss{(\Reward_{\smpltime{k}{i}}^{(i)} - \reward (\State_{\smpltime{k}{i}}^{(i)})) \ExpValTrue (\State_{\smpltime{k}{i}}^{(i)})  (h_{\smpltime{k}{i}}^{(1)} - h_{\smpltime{k}{i}}^{(2)}) (\State_{\smpltime{k}{i}}^{(i)}) }^2 \Big]\\
   &\leq \frac{4 \smoothness_\Fclass^2 \sfactor^2 T^2}{K}  \int_0^T \Exs \big[ \abss{(h_t^{(1)} - h_t^{(2)}) (\State_t)^2} \big] dt \leq \frac{4 \smoothness_\Fclass^2 \sfactor^2 T^2}{K} \energynorm{h^{(1)} - h^{(2)}}{0}{2}{T}
\end{align*}
As for the Orlicz norm of the process $\zeta_i$, we note that conditionally on $(\State_t^{(i)})_{t \in [0, T]}$ and $(\smpltime{k}{i})_{k = 1}^K$, the term $\zeta_i (h^{(1)} - h^{(2)})$ is a sum of zero-mean, bounded and independent random variables. By Hoeffding's inequality, for any $\varepsilon > 0$, we have
\begin{align*}
  \Prob \Big( \abss{\zeta_i (h^{(1)} - h^{(2)})} \geq \varepsilon \mid (\State_t^{(i)})_{t \in [0, T]}, (\smpltime{k}{i})_{k = 1}^K \Big) \leq 2 \exp \Big\{ -\frac{ K^2 \varepsilon^2}{8 \smoothness_\Fclass^2 \sfactor^2 T^2 \sum_{k = 1}^K (h^{(1)} - h^{(2)}) (\State_{\smpltime{k}{i}}^{(i)})^2} \Big\}.
\end{align*}
Consequently, we can bound the Orlicz norm as
\begin{align*}
  \vecnorm{\zeta_i (h^{(1)} - h^{(2)})}{\psi_2} \leq \frac{4 \smoothness_\Fclass \sfactor T }{\sqrt{K}} \vecnorm{h^{(1)} - h^{(2)}}{\holderspace^0}. 
\end{align*}
By Bernstein's inequality for unbounded random variables, for any $\delta \in (0, 1)$, with probability $1 - \delta$, we have
\begin{align*}
  \abss{ \frac{1}{\numobs} \sum_{i = 1}^\numobs \zeta_i (h^{(1)} - h^{(2)}) } &\leq c \frac{\smoothness_\Fclass \sfactor T }{\sqrt{K}} \Big\{ \energynorm{h^{(1)} - h^{(2)}}{0}{2}{T} \sqrt{\frac{\log (1 / \delta)}{\numobs}} + \vecnorm{h^{(1)} - h^{(2)}}{\holderspace^0} \frac{\log (\numobs / \delta)}{\numobs} \Big\}.
\end{align*}
Applying a mixed-norm chaining argument, we can bound the supremum of the empirical process over $h \in \shiftConvSet \cap \strucball (\rho)$ as
\begin{align*}
  \errtermsmpl (\rho) &\leq c \smoothness_\Fclass \sfactor T \Big\{  \frac{\dudley_2 (\shiftConvSet (\rho); \strucnorm{\cdot}{T} )  + \rho \sqrt{\log (1 / \delta)}}{\sqrt{\numobs K}} +  \frac{\dudley_1 (\convSet; \vecnorm{\cdot}{\holderspace^0} )  + \smoothness_\Fclass \log (\numobs / \delta)}{\numobs \sqrt{K}} \Big\},
\end{align*}
which proves \Cref{lemma:errterm-sampling-bound}.

\subsection{Proof of \Cref{prop:iterative-optimization-population}}\label{subsec:proof-iterative-optimization-population}
For any $\beta \in (0, 1)$ and $\ExpVal' \in \Fclass$, the optimality condition in the iteration step implies that
\begin{align*}
  &\strucnorm{\ExpVal^{(\kiter + 1)} - \ExpVal^{(\kiter)}}{T}^2 - 2 \stepsize B \big[ \ExpValTrue - \ExpVal^{(\kiter)}, \ExpVal^{(\kiter + 1)} - \ExpVal^{(\kiter)} \big]\\
   &\leq \strucnorm{ \big( (1 - \beta) \ExpVal^{(\kiter + 1)} + \beta \ExpVal' \big) - \ExpVal^{(\kiter)}}{T}^2 - 2 \stepsize B \big[ \ExpValTrue - \ExpVal^{(\kiter)}, \big( (1 - \beta) \ExpVal^{(\kiter + 1)} + \beta \ExpVal' \big) - \ExpVal^{(\kiter)} \big].
\end{align*}
Taking $\beta \rightarrow 0$ and $\ExpVal' = \fbar$, we arrive at the first-order optimality condition
\begin{align*}
  \strucinprod{\ExpVal^{(\kiter + 1)} - \ExpVal^{(\kiter)}}{\fbar - \ExpVal^{(\kiter + 1)}}{T} \geq  \stepsize B \big[ \ExpValTrue - \ExpVal^{(\kiter)}, \fbar - \ExpVal^{(\kiter + 1)} \big] .
\end{align*}
Note that $\fbar$ satisfies $0 \geq B [\fbar - \ExpValTrue, \fbar - \ExpVal^{(\kiter + 1)}] $. Adding this term (with a factor $\stepsize$) to both sides of the inequality, we have
\begin{align*}
  \strucinprod{\ExpVal^{(\kiter + 1)} - \ExpVal^{(\kiter)}}{\fbar - \ExpVal^{(\kiter + 1)}}{T} \geq \stepsize B \big[ \fbar - \ExpVal^{(\kiter)}, \fbar - \ExpVal^{(\kiter + 1)} \big] .
\end{align*}
Rearranging the terms, we have
\begin{align*}
  &\strucnorm{f^{(\kiter + 1)} - \fbar}{T}^2 \\
  &\leq \strucinprod{\fbar - \ExpVal^{(\kiter)}}{\fbar - \ExpVal^{(\kiter + 1)}}{T} - \stepsize B \big[ \fbar - \ExpVal^{(\kiter)}, \fbar - \ExpVal^{(\kiter + 1)} \big]\\
  &= \frac{1}{2} \strucnorm{\fbar - \ExpVal^{(\kiter)}}{T}^2 + \frac{1}{2} \strucnorm{\fbar - \ExpVal^{(\kiter + 1)}}{T}^2 - \frac{1}{2} \strucnorm{\ExpVal^{(\kiter)} - \ExpVal^{(\kiter + 1)}}{T}^2 - \stepsize B \big[ \fbar - \ExpVal^{(\kiter)}, \fbar - \ExpVal^{(\kiter + 1)} \big].
\end{align*}
Therefore, we arrive at the recursive bound
\begin{align*}
  &\strucnorm{f^{(\kiter + 1)} - \fbar}{T}^2 \\
  &\leq \strucnorm{\ExpVal^{(\kiter)} - \fbar}{T}^2  - \strucnorm{\ExpVal^{(\kiter)} - \ExpVal^{(\kiter + 1)}}{T}^2 - 2 \stepsize B \big[ \fbar - \ExpVal^{(\kiter)}, \fbar - \ExpVal^{(\kiter + 1)} \big]\\
  &= \strucnorm{\ExpVal^{(\kiter)} - \fbar}{T}^2 - \strucnorm{\ExpVal^{(\kiter)} - \ExpVal^{(\kiter + 1)}}{T}^2 - 2 \stepsize B \big[ \fbar - \ExpVal^{(\kiter)}, \fbar - \ExpVal^{(\kiter)} \big] + 2 \stepsize B \big[ \fbar - \ExpVal^{(\kiter)}, \ExpVal^{(\kiter + 1)} - \ExpVal^{(\kiter)} \big].
\end{align*}
By \Cref{lemma:parabolic-positive-definite}, we have
\begin{align*}
  B \big[ \fbar - \ExpVal^{(\kiter)}, \fbar - \ExpVal^{(\kiter)} \big] \geq \frac{\min (\sfactor, \lammin, 1)}{2} \strucnorm{\fbar - \ExpVal^{(\kiter)}}{T}^2.
\end{align*}
On the other hand, by \Cref{lemma:parabolic-op-upper-bound} and Assumption~\ref{assume:moment-comparison}, we have
\begin{align*}
  B \big[ \fbar - \ExpVal^{(\kiter)}, \ExpVal^{(\kiter + 1)} - \ExpVal^{(\kiter)} \big] &\leq c \max \Big\{ 1 + \sfactor, \tau \smoothness (1 + \smoothreg{8})  \Big\} \cdot \strucnorm{\ExpVal^{(\kiter + 1)} - \ExpVal^{(\kiter)}}{T} \cdot \strongnorm{\fbar - \ExpVal^{(\kiter)}}{T}\\
  &\leq \frac{1}{2 \stepsize} \strucnorm{\ExpVal^{(\kiter + 1)} - \ExpVal^{(\kiter)}}{T}^2 +  c^2 \stepsize \kappa^2 \big( 1 + \sfactor + \tau \smoothness (1 + \smoothreg{8})  \big)^2  \strucnorm{\fbar - \ExpVal^{(\kiter)}}{T}^2.
\end{align*}
Substituting these two bounds back to the recursive bound, we have
\begin{align*}
  \strucnorm{f^{(\kiter + 1)} - \fbar}{T}^2 \leq \Big\{1 - \min (\sfactor, \lammin, 1) \stepsize +  2c^2 \kappa^2 \big( 1 + \sfactor + \tau \smoothness (1 + \smoothreg{8})  \big)^2  \stepsize^2 \Big\} \strucnorm{\ExpVal^{(\kiter)} - \fbar}{T}^2.
\end{align*}
Given a stepsize $\stepsize$ satisfying
\begin{align}
  \stepsize \leq \stepsize_{\max} \mydefn \frac{ \min (\sfactor, \lammin, 1) }{4c^2 \kappa^2 \big( 1 + \sfactor + \tau \smoothness (1 + \smoothreg{8})  \big)^2  },\label{eq:stepsize-iterative-optimization-population}
\end{align}
we have
\begin{align*}
  \strucnorm{f^{(\kiter + 1)} - \fbar}{T}^2 \leq \Big\{1 - \frac{\min (\sfactor, \lammin, 1) }{2} \stepsize \Big\} \strucnorm{\ExpVal^{(\kiter)} - \fbar}{T}^2.
\end{align*}
Solving the recursion completes the proof of \Cref{prop:iterative-optimization-population}.

\subsection{Proof of \Cref{thm:iterative-optimization-empirical}}\label
{subsec:proof-iterative-optimization-empirical}
The proof follows the same lines as the proof of \Cref{prop:iterative-optimization-population}. By the empirical optimality condition, we have
\begin{align*}
  \widehat{\mathcal{E}}_\numobs \big(\ExpValHat^{(\kiter + 1)} - \ExpValHat^{(\kiter)}, \fbar - \ExpValHat^{(\kiter + 1)} \big) \geq  \stepsize \widehat{\mathcal{B}}_\numobs \big( \ExpValHat^{(\kiter)}, \fbar - \ExpValHat^{(\kiter + 1)} \big).
\end{align*}
Rearranging the terms similar to the proof of \Cref{prop:iterative-optimization-population}, we have
\begin{multline*}
  \widehat{\mathcal{E}}_\numobs \big(\ExpValHat^{(\kiter + 1)} - \fbar, \ExpValHat^{(\kiter + 1)} - \fbar \big) \leq \widehat{\mathcal{E}}_\numobs \big(\ExpValHat^{(\kiter)} - \fbar, \ExpValHat^{(\kiter)} - \fbar \big) - \widehat{\mathcal{E}}_\numobs \big(\ExpValHat^{(\kiter)} - \ExpValHat^{(\kiter + 1)}, \ExpValHat^{(\kiter)} - \ExpValHat^{(\kiter + 1)} \big)\\
   - 2 \stepsize \widehat{\mathcal{B}}_\numobs \big( \ExpValHat^{(\kiter)}, \fbar - \ExpValHat^{(\kiter)} \big) + 2 \stepsize \widehat{\mathcal{B}}_\numobs \big( \ExpValHat^{(\kiter)},  \ExpValHat^{(\kiter + 1)} -  \ExpValHat^{(\kiter)}\big).
\end{multline*}
In order to relate the empirical Sobolev norm $\widehat{\mathcal{E}}_\numobs$ to the population Sobolev norm $\strucnorm{\cdot}{T}$, we define the following empirical process supremum.
\begin{align*}
  \errtermsobo (\rho) \mydefn \sup_{h \in \Fclass_* \cap \strucball (\rho)} \abss{\widehat{\mathcal{E}}_\numobs (h, h) - \strucnorm{h}{T}^2 }, \quad \mbox{for any} \rho > 0.
\end{align*}
The following lemma provides high-probability bounds on the error term $\errtermsobo (\rho)$.
\begin{lemma}\label{lemma:empirical-sobolev-norm-convergence}
  Under the setup of \Cref{thm:iterative-optimization-empirical}, for any $\rho > 0$, we have
  \begin{align*}
     \errtermsobo (\rho) \leq \frac{c \tau^2 \rho}{\sqrt{\numobs}} \big( \dudley_2 (\Fclass^* (\rho); \strucnorm{\cdot}{T}) + \rho \sqrt{\log (1/ \delta)} \big)  + c \sigma \frac{\rho^\eta}{\numobs} \big( \dudley_1 (\Fclass ; \vecnorm{\cdot}{\holderspace^1}) + \smoothness_\Fclass \log (\numobs / \delta) \big).
  \end{align*}
\end{lemma}
\noindent See \Cref{subsubsec:proof-lemma-empirical-sobolev-norm-convergence} for the proof of this lemma.

We use $s_\delta (\rho)$ to denote the right-hand side of the bound in \Cref{lemma:empirical-sobolev-norm-convergence}. We define the fixed point $\rhofixtilde_{\mathrm{sobo}}$ as the the largest non-negative real $\rho$ satisfying $\rho^2 \leq \frac{128}{\min (\alpha, \lammin, 1) \stepsize} s_\delta (\rho)$.
Applying \Cref{lemma:empirical-sobolev-norm-convergence} with \Cref{lemma:random-radius-empirical-process}, for $\delta < \frac{1}{2 \log \numobs + 2 \log K}$, with probability at least $1 - 3 \delta$, we have
\begin{align*}
  \widehat{\mathcal{E}}_\numobs \big(\ExpValHat^{(\kiter + 1)} - \fbar, \ExpValHat^{(\kiter + 1)} - \fbar \big) &\geq \Big(1 - \frac{\min (\alpha, \lammin, 1) }{16} \stepsize \Big) \strucnorm{\ExpValHat^{(\kiter + 1)} - \fbar}{T}^2 + 3 \rhofixtilde_{\mathrm{sobo}}^2,\\
   \widehat{\mathcal{E}}_\numobs \big(\ExpValHat^{(\kiter)} - \fbar, \ExpValHat^{(\kiter)} - \fbar \big) &\leq  \Big(1 + \frac{\min (\alpha, \lammin, 1) }{16} \stepsize \Big) \strucnorm{\ExpValHat^{(\kiter)} - \fbar}{T}^2 + 3 \rhofixtilde_{\mathrm{sobo}}^2,\\
  \widehat{\mathcal{E}}_\numobs \big(\ExpValHat^{(\kiter)} - \ExpValHat^{(\kiter + 1)}, \ExpValHat^{(\kiter)} - \ExpValHat^{(\kiter + 1)} \big)  &\geq \frac{1}{2} \strucnorm{\ExpValHat^{(\kiter + 1)} -  \ExpValHat^{(\kiter)}}{T}^2 + 3 \rhofixtilde_{\mathrm{sobo}}^2.
\end{align*}
We denote by $\Event_\kiter^{(1)}$ the event that above inequalities hold true. By definition, it is easy to see that $\rhofixtilde_{\mathrm{sobo}} \leq \critradicross_{\numobs, \delta}$.

For the empirical bilinear forms, we note that
\begin{align*}
  &\abss{\widehat{\mathcal{B}}_\numobs \big( \ExpValHat^{(\kiter)},  \fbar - \ExpValHat^{(\kiter)} \big) - B \big[ \ExpValTrue - \ExpValHat^{(\kiter)},  \fbar - \ExpValHat^{(\kiter)} \big]}  \\
  &\leq \errtermmain \big( \strucnorm{\fbar - \ExpValHat^{(\kiter)}}{T} \big) + \errtermgen \big( \strucnorm{\fbar - \ExpValHat^{(\kiter)}}{T} \big) + \errtermsmpl \big( \strucnorm{\fbar - \ExpValHat^{(\kiter)}}{T} \big). 
\end{align*}
Applying \Cref{lemma:errterm-gen-bound,lemma:errterm-main-bound,lemma:errterm-sampling-bound} along with \Cref{lemma:random-radius-empirical-process}, following the arguments in the proof of \Cref{thm:main-error}, we have that with probability at least $1 - 3 \delta$,
\begin{multline}
   \errtermmain \big( \strucnorm{\fbar - \ExpValHat^{(\kiter)}}{T} \big) + \errtermgen \big( \strucnorm{\fbar - \ExpValHat^{(\kiter)}}{T} \big) + \errtermsmpl \big( \strucnorm{\fbar - \ExpValHat^{(\kiter)}}{T} \big) \\
   \leq \frac{3 \min (1, \lammin, \sfactor)}{32} \strucnorm{\fbar - \ExpValHat^{(\kiter)}}{T}^2
    + 3 \big(\critradistar_{\numobs, \delta} (\ExpValTrue - \fbar) \big)^2 + 3 (\critradicross_{\numobs, \delta})^2 + 3 (\critradismpl_{\numobs, K, \delta})^2.\label{eq:bilinear-form-1st-bound-in-iterative-optimization-empirical}
\end{multline}
And similarly, with probability $1 - 3 \delta$, we have
\begin{align}
  &\abss{\widehat{\mathcal{B}}_\numobs \big( \ExpValHat^{(\kiter)},  \ExpValHat^{(\kiter + 1)} - \ExpValHat^{(\kiter)} \big) - B \big[ \ExpValTrue - \ExpValHat^{(\kiter)},  \ExpValHat^{(\kiter + 1)} - \ExpValHat^{(\kiter)} \big]} \nonumber \\
  &\leq \errtermmain \big( \strucnorm{\ExpValHat^{(\kiter + 1)} - \ExpValHat^{(\kiter)}}{T} \big)  + \errtermsmpl \big( \strucnorm{\ExpValHat^{(\kiter + 1)} - \ExpValHat^{(\kiter)}}{T} \big)\nonumber \\
  &\qquad \qquad +  \errtermgen \Big( \max \big\{ \strucnorm{\ExpValHat^{(\kiter + 1)} - \ExpValHat^{(\kiter)}}{T}, \strucnorm{\fbar - \ExpValHat^{(\kiter)}}{T} \big\} \Big) \nonumber\\
  &\leq \frac{\min (1, \lammin, \sfactor)}{32} \Big\{ 3 \strucnorm{\ExpValHat^{(\kiter + 1)} - \ExpValHat^{(\kiter)}}{T}^2 + \strucnorm{\fbar - \ExpValHat^{(\kiter)}}{T}^2 \Big\} \nonumber \\
  &\qquad \qquad + 3 \big(\critradistar_{\numobs, \delta} (\ExpValTrue - \fbar) \big)^2 + 3 (\critradicross_{\numobs, \delta})^2 + 3 (\critradismpl_{\numobs, K, \delta})^2.\label{eq:bilinear-form-2nd-bound-in-iterative-optimization-empirical}
\end{align}
We use $\Event_\kiter^{(2)}$ to denote the event that \Cref{eq:bilinear-form-1st-bound-in-iterative-optimization-empirical,eq:bilinear-form-2nd-bound-in-iterative-optimization-empirical} hold true.

Combining these bounds, on the event $\Event_\kiter^{(1)} \cap \Event_\kiter^{(2)}$, we have
\begin{multline}
  \Big(1 - \frac{\min (\alpha, \lammin, 1) }{16} \stepsize \Big) \strucnorm{\ExpValHat^{(\kiter + 1)} - \fbar}{T}^2\\
   \leq  \Big(1 + \frac{3 \min (\alpha, \lammin, 1) }{16} \stepsize \Big) \strucnorm{\ExpValHat^{(\kiter)} - \fbar}{T}^2 - \frac{1}{4} \strucnorm{\ExpValHat^{(\kiter + 1)} -  \ExpValHat^{(\kiter)}}{T}^2 - 2 \stepsize B \big[ \ExpValTrue - \ExpValHat^{(\kiter)}, \fbar - \ExpValHat^{(\kiter + 1)} \big]\\
  + 6 \stepsize \Big\{ \big(\critradistar_{\numobs, \delta} (\ExpValTrue - \fbar) \big)^2 + (\critradicross_{\numobs, \delta})^2 +  (\critradismpl_{\numobs, K, \delta})^2 \Big\} + 9 (\critradicross_{\numobs, \delta})^2.\label{eq:near-finalized-recursion-iterative-optimization-empirical}
\end{multline}
For the population-level bilinear form $B$, we decompose it as
\begin{align*}
  B \big[ \ExpValTrue - \ExpValHat^{(\kiter)}, \fbar - \ExpValHat^{(\kiter + 1)} \big] = B \big[ \ExpValTrue - \fbar, \fbar - \ExpValHat^{(\kiter + 1)} \big] + B \big[ \fbar - \ExpValHat^{(\kiter)}, \fbar - \ExpValHat^{(\kiter)} \big] + B \big[ \fbar - \ExpValHat^{(\kiter)}, \ExpValHat^{(\kiter)} - \ExpValHat^{(\kiter + 1)} \big]
\end{align*}
Note that $\fbar$ solves the variational problem~\eqref{eq:orthogonality-condition-general}, leading to the inequality
\begin{align*}
  B [ \ExpValTrue - \fbar, \fbar - \ExpValHat^{(\kiter + 1)}] \geq 0.
\end{align*}
For the rest two terms, we use \Cref{lemma:parabolic-positive-definite} to obtain that
\begin{align*}
   B \big[ \fbar - \ExpValHat^{(\kiter)}, \fbar - \ExpValHat^{(\kiter)} \big] \geq \frac{\min (\sfactor, \lammin, 1)}{2} \strucnorm{\fbar - \ExpValHat^{(\kiter)}}{T}^2,
\end{align*}
and by \Cref{lemma:parabolic-op-upper-bound} and Assumption~\ref{assume:moment-comparison}, we have
\begin{align*}
  \abss{B \big[ \fbar - \ExpValHat^{(\kiter)}, \ExpValHat^{(\kiter + 1)} - \ExpValHat^{(\kiter)}\big]} &\leq c \max \Big\{ 1 + \sfactor, \tau \smoothness (1 + \smoothreg{8})  \Big\} \cdot \strucnorm{\ExpValHat^{(\kiter + 1)} - \ExpValHat^{(\kiter)}}{T} \cdot \strongnorm{\fbar - \ExpValHat^{(\kiter)}}{T}\\
  &\leq \frac{1}{8 \stepsize} \strucnorm{\ExpValHat^{(\kiter + 1)} - \ExpValHat^{(\kiter)}}{T}^2 +  4 c^2 \stepsize \kappa^2 \big( 1 + \sfactor + \tau \smoothness (1 + \smoothreg{8})  \big)^2  \strucnorm{\fbar - \ExpValHat^{(\kiter)}}{T}^2.
\end{align*}
Putting them together and substituting into the recursion~\eqref{eq:near-finalized-recursion-iterative-optimization-empirical}, for stepsize satisfying $\stepsize \leq \stepsize_{\max}$ with $\stepsize_{\max}$ defined in \Cref{eq:stepsize-iterative-optimization-population}, on the event $\Event_\kiter^{(1)} \cap \Event_\kiter^{(2)}$, we have
\begin{multline*}
  \strucnorm{\ExpValHat^{(\kiter + 1)} - \fbar}{T}^2 \leq \Big( 1 - \frac{\min (\sfactor, \lammin, 1)}{2} \stepsize \Big) \strucnorm{\ExpValHat^{(\kiter)} - \fbar}{T}^2\\
   + 6 \stepsize \Big\{  \big(\critradistar_{\numobs, \delta} (\ExpValTrue - \fbar) \big)^2 +  (\critradicross_{\numobs, \delta})^2 +  (\critradismpl_{\numobs, K, \delta})^2\Big\} + 9 (\critradicross_{\numobs, \delta})^2.
\end{multline*}
Therefore, on the event $\bigcap_{\kiter = 1}^{M} \Event_\kiter^{(1)} \cap \Event_\kiter^{(2)}$, we can solve the recursion to obtain
\begin{align*}
  \strucnorm{\ExpValHat^{(M)} - \fbar}{T}^2 &\leq \exp \Big( - \frac{\min (\sfactor, \lammin, 1)}{4} M \stepsize \Big) \strucnorm{\ExpValHat^{(0)} - \fbar}{T}^2\\
   &\qquad + \frac{12}{\min (\sfactor, \lammin, 1)}  \Big\{ \big(\critradistar_{\numobs, \delta} (\ExpValTrue - \fbar) \big)^2 +  (\critradicross_{\numobs, \delta})^2 +  (\critradismpl_{\numobs, K, \delta})^2 \Big\} + \frac{18}{\min (\sfactor, \lammin, 1) \stepsize} (\critradicross_{\numobs, \delta})^2.
\end{align*}
Note that such an event admits a lower bound for its probability
\begin{align*}
  \Prob \Big[ \bigcap_{\kiter = 1}^{M} \Event_\kiter^{(1)} \cap \Event_\kiter^{(2)} \Big] \geq 1 - 6 M \delta.
\end{align*}
Since the fixed point equations~\eqref{eq:defn-critical-radii} depend only logarithmically on $\delta$. For $\delta < \frac{1}{M^2}$, we can replace the failure probability $\delta$ by $\frac{\delta}{6M}$ at a cost of universal constant factors in the bound. We therefore conclude the proof of \Cref{thm:iterative-optimization-empirical}.

\subsubsection{Proof of \Cref{lemma:empirical-sobolev-norm-convergence}}
\label{subsubsec:proof-lemma-empirical-sobolev-norm-convergence}
Define the noise term
\begin{align*}
  \varsigma_i (h) \mydefn h (\State_T^{(i)})^2 + h (\State_0^{(i)})^2 + \frac{T}{K} \sum_{k = 1}^K \big( |h (\State_{\smpltime{k}{i}}^{(i)})|^2 + |\nabla h (\State_{\smpltime{k}{i}}^{(i)})|^2  \big).
\end{align*}
We can represent the empirical Sobolev norm as $\widehat{\mathcal{E}}_\numobs (h, h) = \frac{1}{\numobs} \sum_{i = 1}^\numobs \varsigma_i (h)$.For a pair of functions $h^{(1)}, h^{(2)} \in \strucball (\rho)$, we note that
\begin{align*}
  &\var \big(\varsigma_i (h^{(1)}) - \varsigma_i (h^{(2)}) \big) \\
  &\leq 4 \var \big( h^{(1)} (\State_T)^2 - h^{(2)} (\State_T)^2 \big) + 4 \var \big( h^{(1)} (\State_0)^2 - h^{(2)} (\State_0)^2 \big)\\
   &\qquad + 4 \var \Big( \int_0^T (h^{(1)} (\State_t)^2 - h^{(2)} (\State_t)^2) dt \Big) + \frac{4}{K} \int_0^T \var \big( h^{(1)} (\State_t)^2 - h^{(2)} (\State_t)^2 \big) dt\\
   &\qquad \qquad + 4 \var \Big( \int_0^T (|\nabla h^{(1)} (\State_t)|^2 -  | \nabla h^{(2)} (\State_t)|^2) dt \Big) + \frac{4}{K} \int_0^T \var \big( |\nabla h^{(1)} (\State_t)|^2 - |\nabla h^{(2)} (\State_t)|^2 \big) dt\\
   &\leq 4 \lpnorm{h^{(1)} - h^{(2)}}{4}{\density_T}^2  \lpnorm{h^{(1)} + h^{(2)}}{4}{\density_T}^2  + 4 \lpnorm{h^{(1)} - h^{(2)}}{4}{\density_0}^2 \lpnorm{h^{(1)} + h^{(2)}}{4}{\density_0}^2 \\
   &\qquad + 4 (T + 1) \int_0^T \sobokpnorm{h^{(1)} - h^{(2)}}{2}{4}{\density_t}^2 \cdot \sobokpnorm{h^{(1)} + h^{(2)}}{2}{4}{\density_t}^2 dt.
\end{align*}
Noting that $h^{(1)}, h^{(2)} \in \strucball (\rho)$, and invoking Assumption~\ref{assume:moment-comparison}, we have
\begin{align*}
  \var \big(\varsigma_i (h^{(1)}) - \varsigma_i (h^{(2)}) \big) \leq 16 \tau^4 (T + 3) \rho^2 \strucnorm{h^{(1)} - h^{(2)}}{T}^2.
\end{align*}
On the other hand, we can bound the Orlicz norm of the noise process as
\begin{align*}
  \vecnorm{\varsigma_i (h^{(1)}) - \varsigma_i (h^{(2)})}{\psi_1} &\leq \vecnorm{h^{(1)} - h^{(2)}}{\infty} \cdot \Big\{ \vecnorm{h^{(1)} (\State_T) + h^{(2)} (\State_T)}{\psi_1} + \vecnorm{h^{(1)} (\State_0) + h^{(2)} (\State_0)}{\psi_1} \Big\}\\
  &\qquad + T \vecnorm{h^{(1)} - h^{(2)}}{\holderspace^1}   \Big\{ \vecnorm{h^{(1)} (\State_{t_1}) + h^{(2)} (\State_{t_1})}{\psi_1} + \vecnorm{\nabla h^{(1)} (\State_{t_1}) + \nabla h^{(2)} (\State_{t_1})}{\psi_1}   \Big\}.\\
  &\leq 8 \sigma (1 + T)  \vecnorm{h^{(1)} - h^{(2)}}{\holderspace^1} \rho^\eta,
\end{align*}
where the last inequality comes from the sub-Gaussian class assumption~\ref{assume:subgaussian-class} and the fact that $h^{(1)}, h^{(2)} \in \Fclass^* \cap \strucball (\rho)$.

Once again by invoking Bernstein's inequality for unbounded random variables, for any $\delta \in (0, 1)$, with probability $1 - \delta$, we have
\begin{multline*}
  \abss{ \Big( \widehat{\mathcal{E}}_\numobs \big( h^{(1)}, h^{(1)} \big) - \strucnorm{h^{(1)}}{T}^2 \Big) - \Big( \widehat{\mathcal{E}}_\numobs \big( h^{(2)}, h^{(2)} \big) - \strucnorm{h^{(2)}}{T}^2 \Big) }\\
  \leq 4 \tau^2 \rho \sqrt{T + 3} \strucnorm{h^{(1)} - h^{(2)}}{T} \sqrt{\frac{\log (1 / \delta) }{ \numobs} } + 8 (1 + T) \sigma \rho^\eta \vecnorm{h^{(1)} - h^{(2)}}{\holderspace^1} \frac{\log (\numobs / \delta) }{\numobs}.
  \end{multline*}
Applying the mixed-norm chaining bound with this concentration inequality, we conclude that
\begin{align*}
  \errtermsobo (\rho) \leq \frac{c \tau^2 \rho}{\sqrt{\numobs}} \big( J_2 (\Fclass^* (\rho); \strucnorm{\cdot}{T}) + \rho \sqrt{\log (1/ \delta)} \big)  + c \sigma \frac{\rho^\eta}{\numobs} \big( \dudley_1 (\Fclass ; \vecnorm{\cdot}{\holderspace^1}) + \smoothness_\Fclass \log (\numobs / \delta) \big),
\end{align*}
with probability $1 - \delta$, which completes the proof of \Cref{lemma:empirical-sobolev-norm-convergence}.

\section{Discussion}\label{sec:discussion}
In this paper, we present a new class of algorithms for RL fine-tuning of controlled diffusion processes with function approximation. Our approach leverages the underlying structures of the HJB equation. By formulating the value learning problem as a variational inequality in Sobolev spaces, we obtain sharp oracle inequalities for both the value function and the induced policy. The statistical rates depend on the complexity and approximation properties of the function class, and exhibit a self-mitigating error phenomenon, leading to faster rates than standard regression. We also provide an efficient algorithm by iteratively solving regression problems, with exponential convergence to the statistical neighborhood. Our analysis reveals hidden structures that make RL fine-tuning as easy as supervised learning, with even better statistical rates.

Our analysis opens up new avenues for research in RL and diffusion processes.
\begin{itemize}
  \item First, under the framework of diffusion fine-tuning, it is interesting to develop practical algorithms that incorporate our algorithmic framework with practical neural network training. In particular, the Sobolev-norm proximal step in \Cref{alg:iterative-optimization-empirical} could be combined with modern deep learning techniques to improve performance.
  \item Additionally, our analysis techniques deviates from existing analysis of RL algorithms by exploiting the elliptic structures and energy functionals in the HJB equation. These techniques are of independent interest, and contribute to the design of ML-based PDE solvers in a more realistic setting. We conjecture that the insights gained from our analysis could be extended to general controlled Markov diffusions beyond the fine-tuning setting.
  \item Finally, moving beyond diffusion models, it is crucial to explore the potential of our approach in discrete-time and/or discrete-space processes that share similar underlying structures. In particular, it is important to investigate how the techniques developed in this paper can be adapted to the fine-tuning of auto-regressive language models.
\end{itemize}

\section*{Acknowledgments}
This work was partially supported by NSERC grant RGPIN-2024-05092 and a Connaught New Researcher Award to WM. WM thanks Jiequn Han, Wenpin Tang, Yuling Jiao, and Xiaofan Xia for helpful discussion.

\bibliographystyle{alpha}
\bibliography{references}

\appendix
\section{Proof of technical lemmas used in \Cref{thm:main-error}}
In this appendix, we collect the proofs of technical lemmas used in the proof of \Cref{thm:main-error}.

\subsection{Proof of \Cref{lemma:cross-cov-bound-diffusion}}\label{subsubsec:proof-lemma-cross-cov-bound-diffusion}
We note that
\begin{align*}
  \Exs \Big[ \abss{\int_0^T \generator_t f_t (\State_t) \cdot g_t (\State_t) dt}^2 \Big] = \int_0^T \int_0^T \Exs \big[ \generator_s f_s (\State_s) g_s (\State_s) \generator_t f_t (\State_t) g_t (\State_t) \big] ds dt.
\end{align*}
Denoting $ \Gamma (s, t) \mydefn \Exs \big[ \generator_s f_s (\State_s) g_s (\State_s) \generator_t f_t (\State_t) g_t (\State_t) \big]$, for $s < t$, we have
\begin{align}
  \Gamma (s, t) = \iint \generator_s f_s (x) g_s (x) \density_s (x) \generator_t f_t (y) g_t (y)  \density_{s, t} (y | x) dx dy, \label{eq:cross-covariance-term}
\end{align}
where the density $\density_{s, t} (\cdot | x)$ is the transition density of the uncontrolled process~\eqref{eq:uncontrolled-process} from time $s$ to $t$, starting from $x \in \real^\usedim$.

We use the following two bounds on the cross-covariance term $\Gamma (s, t)$.

\begin{lemma}\label{lemma:snapshot-cross-cov-bound-regularity}
  Under the setup of \Cref{lemma:cross-cov-bound-diffusion}, we have
  \begin{align*}
    \abss{\Gamma (s, t)} \leq \frac{\max (T, 1) C (\pq)}{|t - s|} \cdot \sobopstatnorm{g}{4 \pq}{\density_s} \cdot \sobopstatnorm{f}{4 \pq}{\density_s} \cdot \sobopstatnorm{g}{4 \pq}{\density_t} \cdot \sobopstatnorm{f}{4 \pq}{\density_t}
  \end{align*}
  for a constant $C (\pq)$ depending on the exponent $\pq$ and the regularity parameters in Assumptions~\ref{assume:moment-bound} and \ref{assume:density-regular}.
\end{lemma}
\noindent See \Cref{subsubsec:proof-lemma-snapshot-cross-cov-bound-regularity} for the proof of this lemma.

\Cref{lemma:snapshot-cross-cov-bound-regularity} is sharp when $\abss{t - s}$ is not too small. However, its integral blows up when $t$ and $s$ are close.
To resolve this issue, we simply use a na\"{i}ve bound on the cross-covariance term $\Gamma (s, t)$ when $t$ and $s$ are close.
\begin{align}
  |\Gamma (s, t)| &\leq \lpnorm{g_s}{4}{\density_s} \cdot \lpnorm{g_t}{4}{\density_t} \cdot \lpnorm{\generator_s f_s}{4}{\density_s} \cdot \lpnorm{\generator_t f_t}{4}{\density_t} \nonumber\\
  &\leq \smoothness_{\frac{4 \pq}{\pq - 1}}^2 \lpnorm{g_s}{4 \pq}{\density_s} \cdot \lpnorm{g_t}{4 \pq}{\density_t} \cdot \sobokpnorm{f_s}{2}{4 \pq}{\density_s} \cdot \sobokpnorm{ f_t}{2}{4 \pq}{\density_t}.\label{eq:cross-covariance-naive-bound}
\end{align}
Substituting \Cref{lemma:snapshot-cross-cov-bound-regularity} and \Cref{eq:cross-covariance-naive-bound} back to \Cref{eq:cross-covariance-term}, we have
\begin{align*}
  &\Exs \Big[ \abss{\int_0^T \generator_t f_t (\State_t) \cdot g_t (\State_t) dt}^2 \Big]\\
  & \leq \max (T, 1) C (\pq) \cdot \iint_{\substack{s, t \in [0, T]^2\\
  |s - t| \geq \delta}} \sobopstatnorm{g}{4 \pq}{\density_s} \cdot \sobopstatnorm{f}{4 \pq}{\density_s} \cdot \sobopstatnorm{g}{4 \pq}{\density_t} \cdot \sobopstatnorm{f}{4 \pq}{\density_t}  \cdot  \frac{1}{|t - s|} ds dt \\
  &\qquad \qquad  + \smoothness_{\frac{4 \pq}{\pq - 1}}^2 \cdot \iint_{\substack{s, t \in [0, T]^2\\
  |s - t| \geq \delta}} \sobopstatnorm{g}{4 \pq}{\density_s} \cdot \sobokpnorm{f}{2}{4 \pq}{\density_s} \cdot \sobopstatnorm{g}{4 \pq}{\density_t} \cdot \sobokpnorm{f}{2}{4 \pq}{\density_t}  ds dt.
\end{align*}
For the first term, we apply the Cauchy--Schwarz inequality to obtain
\begin{align*}
  &\iint_{\substack{s, t \in [0, T]^2\\
  |s - t| \geq \delta}} \sobopstatnorm{g_s}{4 \pq}{\density_s} \cdot \sobopstatnorm{f_s}{4 \pq}{\density_s} \cdot \sobopstatnorm{g_t}{4 \pq}{\density_t} \cdot \sobopstatnorm{f_t}{4 \pq}{\density_t}  \cdot  \frac{1}{|t - s|} ds dt \\
  &\leq \Big\{ \iint_{\substack{s, t \in [0, T]^2\\
  |s - t| \geq \delta}} \frac{\sobopstatnorm{g_s}{4 \pq}{\density_s}^2 \cdot \sobopstatnorm{f_s}{4 \pq}{\density_s}^2  }{|t - s|} ds dt \Big\}^{1/2} \Big\{ \iint_{\substack{s, t \in [0, T]^2\\
  |s - t| \geq \delta}} \frac{\sobopstatnorm{g_t}{4 \pq}{\density_t}^2 \cdot \sobopstatnorm{f_t}{4 \pq}{\density_t}^2  }{|t - s|} ds dt \Big\}^{1/2}\\
  &\leq \log (T / \delta) \int_0^T \sobopstatnorm{g_t}{4 \pq}{\density_t}^2 \cdot \sobopstatnorm{f_t}{4 \pq}{\density_t}^2 dt\\
  &\leq \log (T / \delta) \Big\{ \int_0^T \sobopstatnorm{g_t}{4 \pq}{\density_t}^4 dt \Big\}^{1/2} \cdot \Big\{ \int_0^T \sobopstatnorm{f_t}{4 \pq}{\density_t}^4 dt \Big\}^{1/2}\\
  &\leq \log (T / \delta) \energynorm{f}{1}{4 \pq}{T}^2 \energynorm{g}{1}{4 \pq}{T}^2.
\end{align*}
Similarly, for the second term, we have
\begin{align*}
  &\iint_{\substack{s, t \in [0, T]^2\\
  |s - t| \geq \delta}} \sobopstatnorm{g}{4 \pq}{\density_s} \cdot \sobokpnorm{f}{2}{4 \pq}{\density_s} \cdot \sobopstatnorm{g}{4 \pq}{\density_t} \cdot \sobokpnorm{f}{2}{4 \pq}{\density_t}  ds dt\\
  &\leq \Big\{ \iint_{\substack{s, t \in [0, T]^2\\
  |s - t| \leq \delta}} \sobopstatnorm{g_s}{4 \pq}{\density_s}^2 \cdot \sobopstatnorm{f_s}{4 \pq}{\density_s}^2   ds dt \Big\}^{1/2} \Big\{ \iint_{\substack{s, t \in [0, T]^2\\
  |s - t| \leq \delta}} \sobopstatnorm{g_t}{4 \pq}{\density_t}^2 \cdot \sobopstatnorm{f_t}{4 \pq}{\density_t}^2  ds dt \Big\}^{1/2}\\
  &\leq \delta \energynorm{f}{2}{4 \pq}{T}^2 \energynorm{g}{0}{4 \pq}{T}^2.
\end{align*}
Putting them together, we complete the proof of \Cref{lemma:cross-cov-bound-diffusion}.

\subsubsection{Proof of \Cref{lemma:snapshot-cross-cov-bound-regularity}}\label{subsubsec:proof-lemma-snapshot-cross-cov-bound-regularity}
Recall that $\generator_t = \inprod{\drift_t}{\nabla} + \frac{1}{2} \inprod{\covMat_t}{\nabla^2}$. We follow Lemma 5 of the paper~\cite{mou2025statistical} to obtain the identity.
\begin{align*}
  \Gamma (s, t) &= \Exs \Bigg[ \Big\{h_s (\State_s)^\top \big(\nabla f_s + f_s  \nabla \log \density_s \big) (\State_s) + \ell_s (\State_s)  f_s (\State_s)  \Big\}\\
  &\qquad \qquad \cdot \Big\{ h_t (\State_t)^\top \big( \nabla f_t (\State_t) + f_t (\State_t) \nabla_y \log p_{s, t} (\State_t \mid \State_s) \big) +  \ell_t (\State_t) f_t (\State_t) \Big\} \Bigg] \\
  &\quad +\Exs \Bigg[ f_s (\State_s) h_s (\State_s)^\top \Big\{\nabla_x \log p_{s, t} (\State_t| \State_s) \nabla f_t (\State_t)^\top + f_t (\State_t) \frac{\nabla_x \nabla_y^\top p_{s, t} (\State_t | \State_s)}{p_{s, t} (\State_t | \State_s)}\Big\} h_t (\State_t) \Bigg]\\
  &\quad \qquad \qquad + \Exs \Big[ f_s (\State_s) f_t (\State_t) \ell_t (\State_t) h_s (\State_s)^\top \nabla \log_x p_{s, t} (\State_t | \State_s)  \Big]\\
  &=: E_1 + E_2 + E_3.
\end{align*}
where we define the functions
\begin{align*}
  h_t (x) = g_t (x) \drift_t (x) + \frac{1}{2} \covMat_t (x) \nabla g_t (x), \quad \mbox{and} \quad \ell_t (x) = g_t (x) \nabla \cdot \drift_t (x) + \frac{1}{2} \big( \nabla \cdot \covMat_t (x) \big)^\top \nabla g_t (x),
\end{align*}
for $t \in [0, T]$ and $x \in \real^\usedim$.

Now we bound the three terms $E_1$, $E_2$, and $E_3$ separately.

\paragraph{Upper bounds on $E_1$:}

By Cauchy--Schwarz inequality, we have the bounds
\begin{multline*}
  |E_1| \leq \Exs \Bigg[ \Big\{h_s (\State_s)^\top \big(\nabla f_s + f_s  \nabla \log \density_s \big) (\State_s) + \ell_s (\State_s)  f_s (\State_s)  \Big\}^2 \Bigg]^{1/2}\\
  \cdot \Exs \Bigg[ \Big\{ h_t (\State_t)^\top \big( \nabla f_t (\State_t) + f_t (\State_t) \nabla_y \log p_{s, t} (\State_t \mid \State_s) \big) +  \ell_t (\State_t) f_t (\State_t) \Big\}^2 \Bigg]^{1/2}.
\end{multline*}
By further applying H\"{o}lder's inequality along with Assumption~\ref{assume:moment-bound} and \ref{assume:density-regular}, we have
\begin{align*}
  &\Exs \Bigg[ \Big\{h_s (\State_s)^\top \big(\nabla f_s + f_s  \nabla \log \density_s \big) (\State_s) + \ell_s (\State_s)  f_s (\State_s)  \Big\}^2 \Bigg]\\
  &\leq \Big( \lpnorm{h_s}{4}{\density_s}^2 + \lpnorm{\ell_s}{4}{\density_s}^2 \Big) \cdot \Big( \lpnorm{\nabla f_s + f_s \nabla \log \density_s}{4}{\density_s}^2 + \lpnorm{f_s}{4}{\density_s}^2 \Big)\\
  &\leq \Big( \lpnorm{h_s}{4}{\density_s}^2 + \lpnorm{\ell_s}{4}{\density_s}^2 \Big) \cdot \Big( 1 + \lpnorm{\nabla \log \density_s}{\frac{4 \pq}{\pq - 1}}{\density_s}^2 \Big) \sobopstatnorm{f_s}{\pq}{\density_s}^2\\
  &\leq 4 \smoothness_{\frac{4 \pq}{\pq - 1}}^2 \Big\{ 1 + \smoothreg{\frac{4 \pq}{\pq - 1}} \Big\}^2 \sobopstatnorm{g_s}{4 \pq}{\density_s}^2 \cdot \sobopstatnorm{f_s}{4 \pq}{\density_s}^2,
\end{align*}
and similarly,
\begin{align*}
  &\Exs \Bigg[ \Big\{ h_t (\State_t)^\top \big( \nabla f_t (\State_t) + f_t (\State_t) \nabla_y \log p_{s, t} (\State_t \mid \State_s) \big) +  \ell_t (\State_t) f_t (\State_t) \Big\}^2 \Bigg]\\
  &\leq \Big( \lpnorm{h_t}{4}{\density_t}^2 + \lpnorm{\ell_t}{4}{\density_t}^2  \Big) \cdot \Big( \Exs \big[ \eucnorm{\nabla f_t (\State_t) + f_t (\State_t) \nabla \log \density_{s, t} (\State_t \mid \State_s)}^4 \big]^{1/2} + \lpnorm{f_t}{4}{\density_t}^2 \Big)\\
  &\leq 4   \smoothness_{\frac{4 \pq}{\pq - 1}}^2 \Big\{ 1 + \frac{1}{\sqrt{t - s}}\smoothreg{\frac{4 \pq}{\pq - 1}} \Big\}^2 \sobopstatnorm{g_t}{4 \pq}{\density_t}^2 \cdot \sobopstatnorm{f_t}{4 \pq}{\density_t}^2.
\end{align*}
Putting them together, we have
\begin{align*}
  |E_1| \leq \frac{16 \sqrt{T}}{\sqrt{t - s}} \smoothness_{\frac{4 \pq}{\pq - 1}}^2 \smoothreg{\frac{4 \pq}{\pq - 1}} \cdot \sobopstatnorm{g}{4 \pq}{\density_s} \cdot \sobopstatnorm{f}{4 \pq}{\density_s} \cdot \sobopstatnorm{g}{4 \pq}{\density_t} \cdot \sobopstatnorm{f}{4 \pq}{\density_t}.
\end{align*}

\paragraph{Upper bounds on $E_2$:}
Similar to the bound on $E_1$, we apply Cauchy--Schwarz inequality to obtain
\begin{align*}
  |E_2| \leq \Exs \Big[ \eucnorm{f_s (\State_s) h_s (\State_s)}^2 \Big]^{1/2} \cdot \Exs \Bigg[ \eucnorm{ \Big\{\nabla_x \log p_t (\State_t| \State_s) \nabla f_t (\State_t)^\top + f_t (\State_t) \frac{\nabla_x \nabla_y^\top \density_{s, t} (\State_t | \State_s)}{\density_{s, t} (\State_t | \State_s)}\Big\} h_t (\State_t)}^2 \Bigg]^{1/2}.
\end{align*}
Invoking H\"{o}lder's inequality, we have
\begin{align*}
  \Exs \Big[ \eucnorm{f_s (\State_s) h_s (\State_s)}^2 \Big] &\leq \lpnorm{f_s}{4}{\density_s}^2 \cdot \lpnorm{h_s}{4}{\density_s}^2 \leq 4 \smoothness_{\frac{4 \pq}{\pq - 1}}^2 \sobopstatnorm{g_s}{4 \pq}{\density_s}^2 \cdot  \sobopstatnorm{f_s}{4 \pq}{\density_s}^2,
\end{align*}
and we have
\begin{align*}
  &\Exs \Bigg[ \eucnorm{ \Big\{\nabla_x \log p_t (\State_t| \State_s) \nabla f_t (\State_t)^\top + f_t (\State_t) \frac{\nabla_x \nabla_y^\top p_{s, t} (\State_t | \State_s)}{\density_{s, t} (\State_t | \State_s)}\Big\} h_t (\State_t)}^2 \Bigg]\\
  &\leq \lpnorm{h_t}{4}{\density_t}^2 \cdot \Exs \Bigg[  \eucnorm{\nabla_x \log p_t (\State_t| \State_s) \nabla f_t (\State_t)^\top + f_t (\State_t) \frac{\nabla_x \nabla_y^\top p_{s, t} (\State_t | \State_s)}{\density_{s, t} (\State_t | \State_s)}}^4 \Bigg]^{1/2}\\
  &\leq 4 \smoothness_{\frac{4 \pq}{\pq - 1}}^2  \sobopstatnorm{g_t}{4 \pq}{\density_t}^2 \cdot \sobopstatnorm{f_t}{4 \pq}{\density_t}^2 \cdot \Big\{  \Exs \big[ \eucnorm{\nabla_x \log p_{s, t} (\State_t| \State_s)}^{\frac{4 \pq}{\pq - 1}} \big]^{\frac{\pq - 1}{2 \pq}} + \Exs \Big[ \eucnorm{\frac{\nabla_x p_{s, t} }{p_{s, t}}(\State_t| \State_s)}^{\frac{4 \pq}{\pq - 1}} \Big]^{\frac{\pq - 1}{2 \pq}} \Big\}\\
  &\leq  4 \smoothness_{\frac{4 \pq}{\pq - 1}}^2 \smoothreg{\frac{4 \pq}{\pq - 1}} \Big\{ \frac{1}{\sqrt{t - s}} + \frac{1}{t - s} \Big\}^2 \sobopstatnorm{g_t}{4 \pq}{\density_t}^2 \cdot \sobopstatnorm{f_t}{4 \pq}{\density_t}^2.
\end{align*}
Putting them together, we have
\begin{align*}
  \abss{E_2} \leq \frac{16 T}{t - s} \smoothness_{\frac{4 \pq}{\pq - 1}}^2 \smoothreg{\frac{4 \pq}{\pq - 1}}^2 \cdot \sobopstatnorm{g}{4 \pq}{\density_s} \cdot \sobopstatnorm{f}{4 \pq}{\density_s} \cdot \sobopstatnorm{g}{4 \pq}{\density_t} \cdot \sobopstatnorm{f}{4 \pq}{\density_t}.
\end{align*}
\paragraph{Upper bounds for $E_3$:} Invoking H\"{o}lder's inequality, we have
\begin{align*}
  |E_3| &\leq \lpnorm{f_s}{4}{\density_s} \cdot \lpnorm{h_s}{4}{\density_s} \cdot \lpnorm{\ell_t}{4}{\density_t} \cdot  \Exs \Big[ \eucnorm{f_t \nabla \log \density_{s,t} (\State_t | \State_s)}^4 \Big]^{1/4}\\
  &\leq 4 \smoothness_{\frac{4 \pq}{\pq - 1}}^2 \smoothreg{\frac{4 \pq}{4 \pq - 1}} \frac{1}{\sqrt{t - s}} \lpnorm{f_s}{4}{\density_s}  \cdot \lpnorm{g_s}{4 \pq}{\density_s}  \cdot \lpnorm{g_t}{4 \pq}{\density_t} \cdot \lpnorm{f_t}{4 \pq}{\density_t}.
\end{align*}

\subsection{Proof of \Cref{lemma:time-derivative-cross-var-bound}}\label{subsubsec:proof-lemma-time-derivative-cross-var-bound}
By Cauchy--Schwarz inequality, we have
\begin{align*}
   \Exs \Big[ \abss{\int_0^T \partial_t f_t (\State_t) g_t (\State_t) dt }^2 \Big] \leq \Big\{ \int_0^T \Exs \big[\partial_t f_t^4 (\State_t) \big]  dt \Big\}^{1/2}  \Big\{ \int_0^T \Exs \big[ \abss{ g_t (\State_t)}^4 \big] dt \Big\}^{1/2} \leq  \dynanorm{f}{1}{4}{T}^2 \energynorm{g}{1}{4}{T}^2.
\end{align*}
On the other hand, by applying the integration by parts formula to the time integral, we have
\begin{align*}
  \int_0^T \partial_t f_t (\State_t) g_t (\State_t) dt =   f_T (\State_T) g_T (\State_T) - f_0 (\State_0) g_0 (\State_0) - \int_0^T f_t (\State_t) \partial_t g_t (\State_t) dt.
\end{align*}
By Young's inequality and Cauchy--Schwarz inequality, we have
\begin{align*}
  &\Exs \Big[ \abss{\int_0^T \partial_t f_t (\State_t) g_t (\State_t) dt}^2 \Big]\\
   &\leq   3 \Exs \big[ f_T^2 (\State_T) g_T^2 (\State_T) \big] +  3 \Exs \big[ f_0^2 (\State_0) g_0^2 (\State_0) \big] + 3 \Exs \Big[ \abss{\int_0^T f_t (\State_t) \partial_t g_t (\State_t) dt}^2 \Big]\\
  &\leq 3 \lpnorm{f_T}{4}{\density_T}^2  \lpnorm{g_T}{4}{\density_T}^2 + 3 \lpnorm{f_0}{4}{\density_0}^2 \lpnorm{g_0}{4}{\density_0}^2 + 3 \Big\{ \int_0^T \Exs \big[f_t^4 (\State_t) \big]  dt \Big\}^{1/2} \cdot \Big\{ \int_0^T \Exs \big[ \abss{\partial_t g_t (\State_t)}^4 \big] dt \Big\}^{1/2}\\
  &\leq 9 \energynorm{f}{1}{4}{T}^2 \cdot \dynanorm{g}{1}{4}{T}^2.
\end{align*}
Combining the two bounds yield the conclusion of \Cref{lemma:time-derivative-cross-var-bound}.

\subsection{Proof of \Cref{lemma:discrete-sampling-in-xi-var-bound}}\label{subsec:proof-lemma-discrete-sampling-in-xi-var-bound}
To start with, since the discrete sampling points $\smpltime{1}{i}, \smpltime{2}{i}, \cdots, \smpltime{K}{i}$ are $\mathrm{i.i.d.}$ uniform for each $i \in [\numobs]$, we note the error decomposition
\begin{align*}
  &\var \Big( \xi_i (f^{(1)}, g^{(1)}) - \xi_i (f^{(2)}, g^{(2)}) \mid (\State_t^{(i)})_{t \in [0, T]} \Big)\\
  &= \frac{T^2}{K} \var \Big\{ \big( \partial_t + \generator_{t_k} + \sfactor \Reward_{t_k} \big)  f_{t_k}^{(1)} (\MyState_{t_k})   g_{t_k}^{(1)} (\MyState_{t_k}) -  \big( \partial_t + \generator_{t_k} + \sfactor \Reward_{t_k} \big)  f_{t_k}^{(2)} (\MyState_{t_k})    g_{t_k}^{(2)} (\MyState_{t_k}) \mid (\State_t)_{t \in [0, T]} \Big\}\\
  &= \frac{3T}{K} \int_0^T \Big\{ (\partial_t f^{(1)}_t \cdot g^{(1)}_t) (\State_t) - (\partial_t f^{(2)}_t \cdot g^{(2)}_t) (\State_t) \Big\}^2 dt  + \frac{3T}{K} \int_0^T \Big\{ (\generator_t f^{(1)}_t \cdot g^{(1)}_t) (\State_t) - (\generator_t f^{(2)}_t \cdot g^{(2)}_t) (\State_t) \Big\}^2 dt\\
  &\qquad + \frac{12 T}{K} \int_0^T \Big\{ ( f^{(1)}_t \cdot g^{(1)}_t) (\State_t) - (f^{(2)}_t \cdot g^{(2)}_t) (\State_t) \Big\}^2 dt.
\end{align*}
By Cauchy--Schwarz inequality, it is easy to see that
\begin{align*}
  &\Exs \Big[ \int_0^T \Big\{ (\partial_t f^{(1)}_t \cdot g^{(1)}_t) (\State_t) - (\partial_t f^{(2)}_t \cdot g^{(2)}_t) (\State_t) \Big\}^2 dt \Big]\\
  &\leq 2 \Exs \Big[ \int_0^T \abss{ \big(\partial_t f^{(1)}_t \cdot (g^{(1)}_t - g^{(2)}_t ) \big) (\State_t)}^2 dt \Big] + 2 \Exs \Big[ \int_0^T \abss{ \big( (\partial_t f^{(1)}_t - \partial_t f^{(2)}_t ) \cdot g^{(2)}_t \big) (\State_t) }^2 dt \Big]\\
  &\leq 2 \dynanorm{f^{(1)}}{0}{4}{T}^2  \energynorm{g^{(1)} - g^{(2)}}{0}{4}{T}^2 + 2 \dynanorm{f^{(1)} - f^{(2)}}{0}{4}{T}^2  \energynorm{ g^{(2)}}{0}{4}{T}^2.
\end{align*}
Similarly, we have
\begin{multline*}
  \Exs \Big[ \int_0^T \Big\{ ( f^{(1)}_t g^{(1)}_t) (\State_t) - (f^{(2)}_t  g^{(2)}_t) (\State_t) \Big\}^2 dt \Big]\\
   \leq 2 \energynorm{f^{(1)}}{0}{4}{T}^2  \energynorm{g^{(1)} - g^{(2)}}{0}{4}{T}^2 + 2 \energynorm{f^{(1)} - f^{(2)}}{0}{4}{T}^2  \energynorm{ g^{(2)}}{0}{4}{T}^2.
\end{multline*}
As for the diffusion generator, we note that
\begin{align*}
  &\int_0^T \Big\{ \generator_t  f_t^{(1)} (\MyState_t) g_t^{(1)} (\MyState_t) -  \generator_t f_t^{(2)} (\State_t) g_t^{(2)} (\State_t) \Big\}^2 dt\\
  &\leq 2 \int_0^T  \abss{\generator_t \big( f_t^{(1)} - f_t^{(2)} \big) (\State_t) g_t^{(2)} (\State_t)}^2 dt + 2 \int_0^T \abss{  \big( g_t^{(1)} - g_t^{(2)} \big) (\State_t) \generator_t f_t^{(1)} (\State_t)}^2 dt.
\end{align*}
Under Assumption~\ref{assume:moment-bound}, for a pair $f, g$ of functions, Cauchy--Schwarz inequality yields
\begin{align*}
  \Exs \big[ \abss{\generator_t f_t (\State_t) g_t (\State_t)}^2 \big] \leq \lpnorm{\generator_t f_t}{4}{\density_t}^2 \cdot \lpnorm{g_t}{4}{\density_t}^2 \leq \smoothness_{\frac{4 \pq}{\pq - 1}}^2 \sobokpnorm{f_t}{2}{4 \pq}{\density_t}^2 \cdot \lpnorm{g_t}{4 \pq}{\density_t}^2.
\end{align*}
Thus, we have
\begin{align*}
  &\Exs \Big[ \int_0^T \Big\{ \generator_t  f_t^{(1)} (\MyState_t) g_t^{(1)} (\MyState_t) -  \generator_t f_t^{(2)} (\State_t) g_t^{(2)} (\State_t) \Big\}^2 dt \Big] \\
  &\leq 2 \smoothness_{\frac{4 \pq}{\pq - 1}}^2\Big\{  \int_0^T \sobokpnorm{f_t^{(1)}}{2}{4 \pq}{\density_t}^2 \cdot \lpnorm{g_t^{(1)} - g_t^{(2)}}{4 \pq}{\density_t}^2 dt + \int_0^T \sobokpnorm{f_t^{(1)} - f_t^{(2)}}{2}{4 \pq}{\density_t}^2 \cdot \lpnorm{g_t^{(2)}}{4 \pq}{\density_t}^2 dt \Big\}\\
  &\leq 2 \smoothness_{\frac{4 \pq}{\pq - 1}}^2 \Big\{ \energynorm{f^{(1)}}{2}{4 \pq}{T}^2 \cdot \energynorm{g^{(1)} - g^{(2)}}{0}{4 \pq}{T}^2 + \energynorm{f^{(1)} - f^{(2)}}{2}{4 \pq}{T}^2 \cdot \energynorm{g^{(2)}}{0}{4 \pq}{T}^2 \Big\}.
\end{align*}
Putting them together, we conclude the proof of \Cref{lemma:discrete-sampling-in-xi-var-bound}.

\subsection{Proof of \Cref{lemma:quasi-sub-additive}}\label{app:subsec-proof-quasi-subadditive}
Recall the definition of Dudley integral
\begin{align*}
  \dudley_\gamma \big( \shiftConvSet \cap \ball_X (\rho); \vecnorm{\cdot}{Y} \big) = \int_0^{+ \infty} \log^\gamma N \big( \shiftConvSet \cap \ball_X (\rho); \vecnorm{\cdot}{Y}, \varepsilon \big) d \varepsilon.
\end{align*}
Given $\rho' > \rho > 0$, by convexity of the class $\Fclass$, we note that
\begin{align*}
    \shiftConvSet \cap \ball_X (\rho') \subseteq  \frac{\rho'}{\rho} \shiftConvSet \cap \ball_X (\rho) ,
\end{align*}
and consequently,
\begin{align*}
   N \big( \shiftConvSet \cap \ball_X (\rho'); \vecnorm{\cdot}{Y}, \varepsilon \big) \leq  N \big( \shiftConvSet \cap \ball_X (\rho); \vecnorm{\cdot}{Y}, \frac{\rho}{\rho'} \varepsilon \big). 
\end{align*}
Applying change-of-variable formula with $t = \frac{\rho}{\rho'} \varepsilon$, we have
\begin{align}
   \dudley_\gamma \big( \shiftConvSet \cap \ball_X (\rho'); \vecnorm{\cdot}{Y} \big) &= 
  \int_0^{+ \infty} \log^\gamma N \big( \shiftConvSet \cap \ball_X (\rho'); \vecnorm{\cdot}{Y}, \varepsilon \big) d \varepsilon \nonumber \\
  &\leq  \int_0^{+ \infty} \log^\gamma N \big( \shiftConvSet \cap \ball_X (\rho); \vecnorm{\cdot}{Y}, \frac{\rho'}{\rho} \varepsilon \big) d \varepsilon \nonumber\\
 & = \frac{\rho'}{\rho} \int_0^{+ \infty} \log^\gamma N \big( \shiftConvSet \cap \ball_X (\rho); \vecnorm{\cdot}{Y}, t \big) d t = \frac{\rho'}{\rho}  \dudley_\gamma \big( \shiftConvSet \cap \ball_X (\rho); \vecnorm{\cdot}{Y} \big).\label{eq:ratio-bound-for-regular-function-proof}
\end{align}
Now for a pair $\rho_1, \rho_2 > 0$, applying \Cref{eq:ratio-bound-for-regular-function-proof}, we have
\begin{align*}
  \dudley_\gamma \big( \shiftConvSet \cap \ball_X (\rho_1 + \rho_2); \vecnorm{\cdot}{Y} \big) \leq \frac{\rho_1 + \rho_2}{\rho_1}  \dudley_\gamma \big( \shiftConvSet \cap \ball_X (\rho_1); \vecnorm{\cdot}{Y} \big),\\
  \dudley_\gamma \big( \shiftConvSet \cap \ball_X (\rho_1 + \rho_2); \vecnorm{\cdot}{Y} \big) \leq \frac{\rho_1 + \rho_2}{\rho_2}  \dudley_\gamma \big( \shiftConvSet \cap \ball_X (\rho_2); \vecnorm{\cdot}{Y} \big).
\end{align*}
Taking the weighted average of these two inequalities, we have
\begin{align*}
   \dudley_\gamma \big( \shiftConvSet \cap \ball_X (\rho_1 + \rho_2); \vecnorm{\cdot}{Y} \big)  \leq \dudley_\gamma \big( \shiftConvSet \cap \ball_X (\rho_1); \vecnorm{\cdot}{Y} \big) + \dudley_\gamma \big( \shiftConvSet \cap \ball_X (\rho_2); \vecnorm{\cdot}{Y} \big).
\end{align*}
And consequently, for $c \in [0, 1]$, we have
\begin{align*}
  h (\rho_1) + h (\rho_2) &= \rho_1^c \dudley_\gamma \big( \shiftConvSet \cap \ball_X (\rho_1); \vecnorm{\cdot}{Y} \big) + \rho_2^c \dudley_\gamma \big( \shiftConvSet \cap \ball_X (\rho_2); \vecnorm{\cdot}{Y} \big)\\
  &\overset{(i)}{\geq} \frac{\rho_1^c + \rho_2^c}{2} \Big\{ \dudley_\gamma \big( \shiftConvSet \cap \ball_X (\rho_1); \vecnorm{\cdot}{Y} \big) + \dudley_\gamma \big( \shiftConvSet \cap \ball_X (\rho_2); \vecnorm{\cdot}{Y} \big) \Big\}\\
  &\geq \frac{\rho_1^c + \rho_2^c}{2} \dudley_\gamma \big( \shiftConvSet \cap \ball_X (\rho_1 + \rho_2); \vecnorm{\cdot}{Y} \big)\\
  &\geq \frac{(\rho_1 + \rho_2)^c }{2} \dudley_\gamma \big( \shiftConvSet \cap \ball_X (\rho_1 + \rho_2); \vecnorm{\cdot}{Y} \big) = \frac{1}{2} h (\rho_1 + \rho_2),
\end{align*}
where in step $(i)$, we use the fact that $\rho \mapsto \dudley_\gamma \big( \shiftConvSet \cap \ball_X (\rho); \vecnorm{\cdot}{Y} \big)$ is a non-decreasing function, so that the weighted average is larger than the average, under a monotonic weight. Additionally, for any $a > 1$, by \Cref{eq:ratio-bound-for-regular-function-proof}, it is easy to see that
\begin{align*}
  h (a \rho) = a^c \dudley_\gamma \big( \shiftConvSet \cap \ball_X (a \rho); \vecnorm{\cdot}{Y} \big) \leq a^c \frac{a}{\rho}  \dudley_\gamma \big( \shiftConvSet \cap \ball_X (\rho); \vecnorm{\cdot}{Y} \big) = a^{c + 1} h (\rho) \leq a^2 h (\rho).
\end{align*} 
This proves that the function $h$ is sub-quadratic.

Suppose that $h_1$ and $h_2$ are both sub-quadratic functions. Let $h = a_1 h_1 + a_2 h_2$ for some $a_1, a_2 > 0$. Then for any $x, y > 0$, we have
\begin{align*}
   h (x + y) = a_1 h_1 (x + y) + a_2 h_2 (x + y) \leq 2 a_1 \big( h_1 (x) + h_1 (y) \big) + 2 a_2 \big( h_2 (x) + h_2 (y) \big) = 2 h (x) + 2 h (y).
\end{align*}
For a scaling factor $a > 1$, we have
\begin{align*}
  h (ax) = a_1 h_1 (ax) + a_2 h_2 (ax) \leq a^2 \big( a_1 h_1 (x) + a_2 h_2 (x) \big) = a^2 h (x),
\end{align*}
which proves that $h$ is also sub-quadratic. This concludes the proof of \Cref{lemma:quasi-sub-additive}.

\subsection{Proof of \Cref{lemma:random-radius-empirical-process}}\label{app:subsec-proof-lemma-random-radius-empirical-process}
Let $N \mydefn \lceil 2\log \numobs \rceil +  \lceil 2\log K \rceil$, and consider the doubling grid $\rho_k \mydefn 2^{-k} D$ for $k = 0, 1, \ldots, N - 1$. By union bound, with probability $1 - N \delta$, we have
\begin{align}
   Z (\rho_k) \leq s_{\delta} (\rho_k) ,\quad \mbox{for all } k = 0, 1, \ldots, N - 1.\label{eq:union-bound-random-radii}
\end{align}
Define the random index
\begin{align*}
  \widehat{k} \mydefn \max \big\{ k \in [N]: \rho_k \geq \rhohat \big\}.
\end{align*}
Since $\rhohat \in [0, D]$, the set of indices $\{ k \in [N]: \rho_k \geq \rhohat \}$ is non-empty, and $\widehat{k}$ is well-defined. By the definition of $\widehat{k}$, we have
\begin{align*}
  \rhohat \leq \rho_{\widehat{k}} \leq 2 \rhohat + 2^{- N} D.
\end{align*}
By monotonicity of the functions $Z (\cdot)$ and $s$, on the event that \eqref{eq:union-bound-random-radii} holds, we have
\begin{align*}
  Z (\rhohat) \leq Z (\rho_{\widehat{k}}) \leq s_\delta (\rho_{\widehat{k}}) \leq s_\delta (2 \rhohat + 2^{- N} D) = s_\delta \Big(2 \rhohat + \frac{D}{\numobs^2K^2} \Big) \leq s_\delta \big(2 \rhohat + \rhofix \big),
\end{align*}
where in the last step, we use monotonicity of the function $s$ and the fact that $\rhofix \geq \frac{D}{\numobs^2 K^2}$.

By sub-quadraticity of the function $s$, we have
\begin{align*}
  s_\delta \big(2 \rhohat + \rhofix \big) \leq 8 s_\delta (\rhohat) + 2 s_\delta (\rhofix) = 8 s_\delta (\rhohat) + 2  (\rhofix)^2.
\end{align*}
Furthermore, by sub-quadraticity of the function $s$, we have
\begin{align*}
 s_\delta (\rhohat) \leq s_\delta (\rhofix) \cdot \max \Big\{1, \Big( \frac{\rhohat}{\rhofix} \Big)^2 \Big\} \leq \frac{1}{8 a} (\rhofix)^2  \max \Big\{1, \Big( \frac{\rhohat}{\rhofix} \Big)^2 \Big\} = \frac{1}{8 a} \max \big\{ (\rhofix)^2, \rhohat^2 \big\}.
\end{align*}
Consequently, on the event that \Cref{eq:union-bound-random-radii} holds true, we have
\begin{align*}
  a Z (\rhohat) - \rhohat^2 \leq \max \big\{ (\rhofix)^2, \rhohat^2 \big\} + 2 a (\rhofix)^2 - \rhohat^2 \leq 3 a (\rhofix)^2,
\end{align*}
which proves the conclusion of \Cref{lemma:random-radius-empirical-process}.

\section{Proof for auxiliary results}\label{sec:proof-examples}
In this section, we prove some auxiliary results that support the examples in this paper.

\subsection{Proof of \Cref{prop:bandwidth-limited-func-norm-domination}}\label{subsec:app-proof-prop-bandwidth-limited-func-norm-domination}
Since $(\alpha_j)_{j = 1}^{+ \infty}$ is an orthonormal basis of $\ltwospace ([0, T])$, given the representation $f_t (x) = \sum_{j = 1}^m \alpha_j (x) \phi_j (t)$, we have
\begin{align*}
  \int_0^T \Exs \big[ |f_t (\State_t)|^2 \big] dt &= \int_0^T \Exs \Big[ |\sum_{j = 1}^m \alpha_j (\State_t) \phi_j (t) |^2 \Big] dt = \int_0^T \phi (t)^\top \Exs \big[ \alpha (\MyState_t) \alpha (\MyState_t)^\top \big] \phi(t) dt\\
  &\geq \frac{1}{c_1} \int_0^T \phi (t)^\top \cdot \int_0^T \Exs \big[ \alpha (\MyState_s) \alpha (\MyState_s)^\top \big] ds \cdot \phi (t) dt\\
  &= \frac{1}{c_1} \sum_{j = 1}^m \int_0^T \Exs \big[ \abss{\alpha_j (\MyState_t)}^2 \big] dt.
\end{align*}
On the other hand, we note that
\begin{align*}
  \int_0^T \Exs \big[ \abss{\partial_t f_t (\State_t)}^2 \big] dt &= \int_0^T \Exs \Big[ |\sum_{j = 1}^m \alpha_j (\State_t) \partial_t \phi_j (t) |^2 \Big] dt \\
  &\leq c_2 \int_0^T \partial_t \phi (t)^\top \cdot \int_0^T \Exs \big[ \alpha (\MyState_s) \alpha (\MyState_s)^\top \big] ds \cdot \partial_t \phi (t) dt\\
  &\leq c_2 \matsnorm{\int_0^T \Exs \big[ \alpha (\MyState_t) \alpha (\MyState_t)^\top \big] dt}{\mathrm{nuc}} \cdot \opnorm{\int_0^T \partial_t \phi (t) \partial_t \phi (t)^\top dt}\\
  &= c_2 \opnorm{\int_0^T \partial_t \phi (t) \partial_t \phi (t)^\top dt} \sum_{j = 1}^m \int_0^T \Exs \big[ \abss{\alpha_j (\MyState_t)}^2 \big] dt.
\end{align*}
For the operator norm, we note that each entry of the matrix satisfies
\begin{align*}
  \abss{\int_0^T \partial_t \phi_i(t) \partial_t \phi_j(t) dt} \leq \sqrt{\int_0^T \partial_t \phi_i(t)^2 dt} \cdot \sqrt{\int_0^T \partial_t \phi_j(t)^2 dt} \leq ij \leq m^2.
\end{align*}
Consequently, for any vector $u \in \sphere^{m - 1}$, we have 
\begin{align*}
  \vecnorm{\int_0^T \partial_t \phi (t) \partial_t \phi (t)^\top dt \cdot u}{2} \leq m^2 \sum_{j = 1}^{m} |u_j| \cdot \sqrt{m} \leq m^3,
\end{align*}
which leads to the operator norm bound $\opnorm{\int_0^T \partial_t \phi (t) \partial_t \phi (t)^\top dt} \leq m^3$. Putting the pieces together, we conclude that
\begin{align*}
  \int_0^T \Exs \big[ \abss{\partial_t f_t (\State_t)}^2 \big] dt \leq c_2 m^3 \sum_{j = 1}^m \int_0^T \Exs \big[ \abss{\alpha_j (\MyState_t)}^2 \big] dt \leq \frac{c_2}{c_1} m^3 \int_0^T \Exs \big[ |f_t (\State_t)|^2 \big] dt,
\end{align*}
which concludes the proof of \Cref{prop:bandwidth-limited-func-norm-domination}.

\subsection{Proof of \Cref{prop:example-vc-dimension}}\label{subsec:app-proof-prop-example-vc-dimension}
By substituting the covering number bound into the Dudley integral, we obtain
\begin{align*}
  \dudley_2 \big( \shiftConvSet (\rho); \strucnorm{\cdot}{T} \big) &\leq \int_0^\rho \sqrt{\log N \big( \varepsilon; \mathcal{G}; \vecnorm{\cdot}{\infty} \big)} d\varepsilon \leq \int_0^\rho \sqrt{d_0 \log \frac{1}{\varepsilon}} d\varepsilon \leq c \rho \sqrt{d_0},\\
  \dudley_2 (\convSet; \vecnorm{\cdot}{\Ctwospace}) &\leq \int_0^{\smoothness_\Fclass} \sqrt{\log N \big( \varepsilon; \mathcal{G}; \vecnorm{\cdot}{\infty} \big)} d\varepsilon \leq \int_0^{\smoothness_\Fclass} \sqrt{d_0 \log \frac{1}{\varepsilon}} d\varepsilon \leq c \smoothness_\Fclass \sqrt{d_0},\\
  \dudley_1 (\convSet; \vecnorm{\cdot}{\Ctwospace}) &\leq \int_0^{\smoothness_\Fclass} {\log N \big( \varepsilon; \mathcal{G}; \vecnorm{\cdot}{\infty} \big)} d\varepsilon \leq \int_0^{\smoothness_\Fclass} {d_0 \log \frac{1}{\varepsilon}} d\varepsilon \leq c \smoothness_\Fclass d_0
\end{align*}
for a universal constant $c > 0$.

Substituting to the definitions~\eqref{eq:defn-critical-radii} of the critical radii, for $\delta \asymp \frac{1}{\numobs^2 K^2}$, some algebra yields
\begin{align*}
  \critradistar_{\numobs, \delta} (\ExpValTrue - \fbar) &\lesssim \dynanorm{\ExpValTrue - \fbar}{1}{4 \pq}{T} \sqrt{\frac{d_0}{\numobs} \log K} + \sqrt{\vecnorm{\ExpValTrue - \fbar}{\Ctwospace} \frac{d_0 \smoothness_\Fclass}{\numobs} \log \numobs} \lesssim \sqrt{\vecnorm{\ExpValTrue - \fbar}{\Ctwospace}  \smoothness_\Fclass \frac{d_0\log (\numobs K)}{\numobs} },\\
  \critradicross_{\numobs, \delta}  &\lesssim \Big( \frac{\smoothness_\Fclass d_0 \log (\numobs K)}{\numobs} \Big)^{\frac{1}{2 - \eta}}, \quad \mbox{when } \frac{\numobs}{\log (\numobs K)} \gtrsim d_0, \quad \mbox{and}\\
  \critradismpl_{\numobs, K, \delta} &\lesssim \Big( \frac{\smoothness_\Fclass d_0}{\numobs K} \log (\numobs K) \Big)^{\frac{1}{2 (2 - \eta)}} + K^{-1/4} \sqrt{\frac{\smoothness_\Fclass d_0}{\numobs} \log (\numobs K)}.
\end{align*}
When $K \asymp \numobs^2$, the last term is dominated by the first two terms, and we conclude the proof of \Cref{prop:example-vc-dimension}.

\subsection{Proof of \Cref{prop:example-nonparametric}} \label{subsec:app-proof-prop-example-nonparametric}
Similar to the proof of \Cref{prop:example-vc-dimension}, by substituting the covering number bound into the Dudley integral, we obtain
\begin{align*}
    \dudley_2 \big( \shiftConvSet (\rho); \strucnorm{\cdot}{T} \big) &\leq \int_0^\rho \sqrt{\log N \big( \varepsilon; \mathcal{G}; \vecnorm{\cdot}{\infty} \big)} d\varepsilon \leq \int_0^\rho (c/\varepsilon)^{\beta_0 / 2} d\varepsilon \lesssim \rho^{1 - \beta_0 / 2},\\
    \dudley_1 (\convSet; \vecnorm{\cdot}{\Ctwospace}) &\leq \int_0^{\smoothness_\Fclass} {\log N \big( \varepsilon; \mathcal{G}; \vecnorm{\cdot}{\infty} \big)} d\varepsilon \leq \int_0^{\smoothness_\Fclass} \frac{c}{\varepsilon^{\beta_1}} d\varepsilon = O (1).
\end{align*}
Substituting to the definitions~\eqref{eq:defn-critical-radii} of the critical radii, for $\delta \asymp \frac{1}{\numobs^2 K^2}$, some algebra yields
\begin{align*}
  \critradistar_{\numobs, \delta} (\ExpValTrue - \fbar) &\lesssim  \Big(\dynanorm{\ExpValTrue - \fbar}{1}{4 \pq}{T}^2 \frac{\log \numobs K}{\numobs} \Big)^{\frac{1}{2 + \beta_0}} + \sqrt{\vecnorm{\ExpValTrue - \fbar}{\Ctwospace}\frac{ \log \numobs K}{\numobs}},\\
  \critradicross_{\numobs, \delta} & \lesssim \Big( \frac{\log \numobs K}{\numobs} \Big)^{1 / \beta_0} + \Big( \frac{\log \numobs K}{\numobs} \Big)^{1 / (2 - \eta)},\\
  \critradismpl_{\numobs, K, \delta} &\lesssim \Big(  \frac{\log \numobs K}{\numobs K}\Big)^{\frac{1}{2 (2 - \eta)}} + \Big(  \frac{\log \numobs K}{\numobs K}\Big)^{\frac{1}{2 + \beta_0}} + \Big( \frac{\log \numobs K}{\numobs \sqrt{K}} \Big)^{1/2}.
\end{align*}
When $K \asymp \numobs^2$, the last term is dominated by the first two terms, and we conclude the proof of \Cref{prop:example-nonparametric}.

\subsection{Proof of \Cref{prop:from-valfunc-to-policy}}\label{subsec:proof-from-valfunc-to-policy}
Consider a pair of diffusion processes
\begin{align*}
  d \State^{(1)}_t &= \big( \drift_t (\State^{(1)}_t) +  \policy_t^{(1)} (\State^{(1)}_t) \big) dt + \covMat_t^{1/2} d B_t,\\
  d \State^{(2)}_t &= \big( \drift_t (\State^{(2)}_t) +  \policy_t^{(2)} (\State^{(2)}_t) \big) dt + \covMat_t^{1/2} d B_t
\end{align*}
Let $\Prob^{(1)}$ and $\Prob^{(2)}$ be their probability distributions in the path space, respectively. By Girsanov's theorem, we have
\begin{multline*}
  \frac{d \Prob^{(1)}}{d \Prob^{(2)}}= \exp \Big( \int_0^T \big(\policy_t^{(2)} (\State^{(2)}_t) - \policy^{(1)}_t (\State^{(2)}_t) \big)^\top \covMat_t^{- 1/2} d B_t\\
   - \frac{1}{2} \int_0^T \eucnorm{\covMat_t^{- 1/2} \big(\policy_t^{(2)} (\State^{(2)}_t) - \policy^{(1)}_t (\State^{(2)}_t) \big)}^2 dt \Big).
\end{multline*}
Consequently, we can compute the KL divergence
\begin{align}
  \kull{\Prob^{(2)}}{\Prob^{(1)}} &= - \Exs_{\Prob^{(2)}} \Big[ \log \frac{d \Prob^{(1)}}{d \Prob^{(2)}} \Big] = \frac{1}{2} \int_0^T \Exs \Big[ \eucnorm{\covMat_t^{- 1/2} \big(\policy_t^{(2)} (\State^{(2)}_t) - \policy^{(1)}_t (\State^{(2)}_t) \big)}^2 \Big] dt \nonumber \\
  &=  \frac{1}{2} \int_0^T \Exs \Big[ \eucnorm{\covMat_t^{1/2}  \Big(\nabla \log \ExpVal_t^{(2)} (\State^{(2)}_t) - \nabla \log \ExpVal_t^{(1)} (\State^{(2)}_t) \Big)}^2  \Big] dt \nonumber \\
  &\leq \frac{\lammax}{2} \int_0^T \Exs \Big[ \eucnorm{\nabla \log \ExpVal_t^{(2)} (\State^{(2)}_t) - \nabla \log \ExpVal_t^{(1)} (\State^{(2)}_t)}^2 \Big] dt.\label{eq:kl-bound-in-valfunc-to-policy-proof}
\end{align}
\begin{subequations}
By Pinsker's inequality and variational formulation of the total variation distance, we have
\begin{align}
  \abss{\Exs \big[ \termReward (\State^{(1)}_T) \big] - \Exs \big[ \termReward (\State^{(2)}_T) \big]} &\leq 2 \totalvariation \big( \Prob^{(2)}, \Prob^{(1)}  \big) \leq \sqrt{2 \kull{\Prob^{(2)}}{\Prob^{(1)}}} \nonumber\\
  &\leq \Big\{ \lammax \int_0^T \Exs \Big[ \eucnorm{\nabla \log \ExpVal_t^{(2)} (\State^{(2)}_t) - \nabla \log \ExpVal_t^{(1)} (\State^{(2)}_t)}^2 \Big] dt \Big\}^{1/2},\label{eq:termReward-relative-bound-in-value-to-policy-proof}
\end{align}
and similarly, for every $t \in [0, T]$,
\begin{align}
  \abss{\Exs \big[ \reward_t (\State^{(1)}_t) \big] - \Exs \big[ \reward_t (\State^{(2)}_t) \big]} \leq  \rmax \Big\{ \lammax \int_0^T \Exs \Big[ \eucnorm{\nabla \log \ExpVal_t^{(2)} (\State^{(2)}_t) - \nabla \log \ExpVal_t^{(1)} (\State^{(2)}_t)}^2 \Big] dt \Big\}^{1/2}.\label{eq:instant-reward-relative-bound-in-value-to-policy-proof}
\end{align}
\end{subequations}
In order to bound the gradient difference between the true value function and the estimated value function. We use the following lemma.
\begin{lemma}\label{lemma:grad-diff-bound-in-value-to-policy-proof}
  Under the setup of \Cref{prop:from-valfunc-to-policy}, for any $\varepsilon > 0$, we have
  \begin{align*}
    \Exs \Big[ \eucnorm{\nabla \log \ExpVal_t^{(2)} (\State^{(2)}_t) - \nabla \log \ExpVal_t^{(1)} (\State^{(2)}_t)}^2 \Big] \leq 2 (1 + \smoothness_\Fclass) \exp \Big( \frac{2 + T }{\sfactor} +  \frac{ 2 \smoothness_\Fclass^2  T}{ \varepsilon^2 \sfactor^2}    \Big) \sobopstatnorm{\ExpVal_t^{(2)} - \ExpVal_t^{(1)}}{2 (1 + \varepsilon)}{\density_t}^2,
  \end{align*}
  for any $t \in [0, T]$.
\end{lemma}
\noindent See \Cref{subsubsec:proof-lemma-grad-diff-bound-in-value-to-policy-proof} for the proof of this lemma.

Taking this lemma as given, we now proceed with the proof of \Cref{prop:from-valfunc-to-policy}. By substituting the bounds in \Cref{lemma:grad-diff-bound-in-value-to-policy-proof} to \Cref{eq:termReward-relative-bound-in-value-to-policy-proof,eq:instant-reward-relative-bound-in-value-to-policy-proof}, we have
\begin{multline}
 \abss{\Big\{ \Exs \big[ \termReward (\State^{(2)}_T) \big] + \int_0^T \Exs \big[ \reward (\State^{(2)}_t) \big] dt \Big\} - \Big\{ \Exs \big[ \termReward (\State^{(1)}_T) \big] + \int_0^T \Exs \big[ \reward (\State^{(1)}_t) \big] dt \Big\}}\\
\leq  \frac{4 (1 + \smoothness_\Fclass)}{\sfactor} \exp \Big( \frac{2 + T }{\sfactor} +  \frac{ 2 \smoothness_\Fclass^2  T}{ \varepsilon^2 \sfactor^2}    \Big) \sqrt{\lammax \int_0^T \Exs \big[ \sobopstatnorm{\ExpVal_t^{(2)} - \ExpVal_t^{(1)}}{2 (1 + \varepsilon)}{\density_t}^2 \big] dt}.\label{eq:reward-diff-bound-in-value-to-policy-proof}
\end{multline}
As for the action-induced cost term, we have
\begin{align*}
  &\sfactor \abss{ \Exs \Big[ \policy^{(1)} (\State^{(1)}_t)^\top \covMat_t^{-1} \policy^{(1)} (\State^{(1)}_t) \Big] - \Exs \Big[ \policy^{(2)} (\State^{(2)}_t)^\top \covMat_t^{-1}  \policy^{(2)} (\State^{(2)}_t) \Big]}\\
  &=  \abss{\Exs \Big[ \Big( \big(\nabla \log \ExpVal_t^{(1)}  \big)^\top \covMat_t \big(\nabla \log \ExpVal_t^{(1)}  \big) \Big) (\State^{(1)}_t) \Big] - \Exs \Big[ \Big( \big(\nabla \log \ExpVal_t^{(2)}  \big)^\top \covMat_t \big(\nabla \log \ExpVal_t^{(2)}  \big) \Big) (\State^{(2)}_t) \Big]}\\
  &\leq\lammax \abss{\Exs \big[ \eucnorm{\nabla \log \ExpVal_t^{(1)} (\State^{(1)}_t )}^2 \big] - \Exs \big[ \eucnorm{\nabla \log \ExpVal_t^{(2)} (\State^{(2)}_t)}^2 \big]}
\end{align*}
We bound such a difference term in the following lemma.
\begin{lemma}\label{lemma:actcost-diff-bound-in-value-to-policy-proof}
  Under the setup of \Cref{prop:from-valfunc-to-policy}, we have
  \begin{multline*}
   \int_0^T  \abss{\Exs \big[ \eucnorm{\nabla \log \ExpVal_t^{(1)} (\State^{(1)}_t )}^2 \big] - \Exs \big[ \eucnorm{\nabla \log \ExpVal_t^{(2)} (\State^{(2)}_t)}^2 \big]} dt \\
   \leq 4 \Big\{ \frac{\smoothness_\Fclass^2 (T + 1)}{\sfactor^2} + 1\Big\}^2  \exp \Big( \frac{2 + T }{\sfactor} +  \frac{ 2 \smoothness_\Fclass^2  T}{ \varepsilon^2 \sfactor^2}    \Big)\sqrt{ \int_0^T \sobopstatnorm{\ExpVal_t^{(2)} - \ExpVal_t^{(1)}}{2 (1 + \varepsilon)}{\density_t}^2 dt}.
  \end{multline*}
\end{lemma}
\noindent See \Cref{subsubsec:proof-lemma-actcost-diff-bound-in-value-to-policy-proof} for the proof of this lemma.

Combining \Cref{eq:reward-diff-bound-in-value-to-policy-proof} and \Cref{lemma:actcost-diff-bound-in-value-to-policy-proof}, we have
\begin{align*}
  \abss{J (\policy^{(2)}) - J (\policy^{(1)})} \leq C (\sfactor, \varepsilon, T, \rmax)\sqrt{ \int_0^T \sobopstatnorm{\ExpVal_t^{(2)} - \ExpVal_t^{(1)}}{2 (1 + \varepsilon)}{\density_t}^2 dt},
\end{align*}
where we define the constant
\begin{align}
  C (\sfactor, \varepsilon, T, \rmax) \mydefn 8 (1 + \rmax\lammax) \Big\{ \frac{\smoothness_\Fclass^2 (T + 1)}{\sfactor^2} + 1\Big\}^2  \exp \Big( \frac{2 + T }{\sfactor} +  \frac{ 2 \smoothness_\Fclass^2  T}{ \varepsilon^2 \sfactor^2}    \Big) 
\end{align}

\subsubsection{Proof of \Cref{lemma:grad-diff-bound-in-value-to-policy-proof}}\label{subsubsec:proof-lemma-grad-diff-bound-in-value-to-policy-proof}
Let $\density_t^{(2)}$ be the one-time marginal density of $\State_t^{(2)}$. We note that
\begin{align*}
  &\Exs \Big[ \eucnorm{\nabla \log \ExpVal_t^{(2)} (\State_t^{(2)}) - \nabla \log \ExpVal_t^{(1)} (\State_t^{(2)})}^2 \Big] = \int \eucnorm{\frac{\nabla \ExpVal_t^{(2)}}{\ExpVal_t^{(2)}} - \frac{\nabla \ExpVal_t^{(1)}}{\ExpVal_t^{(1)}} }^2 \density_t^{(2)}  d \state\\
  &\leq 2 \int \frac{1}{(\ExpVal_t^{(1)})^2} \eucnorm{\nabla \ExpVal_t^{(2)} - \nabla \ExpVal_t^{(1)}}^2 \density_t^{(2)}  d \state + 2 \int \frac{\eucnorm{\nabla \ExpVal_t^{(2)}}^2}{(\ExpVal_t^{(2)} \ExpVal_t^{(1)})^2} \abss{\ExpVal_t^{(2)} - \ExpVal_t^{(1)}}^2 \density_t^{(2)} d \state.
\end{align*}
By our assumption, we have $\eucnorm{\nabla \log f_t (\state)} \leq \smoothness_\Fclass$. As for the optimal value function, we note that the na\"{i}ve policy $\policy_t (\state) \equiv 0$ achieves a value function of
\begin{align*}
  \ValFun_t^0 (\state) \geq - T - 2, \quad \mbox{for any }t \in [0, T], ~\state \in \real^\usedim.
\end{align*}
Therefore, we have
\begin{align*}
  \ExpVal_t^{(1)} (\state) \geq \exp \big(\ValFun_t^0 (\state) / \sfactor \big) \geq \exp \big( - (T + 2) / \sfactor \big).
\end{align*}
Substituting back to the gradient bound, we have
\begin{align*}
  \Exs \Big[ \eucnorm{\nabla \log \ExpVal_t^{(2)} (\State_t^{(2)}) - \nabla \log \ExpVal_t^{(1)} (\State_t^{(2)})}^2 \Big] \leq 2 (1 + \smoothness_\Fclass) \exp \big( \frac{2 + T }{\sfactor} \big) \sobonorm{\ExpVal_t^{(2)} - \ExpVal_t^{(1)}}{\density_t^{(2)}}^2.
\end{align*}
The Sobolev norm $\sobonorm{\ExpVal_t^{(2)} - \ExpVal_t^{(1)}}{\density_t^{(2)}}$ is defined with respect to the density $\density_t^{(2)}$, while the error guarantees in \Cref{thm:main-error} are defined with respect to the density $\density_t$ of the uncontrolled process. To relate the weighted norms under different distributions, we use the following upper bound on the R\'{e}nyi divergence.
  \begin{align}
    \int \Big( \frac{\density_t^{(2)}}{\density_t} (\state) \Big)^q \density_t (\state) d\state \leq \exp \Big( \frac{q^2 \smoothness_\Fclass^2}{2 \sfactor^2} T \Big)\quad \mbox{for any } q > 1.\label{eq:density-controlled-renyi-bound}
  \end{align}
We prove this bound at the end of this section.
Invoking H\"{o}lder's inequality with \Cref{eq:density-controlled-renyi-bound}, for any smooth function $f$, we have
\begin{align*}
  \sobonorm{f}{\density_t^{(2)}}^2 &= \int \Big\{ f^2 + \eucnorm{\nabla f}^2 \Big\} (x) \density_t^{(2)} (x) dx\\
  &\leq \Big\{ \int \Big\{ f^2 + \eucnorm{\nabla f}^2 \Big\}^{1 + \varepsilon} \density (x) dx \Big\}^{\frac{1}{1 + \varepsilon}} \cdot \int \Big( \frac{\density_t^{(2)}}{\density_t} (x) \Big)^{1 + \frac{1}{\varepsilon}} \density_t (x) dx\\
  &\leq \exp \Big[ \frac{ \smoothness_\Fclass^2}{ \sfactor^2}  \big(1 + \varepsilon^{-2} \big) T \Big]\Big\{ \int \Big\{ f^2 + \eucnorm{\nabla f}^2 \Big\}^{1 + \varepsilon} \density_t (x) dx \Big\}^{\frac{1}{1 + \varepsilon}} .
\end{align*}
So we have the bound
\begin{align*}
  \Exs \Big[ \eucnorm{\nabla \log \ExpVal_t^{(2)} (\State_t^{(2)}) - \nabla \log \ExpVal_t^{(1)} (\State_t^{(2)})}^2 \Big] \leq 2 (1 + \smoothness_\Fclass) \exp \Big( \frac{2 + T }{\sfactor} +  \frac{ 2 \smoothness_\Fclass^2  T}{ \varepsilon^2 \sfactor^2}    \Big) \sobopstatnorm{\ExpVal_t^{(2)} - \ExpVal_t^{(1)}}{2 (1 + \varepsilon)}{\density_t}^2.
\end{align*}
\paragraph{Proof of \Cref{eq:density-controlled-renyi-bound}:}
Let $\Prob_0$ denote the probability measure of the uncontrolled process~\eqref{eq:uncontrolled-process} in the path space. By Girsanov's theorem, we have
\begin{align*}
  \frac{d \Prob^{(2)}}{d \Prob_0} = \exp \Big( \int_0^T \policy_t^{(2)} (\State_t)^\top  \covMat_t^{- 1/2} d B_t - \frac{1}{2} \int_0^T \eucnorm{\covMat_t^{- 1/2}  \policy_t^{(2)} (\State_t) }^2 dt \Big).
\end{align*}
Note that
\begin{align*}
  \frac{\density_t^{(2)}}{\density_t} (\State_t) = \frac{d \Prob^{(2)}|_t}{d \Prob_0 |_t} = \Exs \Big[ \frac{d \Prob^{(2)}}{d \Prob_0} \mid \State_t \Big].
\end{align*}
Consequently, by Jensen's inequality, since the function $x \mapsto x^q$ is convex on $[0, + \infty)$, we have
\begin{multline*}
  \Exs \Big[\frac{\density_t^{(2)}}{\density_t} (\State_t)^q \Big] \leq \Exs \Big[ \Big( \frac{d \Prob^{(2)}}{d \Prob_0} \Big)^q \Big]
  \leq \Exs \Big[ \exp \Big( q \int_0^T \policy_t^{(2)} (\State_t)^\top  \covMat_t^{- 1/2} d B_t \Big) \Big]\\
  \leq \Exs \Big[ \exp \Big( \frac{q^2}{2 \sfactor^2} \int_0^T \eucnorm{\nabla \log f_t^{(2)} (\State_t) }^2 dt \Big) \Big]
  \leq \exp \Big( \frac{q^2 \smoothness_\Fclass^2}{2 \sfactor^2} T \Big),
\end{multline*}
which completes the proof of \Cref{eq:density-controlled-renyi-bound}.

\subsubsection{Proof of \Cref{lemma:actcost-diff-bound-in-value-to-policy-proof}}\label{subsubsec:proof-lemma-actcost-diff-bound-in-value-to-policy-proof}
We decompose the error as
\begin{align}
  &\abss{\Exs \big[ \eucnorm{\nabla \log \ExpVal^{(1)}_t (\State^{(1)}_t )}^2 \big] - \Exs \big[ \eucnorm{\nabla \log \ExpVal^{(2)}_t (\State^{(2)}_t)}^2 \big]} \nonumber\\
  & \leq \abss{\Exs \big[ \eucnorm{\nabla \log \ExpVal^{(1)}_t (\State^{(1)}_t )}^2 \big] - \Exs \big[ \eucnorm{\nabla \log \ExpVal^{(1)}_t (\State^{(2)}_t)}^2 \big]} + \abss{\Exs \big[ \eucnorm{\nabla \log \ExpVal^{(1)}_t (\State^{(2)}_t )}^2 \big] - \Exs \big[ \eucnorm{\nabla \log \ExpVal^{(2)}_t (\State^{(2)}_t)}^2 \big]} .\label{eq:error-decomposition-in-actcost-diff-bound-in-value-to-policy-proof}
\end{align}
For the first term, we note that $\eucnorm{\nabla \log \ExpVal^{(1)}_t} \leq \smoothness_\Fclass$.
\begin{align*}
  \abss{\Exs \big[ \eucnorm{\nabla \log \ExpVal^{(1)}_t (\State^{(1)}_t )}^2 \big] - \Exs \big[ \eucnorm{\nabla \log \ExpVal^{(1)}_t (\State^{(2)}_t)}^2 \big] } \leq \smoothness_\Fclass^2 \totalvariation \big( \Prob^{(1)}, \Prob^{(2)} \big) \leq \smoothness_\Fclass^2 \sqrt{\frac{1}{2} \kull{\Prob^{(1)}}{\Prob^{(2)}}}.
\end{align*}
Invoking \Cref{eq:kl-bound-in-valfunc-to-policy-proof}, we have that
\begin{align*}
  \abss{\Exs \big[ \eucnorm{\nabla \log \ExpVal^{(1)}_t (\State^{(1)}_t )}^2 \big] - \Exs \big[ \eucnorm{\nabla \log \ExpVal^{(1)}_t (\State^{(2)}_t)}^2 \big] } \leq \frac{\smoothness_\Fclass^2 }{\sfactor^2} \sqrt{\int_0^T \Delta_s^2 ds}.
\end{align*}
For the second term in the decomposition~\eqref{eq:error-decomposition-in-actcost-diff-bound-in-value-to-policy-proof}, we use Cauchy--Schwarz inequality to obtain that
\begin{align*}
  &\abss{\Exs \big[ \eucnorm{\nabla \log \ExpVal^{(1)}_t (\State^{(2)}_t )}^2 \big] - \Exs \big[ \eucnorm{\nabla \log \ExpVal^{(2)}_t (\State^{(2)}_t)}^2 \big]}\\
  &\leq \int \eucnorm{\nabla \log \ExpVal^{(2)}_t (\state) - \nabla \log \ExpVal^{(1)}_t (\state)}^2 \density_t^{(2)} (\state) d\state + 2 \int \eucnorm{\nabla \log \ExpVal^{(2)}_t (\state) - \nabla \log \ExpVal^{(1)}_t (\state)} \cdot \eucnorm{\nabla \log \ExpVal^{(2)}_t (\state)} \density_t^{(2)} (\state) d\state\\
  &\leq  \int \eucnorm{\nabla \log \ExpVal^{(2)}_t (\state) - \nabla \log \ExpVal^{(1)}_t (\state)}^2 \density_t^{(2)} (\state) d\state + 2 \smoothness_\Fclass \sqrt{\int \eucnorm{\nabla \log \ExpVal^{(2)}_t (\state) - \nabla \log \ExpVal^{(1)}_t (\state)}^2 \density_t^{(2)}(\state) d\state}.
\end{align*}
For notational simplicity, let us define
\begin{align*}
  \Delta_t \mydefn  \sqrt{\Exs \Big[ \eucnorm{\nabla \log \ExpVal^{(2)}_t (\State^{(2)}_t) - \nabla \log \ExpVal^{(1)}_t (\State^{(2)}_t)}^2 \Big]},
\end{align*}
The decomposition~\eqref{eq:error-decomposition-in-actcost-diff-bound-in-value-to-policy-proof} implies that
\begin{align*}
  \abss{\Exs \big[ \eucnorm{\nabla \log \ExpVal^{(1)}_t (\State^{(1)}_t )}^2 \big] - \Exs \big[ \eucnorm{\nabla \log \ExpVal^{(2)}_t (\State^{(2)}_t)}^2 \big]} \leq  \frac{\smoothness_\Fclass^2 }{\sfactor^2} \sqrt{\int_0^T \Delta_s^2 ds} + 2 \smoothness_\Fclass \Delta_t + \Delta_t^2.
\end{align*}
Integrating with respect to time, we have
\begin{align*}
  &\int_0^T \abss{\Exs \big[ \eucnorm{\nabla \log \ExpVal^{(1)}_t (\State^{(1)}_t )}^2 \big] - \Exs \big[ \eucnorm{\nabla \log \ExpVal^{(2)}_t (\State^{(2)}_t)}^2 \big]} dt \\
  &\leq \frac{\smoothness_\Fclass^2 T}{\sfactor^2}\sqrt{\int_0^T \Delta_s^2 ds} + 2 \smoothness_\Fclass \int_0^T \Delta_t dt + \int_0^T \Delta_t^2 dt\\
  &\leq \Big\{ \frac{\smoothness_\Fclass^2 T}{\sfactor^2} + 2 \smoothness_\Fclass  \sqrt{T} \Big\}\sqrt{\int_0^T \Delta_t^2 dt} + \int_0^T \Delta_t^2 dt,
\end{align*}
where in the second step, we use Cauchy--Schwarz inequality to bound the term $ \int_0^T \Delta_t dt$.

Invoking \Cref{lemma:grad-diff-bound-in-value-to-policy-proof}, we have
\begin{align*}
  \int_0^T \Delta_t^2 dt \leq2 (1 + \smoothness_\Fclass) \exp \Big( \frac{2 + T }{\sfactor} +  \frac{ 2 \smoothness_\Fclass^2  T}{ \varepsilon^2 \sfactor^2}    \Big) \int_0^T \sobopstatnorm{\ExpVal^{(2)}_t - \ExpVal^{(1)}_t}{2 (1 + \varepsilon)}{\density_t}^2 dt.
\end{align*}
By assumption of \Cref{prop:from-valfunc-to-policy}, $\int_0^T \sobopstatnorm{\ExpVal^{(2)}_t - \ExpVal^{(1)}_t}{2 (1 + \varepsilon)}{\density_t}^2 dt \leq 1$. Substituting back to the bound above, we conclude that
\begin{multline*}
  \int_0^T \abss{\Exs \big[ \eucnorm{\nabla \log \ExpVal^{(1)}_t (\State^{(1)}_t )}^2 \big] - \Exs \big[ \eucnorm{\nabla \log \ExpVal^{(2)}_t (\State^{(2)}_t)}^2 \big]} dt\\
   \leq  4 \Big\{ \frac{\smoothness_\Fclass^2 (T + 1)}{\sfactor^2} + 1\Big\}^2  \exp \Big( \frac{2 + T }{\sfactor} +  \frac{ 2 \smoothness_\Fclass^2  T}{ \varepsilon^2 \sfactor^2}    \Big)\sqrt{ \int_0^T \sobopstatnorm{\ExpVal^{(2)}_t - \ExpVal^{(1)}_t}{2 (1 + \varepsilon)}{\density_t}^2 dt},
\end{multline*}
which completes the proof of \Cref{lemma:actcost-diff-bound-in-value-to-policy-proof}.

\end{document}